\documentclass{article}
\usepackage{iclr2022_conference,times}
\usepackage{graphicx}
\usepackage{multirow}
\usepackage{xcolor}
\definecolor{C0}{HTML}{3182ce}
\definecolor{C1}{HTML}{dd6b20}
\usepackage{hyperref}

\definecolor{grey}{rgb}{0.8,0.8,0.8}
\definecolor{green}{rgb}{0.0,0.65,0.0}
\definecolor{lgreen}{rgb}{0.5,0.93,0.5}
\definecolor{red}{rgb}{0.8,0.0,0.0}
\definecolor{lred}{rgb}{0.93,0.5,0.5}
\definecolor{darkblue}{rgb}{0, 0, 0.5}
\hypersetup{colorlinks=true, citecolor=darkblue, linkcolor=darkblue, urlcolor=darkblue}

\usepackage{wrapfig,booktabs}
\usepackage{subcaption}
\usepackage{amsthm}
\usepackage{mathrsfs}
\usepackage{verbatim}

\usepackage{amsmath,amssymb}

\usepackage{sidecap}
\usepackage[utf8]{inputenc} 
\usepackage[T1]{fontenc}    
\usepackage{url}            
\usepackage{amsfonts}       
\usepackage{nicefrac}       
\usepackage{microtype}      
\usepackage{algorithm}
\usepackage{forloop}
\usepackage{algorithmic}

\usepackage{mathtools}
\usepackage{bm}

\DeclarePairedDelimiter\norm{\lVert}{\rVert}

\usepackage{bbm}
\def\one{\mathbbm{1}}

\title{\textsc{Scarf}: Self-Supervised Contrastive Learning using Random Feature Corruption}

\author{Dara Bahri, Heinrich Jiang, Yi Tay, Donald Metzler\\
Google Research\\
\texttt{\{dbahri,heinrichj,yitay,metzler\}@google.com}\\
}

\iclrfinalcopy
\begin{document}

\maketitle

\begin{abstract}
Self-supervised contrastive representation learning has proved incredibly successful in the vision and natural language domains, enabling state-of-the-art performance with orders of magnitude less labeled data. However, such methods are domain-specific and little has been done to leverage this technique on real-world \emph{tabular} datasets. We propose \textsc{Scarf}, a simple, widely-applicable technique for contrastive learning, where views are formed by corrupting a random subset of features. When applied to pre-train deep neural networks on the 69 real-world, tabular classification datasets from the OpenML-CC18 benchmark, \textsc{Scarf} not only improves classification accuracy in the fully-supervised setting but does so also in the presence of label noise and in the semi-supervised setting where only a fraction of the available training data is labeled. We show that \textsc{Scarf} complements existing strategies and outperforms alternatives like autoencoders. We conduct comprehensive ablations, detailing the importance of a range of factors.
\end{abstract}

\section{Introduction}

In many machine learning tasks, unlabeled data is abundant but labeled data is costly to collect, requiring manual human labelers. The goal of self-supervised learning is to leverage large amounts of unlabeled data to learn useful representations for downstream tasks such as classification. Self-supervised learning has proved critical in computer vision~\citep{grill2020bootstrap,misra2020self,he2020momentum,tian2019contrastive} and natural language processing~\citep{song2020mpnet,wang2019denoising,raffel2019exploring}. Some recent examples include the following: \citet{chen2020simple} showed that training a linear classifier on the representations learned by their proposed method, SimCLR, significantly outperforms previous state-of-art image classifiers and requires 100x fewer labels to do so; \citet{brown2020language} showed through their GPT-3 language model that by pre-training on a large corpus of text, only few labeled examples were required for task-specific fine-tuning for a wide range of tasks.

A common theme of these advances is learning representations that are robust to different views or distortions of the same input; this is often achieved by maximizing the similarity between views of the same input and minimizing those of different inputs via a contrastive loss. However, techniques to generate views or corruptions have thus far been, by and large, domain-specific (e.g. color distortion~\citep{zhang2016colorful} and cropping~\citep{chen2020simple} in vision, and token masking~\citep{song2020mpnet} in NLP). Despite the importance of self-supervised learning, there is surprisingly little work done in finding methods that are applicable across domains and in particular, ones that can be applied to tabular data.

In this paper, we propose \textsc{Scarf}, a simple and versatile contrastive pre-training procedure. We generate a view for a given input by selecting a random subset of its features and replacing them by random draws from the features' respective empirical marginal distributions.
Experimentally, we test \textsc{Scarf} on the OpenML-CC18 benchmark~\citep{OpenML2013,bischl2017openml,OpenMLPython2019}, a collection of 72 real-world classification datasets. We show that not only does \textsc{Scarf} pre-training improve classification accuracy in the fully-supervised setting but does so also in the presence of label noise and in the semi-supervised setting where only a fraction of the available training data is labeled. Moreover, we show that combining \textsc{Scarf} pre-training with other solutions to these problems further improves them, demonstrating the versatility of \textsc{Scarf} and its ability to learn effective task-agnostic representations. We then conduct extensive ablation studies, showing the effects of various design choices and stability to hyperparameters. Our ablations show that \textsc{Scarf}'s way of constructing views is more effective than alternatives. We show that \textsc{Scarf} is less sensitive to feature scaling and is stable to various hyperparameters such as batch size, corruption rate, and softmax temperature.

\begin{figure}[!t]
\centering
\includegraphics[trim=0 0 0 0,clip,width=0.90\linewidth]{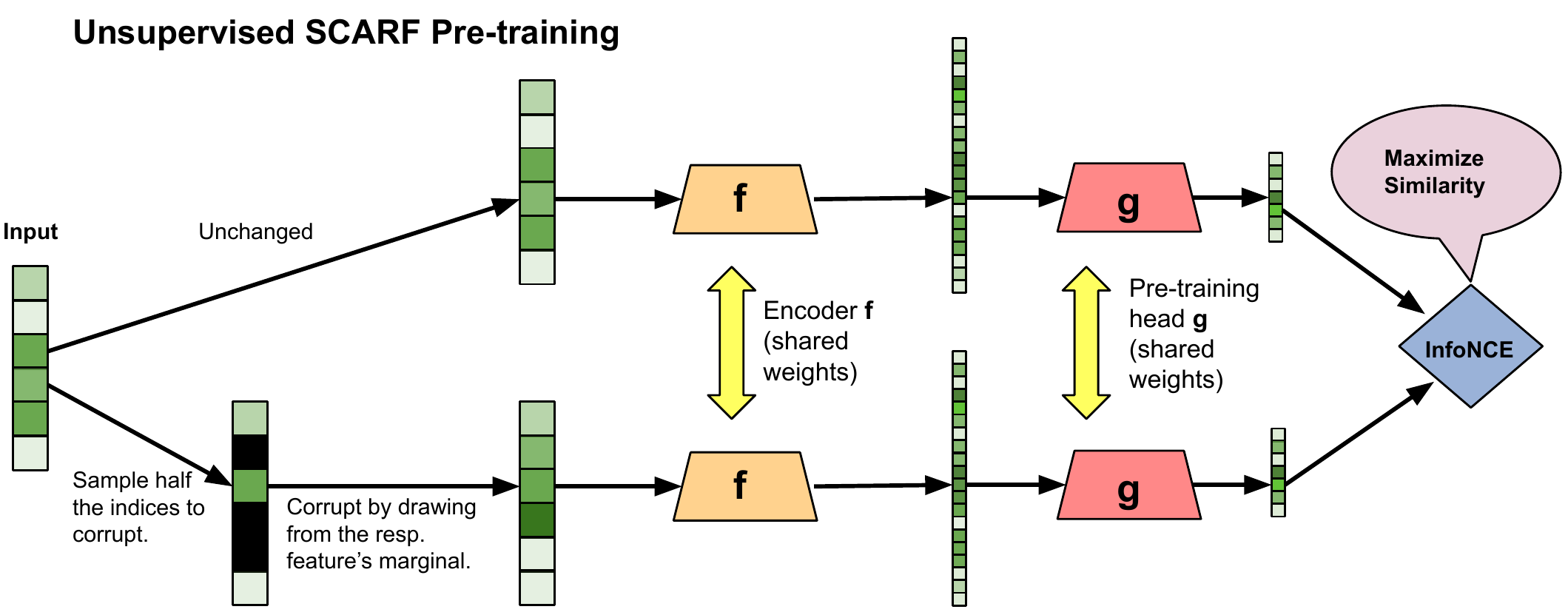} \\
\includegraphics[trim=0 0 0 0,clip,width=0.90\linewidth]{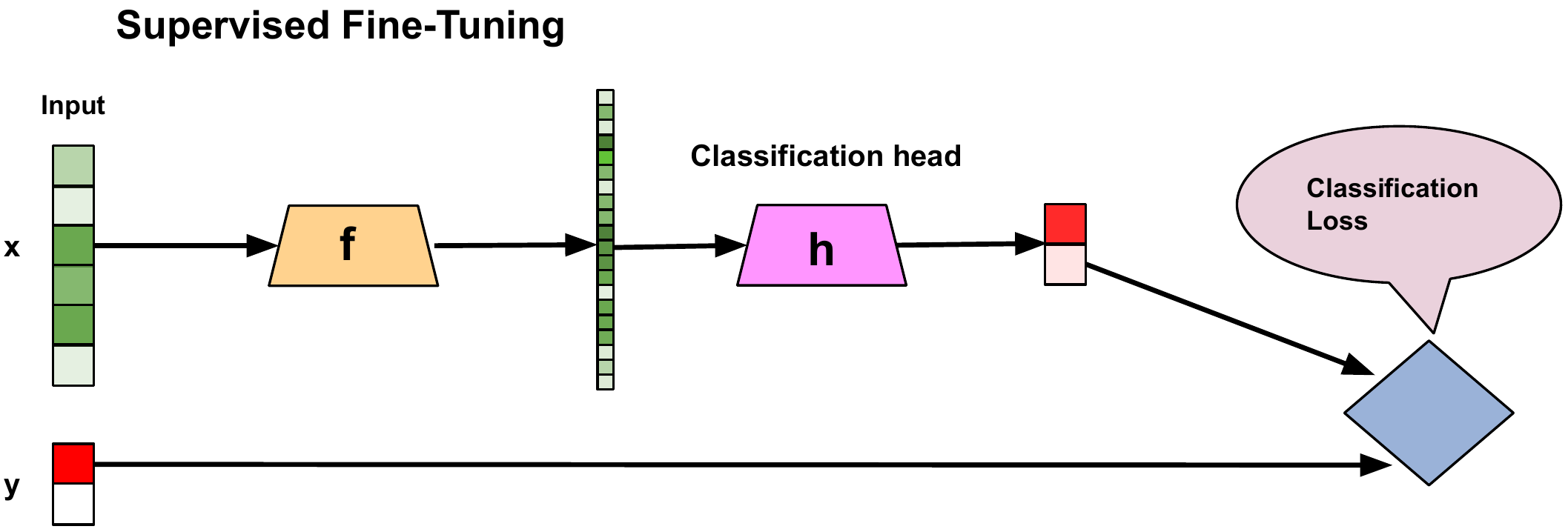} \\
\caption{\label{fig:arch} Diagram showing unsupervised \textsc{Scarf} pre-training ({\bf Top}) and subsequent supervised fine-tuning ({\bf Bottom}). During pre-training, networks $f$ and $g$ are learned to produce good representations of the input data. After pre-training, $g$ is discarded and a classification head $h$ is applied on top of the learned $f$ and both $f$ and $h$ are subsequently fine-tuned for classification.}
\end{figure}

\section{Related Works}

A number of self-supervised learning techniques have been proposed in computer vision~\citep{zhai2019s4l,tung2017self,jing2020self}. One framework involves learning features based on generated images through various methods, including using a GAN~\citep{goodfellow2014generative,donahue2016adversarial,radford2015unsupervised,chen2018self}, predicting pixels~\citep{larsson2016learning}, predicting colorizations \citep{zhang2016colorful,larsson2017colorization}, ensuring local and global consistency~\citep{iizuka2017globally}, and learning synthetic artifacts~\citep{jenni2018self}.
Most related to our approach are contrastive learning ones ~\citep{tian2019contrastive,hassani2020contrastive,oord2018representation,henaff2020data,li2016unsupervised,he2020momentum,bojanowski2017unsupervised,wang2015unsupervised,gidaris2018unsupervised}. In particular, our framework is similar to SimCLR~\citep{chen2020simple}, which involves generating views of a single image via image-based corruptions like random cropping, color distortion and blurring; however, we generate views that are applicable to tabular data.

Self-supervised learning has had an especially large impact in language modeling~\citep{qiu2020pre}. One popular approach is masked language modeling, wherein the model is trained to predict input tokens that have been intentionally masked out~\citep{devlin2018bert,raffel2019exploring,song2019mass} as well as enhancements to this approach~\citep{liu2019roberta,dong2019unified,bao2020unilmv2,lample2019cross,joshi2020spanbert} and variations involving permuting the tokens~\citep{yang2019xlnet,song2020mpnet}. Denoising autoencoders have been used by training them to reconstruct the input from a corrupted version (produced by, for example, token masking, deletion, and infilling)~\citep{lewis2019bart,wang2019denoising,freitag2018unsupervised}. Contrastive approaches include randomly replacing words and distinguishing between real and fake phrases~\citep{collobert2011natural,mnih2013learning}, random token replacement~\citep{mikolov2013distributed,clark2020electra}, and adjacent sentences~\citep{joshi2020spanbert,lan2019albert,de2019bertje}.

Within the contrastive learning framework, the choice of loss function is significant. InfoNCE~\citep{gutmann2010noise,oord2018representation}, which can be interpreted as a
non-parametric estimation of the entropy of the representation~\citep{wang2020understanding}, is a popular choice. Since then there have been a number of proposals~\citep{zbontar2021barlow,grill2020bootstrap,hjelm2018learning}; however, we show that InfoNCE is effective for our framework.

Recently, \citet{yao2020self} adapted the contrastive framework to large-scale recommendation systems in a way similar to our approach. The key difference is in the way the methods generate multiple views. \citet{yao2020self} proposes masking random features in a correlated manner and applying a dropout for categorical features, while our approach involves randomizing random features based on the features' respective empirical marginal distribution (in an uncorrelated way). Generating such views for a task is a difficult problem: there has been much work done in understanding and designing them~\citep{wu2018unsupervised,purushwalkam2020demystifying,gontijo2020affinity,lopes2019improving,perez2017effectiveness,park2019specaugment} and learning them~\citep{ratner2017learning,cubuk2020randaugment,ho2019population,lim2019fast,zhang2019adversarial,tran2017bayesian,tamkin2020viewmaker}.

Lastly, also similar to our work is VIME~\citep{yoon2020vime}, which proposes the same corruption technique for tabular data that we do.
They pre-train an encoder network on unlabeled data by attaching ``mask estimator'' and ``feature estimator'' heads on top of the encoder state and teaching the model to recover both the binary mask that was used for corruption as well as the original uncorrupted input, given the corrupted input. The pre-trained encoder network is subsequently used for semi-supervised learning via attachment of a task-specific head and minimization of the supervised loss as well as an auto-encoder reconstruction loss. VIME was shown to achieve state-of-art results on genomics and clinical datasets.
The key differences with our work is that we pre-train using a contrastive loss, which we show to be more effective than the denoising auto-encoder loss that partly constitutes VIME. Furthermore, after pre-training we fine-tune all model weights, including the encoder (unlike VIME, which only fine-tunes the task head), and we do so using task supervision only. 


\section{\textsc{Scarf}}

\begin{algorithm}[!t]
\caption{\label{alg:main} \textsc{Scarf} pre-training algorithm.}
\begin{algorithmic}[1]
    \STATE \textbf{input:} unlabeled training data $\mathcal{X} \subseteq \mathbb{R}^M$, batch size $N$, temperature $\tau$, corruption rate $c$, encoder network $f$, pre-train head network $g$.
    \STATE let $\widehat{\mathcal{X}_j}$ be the uniform distribution over $\mathcal{X}_j = \{x_j : x \in \mathcal{X}\}$, where $x_j$ denotes the $j$-th coordinate of $x$.
    \STATE let $q = \lfloor c\cdot  M\rfloor$ be the number of features to corrupt.
    \FOR{sampled mini-batch $\left\{x^{(i)}\right\}_{i=1}^N \subseteq \mathcal{X}$}
    \STATE for $i \in [N]$, uniformly sample subset $\mathcal{I}_i$ from $\{1,...,M\}$ of size $q$ and define  $\tilde{x}^{(i)} \in \mathbb{R}^M$ as follows:
    $\tilde{x}^{(i)}_j = x_j$ if $j \not\in \mathcal{I}_i$, otherwise $\tilde{x}^{(i)}_j = v$, where $v \sim \widehat{\mathcal{X}_j}$.  \textcolor{gray}{~~~~~\# generate corrupted view.}
    \STATE let $z^{(i)} = g\left(f\left(x^{(i)}\right)\right)$, $\tilde{z}^{(i)} = g\left(f\left(\tilde{x}^{(i)}\right)\right)$, for $i \in [N]$.  \textcolor{gray}{~~~~~~~~~~~~~~~~~~~~~~\# embeddings for views.}    
    \STATE let $s_{i,j} = {z^{(i)}}^\top \tilde{z}^{(j)} / \left(\norm{z^{(i)}}_2 \cdot \norm{\tilde{z}^{(j)}}_2\right)$, for $i, j \in [N]$. \textcolor{gray}{~~~~~~~~~~~~~~~~~~~~~~~~~~\# pairwise similarity.}\\
    \STATE define $\mathcal{L}_{\text{cont}} := \frac{1}{N} \sum_{i=1}^N -\log\left( \frac{\exp(s_{i,i}/\tau)}{\frac{1}{N}\sum_{k=1}^{N} \exp(s_{i, k}/\tau)}\right)$.
    \STATE update networks $f$ and $g$ to minimize $\mathcal{L}_{\text{cont}}$ using SGD.
    \ENDFOR
    \RETURN{encoder network $f$.}
\end{algorithmic}
\end{algorithm}

We now describe our proposed method (Algorithm~\ref{alg:main}), which is also described in Figure~\ref{fig:arch}. For each mini-batch of examples from the unlabeled training data, we generate a corrupted version $\tilde{x}^{(i)}$ for each example $x^{(i)}$ as follows. We sample some fraction of the features uniformly at random and replace each of those features by a random draw from that feature's empirical marginal distribution, which is defined as the uniform distribution over the values that feature takes on across the training dataset. Then, we pass both $x^{(i)}$ and $\tilde{x}^{(i)}$ through the encoder network $f$, whose output we pass through the pre-train head network $g$, to get $z^{(i)}$ and $\tilde{z}^{(i)}$ respectively. Note that the pre-train head network $\ell_2$-normalizes the outputs so that they lie on the unit hypersphere -- this has been found to be crucial in practice~\citep{chen2020simple,wang2020understanding}. We train on the InfoNCE contrastive loss, encouraging $z^{(i)}$ and $\tilde{z}^{(i)}$ to be close for all $i$ and $z^{(i)}$ and $\tilde{z}^{(j)}$ to be far apart for $i\neq j$, and we optimize over the parameters of $f$ and $g$ via SGD. 

Then, to train a classifier for the task via fine-tuning, we take the encoder network $f$ and attach a classification head $h$ which takes the output of $f$ as its input and predicts the label of the example. We optimize the cross-entropy classification loss and tune the parameters of both $f$ and $h$.

While pre-training can be run for a pre-determined number of epochs, much like normal supervised training, how many is needed largely depends on the model and dataset. To this end, we propose using early stopping on the validation InfoNCE loss. Given unlabeled validation data, we cycle through it for some number of epochs, running our proposed method to generate $\left(x^{(i)}, \tilde{x}^{(i)}\right)$ pairs. Once built, the loss on this static set is tracked during pre-training. Prototypical loss curves are shown in the Appendix.

\section{Experiments}
We evaluate the impact of \textsc{Scarf} pre-training on test accuracy after supervised fine-tuning in three distinct settings: on the full dataset, on the full dataset but where only 25\% of samples have labels and the remaining 75\% do not, and on the full dataset where 30\% of samples undergo label corruption.

{\bf Datasets.} We use 69 datasets from the public OpenML-CC18 benchmark\footnote{\url{https://docs.openml.org/benchmark/}} under the CC-BY licence. It consists of 72 real-world classification datasets that have been manually curated for effective benchmarking. Since we're concerned with tabular datasets in this work, we remove MNIST, Fashion-MNIST, and CIFAR10. For each OpenML dataset, we form $70\%/10\%/20\%$ train/validation/test splits, where a different split is generated for every trial and all methods use the same splits. The percentage used for validation and test are never changed and only training labels are corrupted for the label noise experiments.

{\bf Dataset pre-processing}. We represent categorical features by a one-hot encoding, and most of the corruption methods explored in the ablations are on these one-hot encoded representations of the data (with the exception of \textsc{Scarf}, where the marginal sampling is done before one-hot encoding). We pre-process missing data as follows: if a feature column is always missing, we drop it. Otherwise, if the feature is categorical, we fill in missing entries with the mode, or most frequent, category computed over the full dataset. For numerical features, we impute it with the mean. We explore scaling numerical features by z-score, min-max, and mean normalization. We find that for a vanilla network (i.e. control), z-score normalization performed the best for all but three datasets (OpenML dataset ids 4134, 28, and 1468), for which no scaling was optimal. We thus do not scale these three datasets and z-score normalize all others.

{\bf Model architecture and training}.
Unless specified otherwise, we use the following settings across all experiments.
As described earlier, we decompose the neural network into an encoder $f$, a pre-training head $g$, and a classification head $h$, where the inputs to $g$ and $h$ are the outputs of $f$. We choose all three component models to be ReLU networks with hidden dimension $256$. $f$ consists of 4 layers, whereas both $g$ and $h$ have 2 layers. Both \textsc{Scarf} and the autoencoder baselines use $g$ (for both pre-training and co-training, described later), but for autoencoders, the output dimensionality is the input feature dimensionality, and the mean-squared error reconstruction loss is applied. We train all models and their components with the Adam optimizer using the default learning rate of $0.001$. For both pre-training and fine-tuning we use 128 batch size. Unsupervised pre-training methods all use early stopping with patience 3 on the validation loss, unless otherwise noted. Supervised fine-tuning uses this same criterion (and validation split), but classification error is used as the validation metric for early stopping, as it performs slightly better. We set a max number of fine-tune epochs of $200$ and pre-train epochs of $1000$, We use 10 epochs to build the static validation set. Unless otherwise noted, we use a corruption rate $c$ of 0.6 and a temperature $\tau$ of 1, for \textsc{Scarf}-based methods. All runs are repeated 30 times using different train/validation/test splits. Experiments were run on a cloud cluster of CPUs, and we used about one million CPU core hours in total for the experiments.

\begin{figure}[!t]
    \centering
    \includegraphics[width=0.49\textwidth]{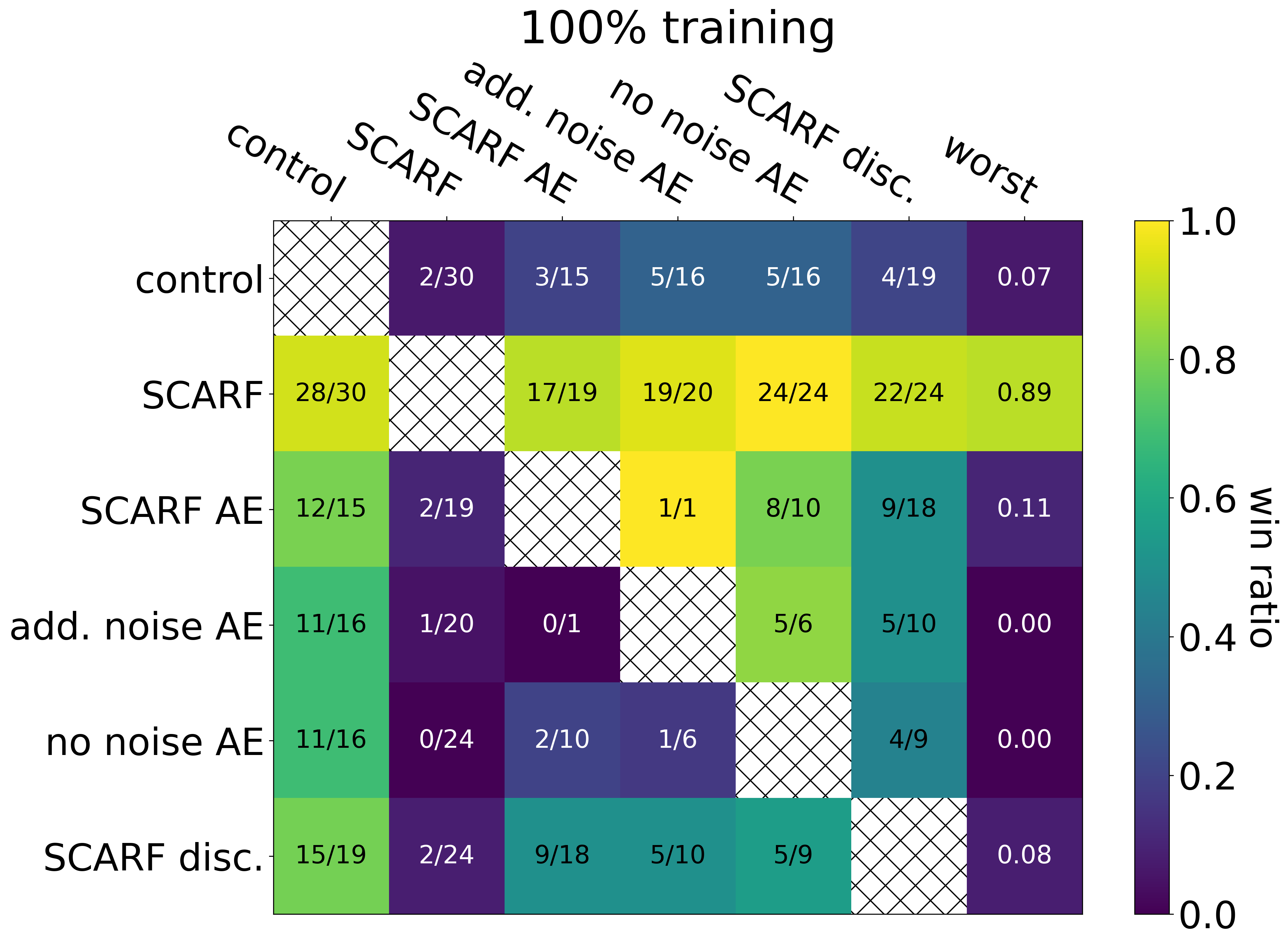}
    \includegraphics[width=0.49\textwidth]{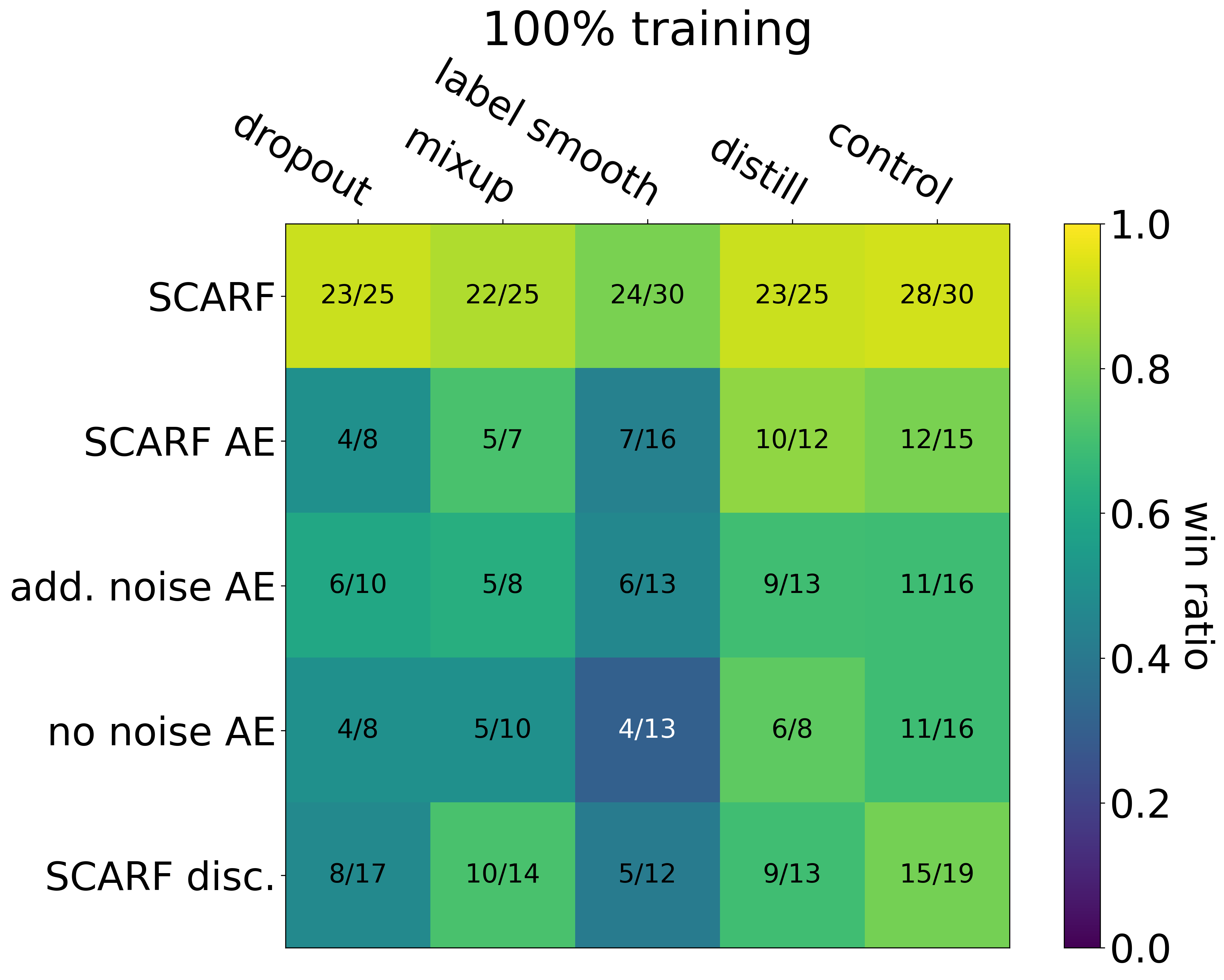}\\
    \includegraphics[width=0.32\textwidth]{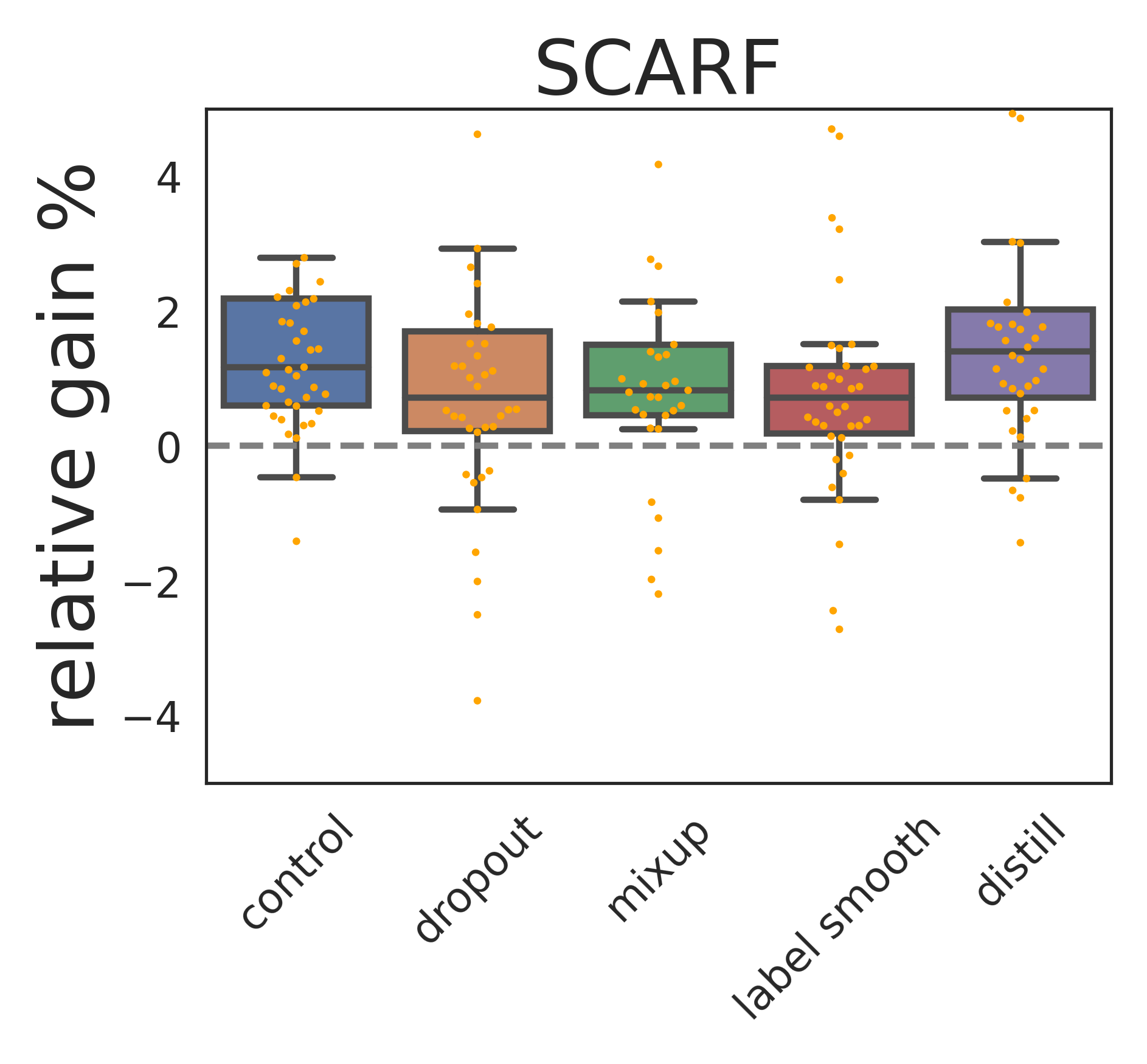} \includegraphics[width=0.32\textwidth]{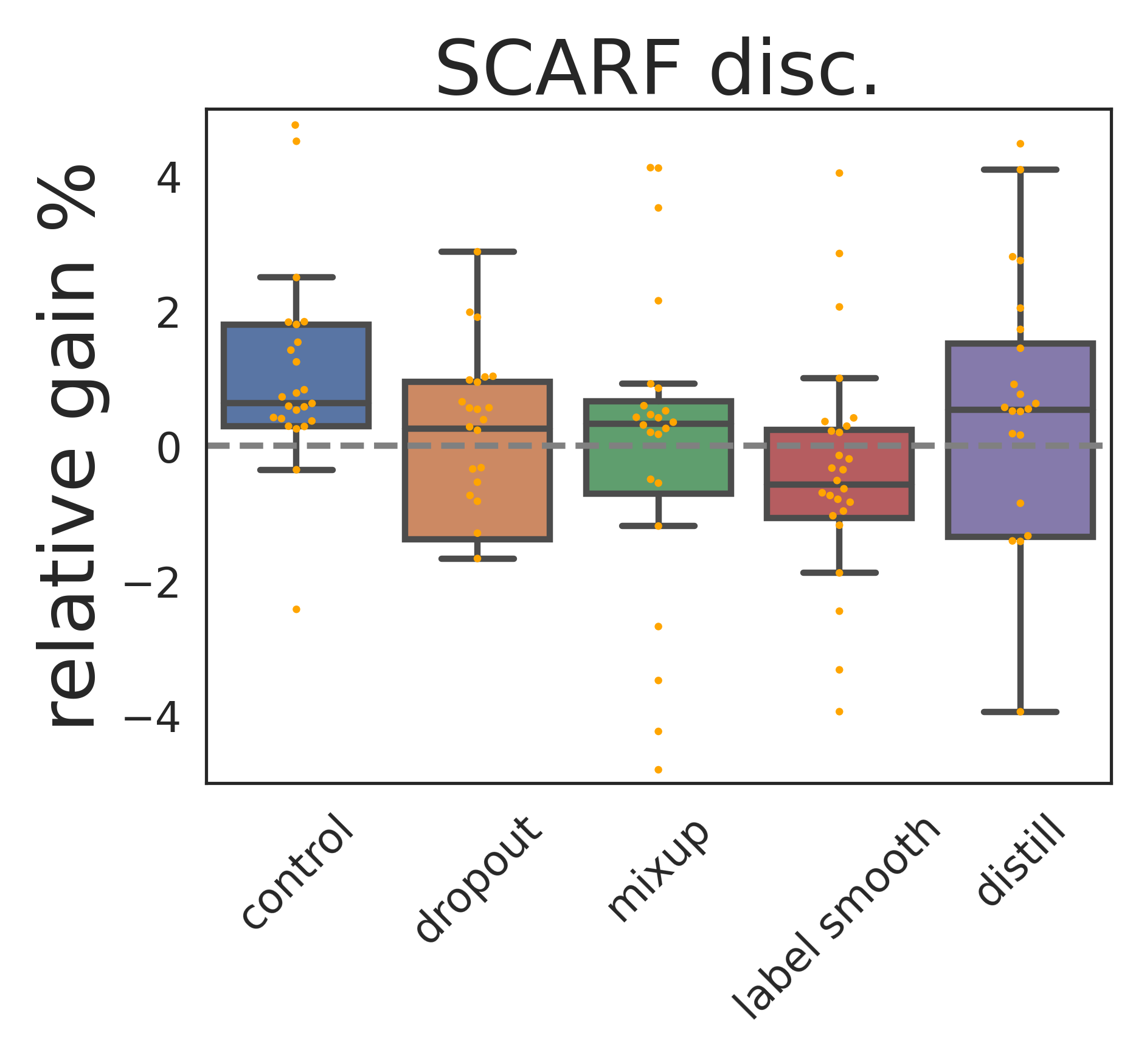}\\
    \includegraphics[width=0.32\textwidth]{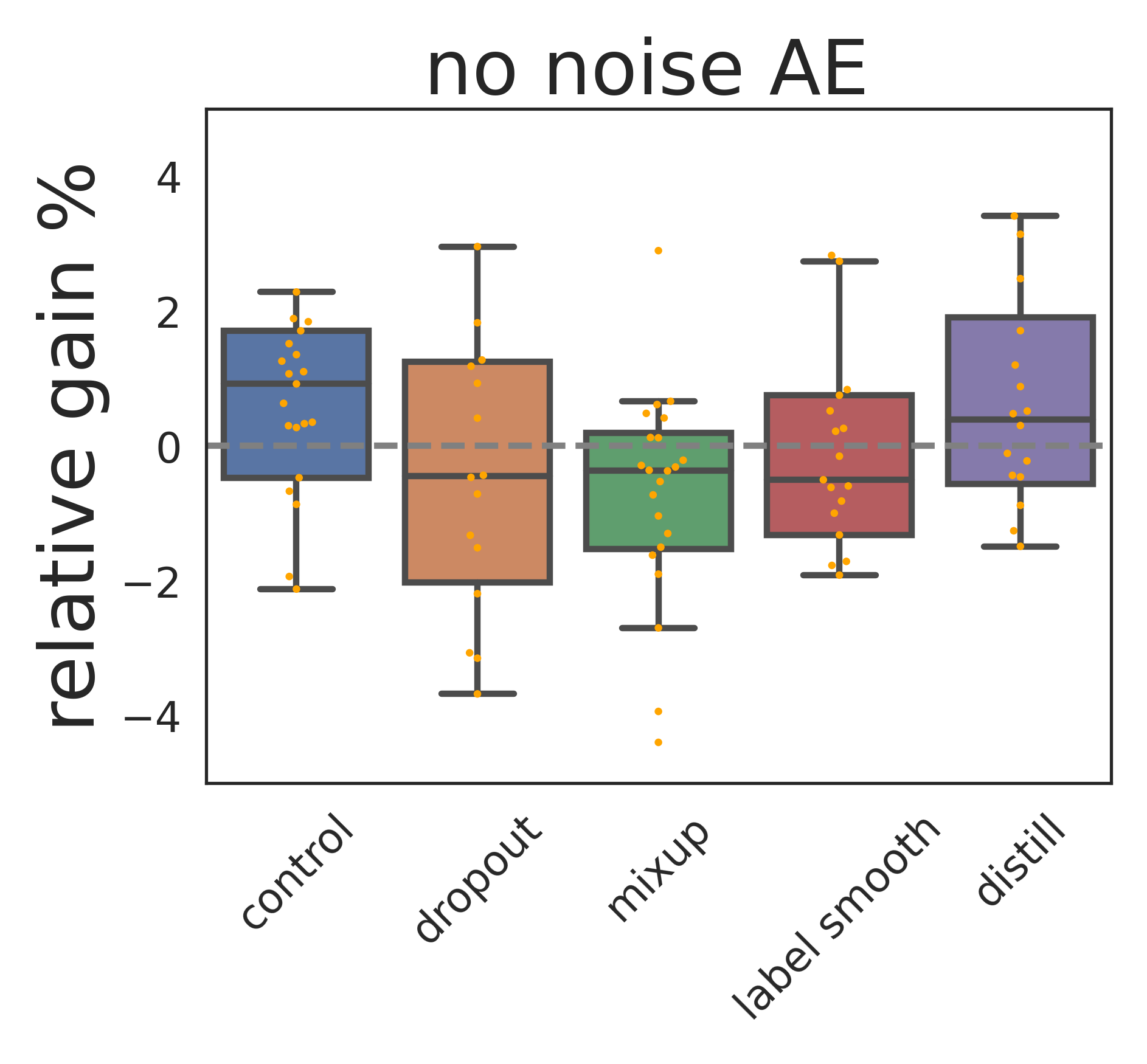}
    \includegraphics[width=0.32\textwidth]{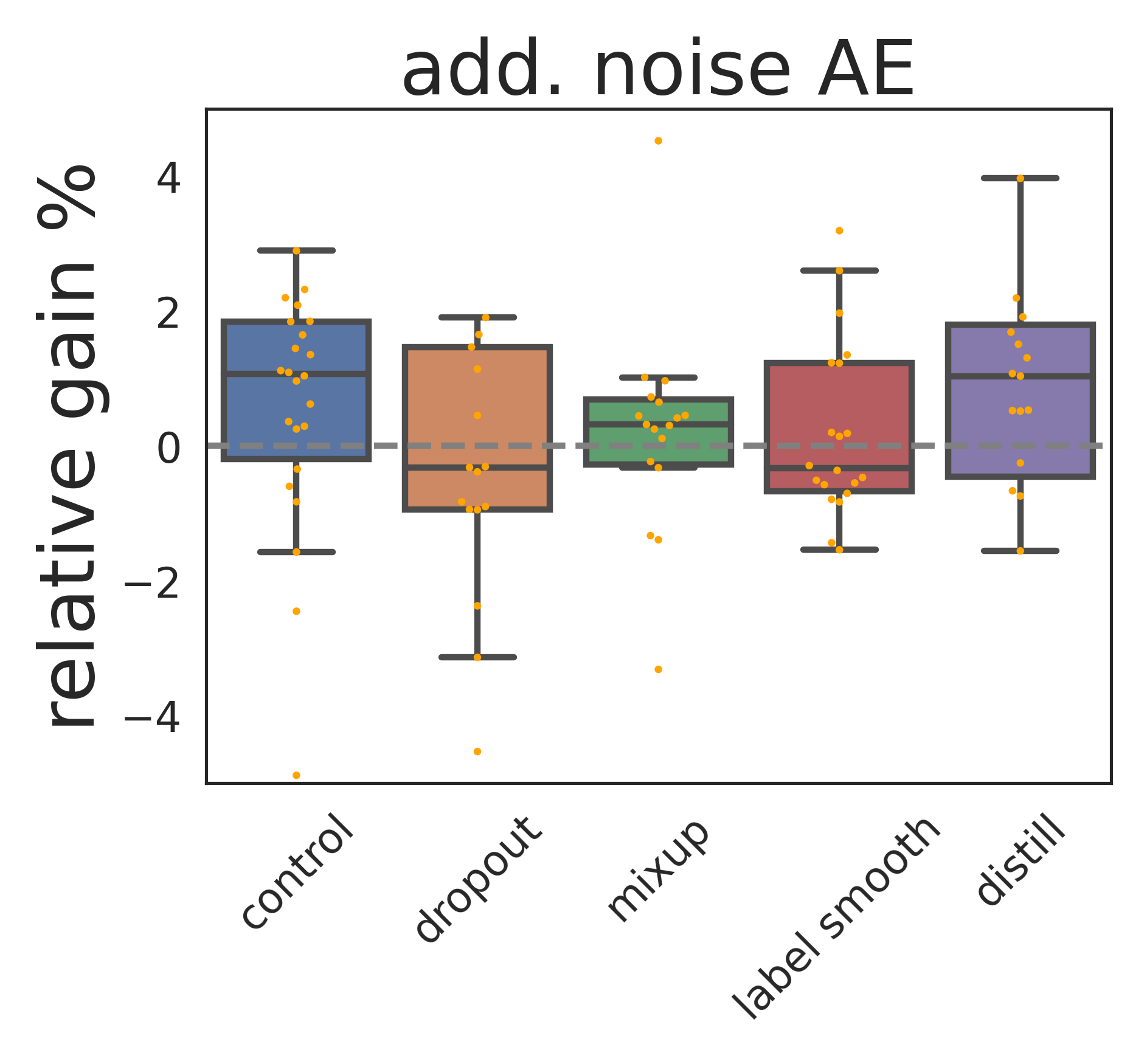}    
    \includegraphics[width=0.32\textwidth]{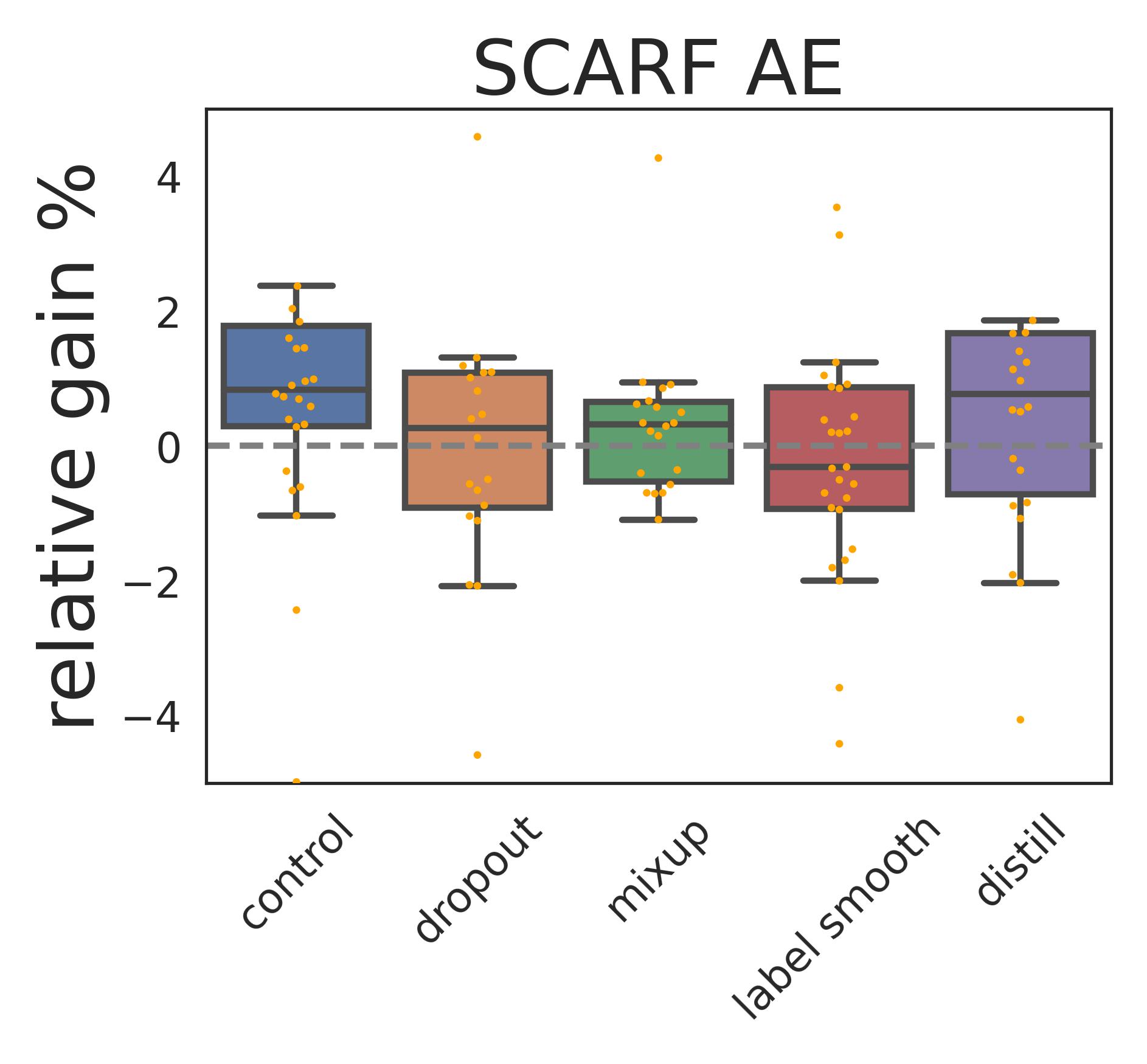}
    \caption{\textbf{Top}: Win matrices comparing pre-training methods against each other, and their improvement to existing solutions. \textbf{Bottom}: Box plots showing the relative improvement of different pre-training methods over baselines (y-axis is zoomed in). We see that \textsc{Scarf} pre-training adds value \emph{even} when used in conjunction with known techniques.} 
    \label{fig:acc}
\end{figure}

{\bf Evaluation methods}. We use the following to effectively convey the results across all datasets.

\noindent\emph{Win matrix}. Given $M$ methods, we compute a ``win'' matrix $W$ of size $M \times M$, where the $(i,j)$ entry is defined as:
\begin{align*}
    W_{i,j} = \frac{\sum_{d=1}^{69} \one[\text{method } i \text{ beats } j \text{ on dataset } d]}{\sum_{d=1}^{69}  \one[\text{method } i \text{ beats } j \text{ on dataset } d] + \one[\text{method } i \text{ loses to } j \text{ on dataset } d]}.
\end{align*}
 ``Beats'' and ``loses'' are only defined when the means are not a statistical tie (using Welch's $t$-test with unequal variance and a $p$-value of $0.05$). A win ratio of $0/1$ means that out of the 69 (pairwise) comparisons, only one was significant and it was a loss. Since $0/1$ and $0/69$ have the same value but the latter is more confident indication that $i$ is worse than $j$, we present the values in fractional form and use a heat map. We add an additional column to the matrix that represents the minimum win ratio across each row.
 
\noindent\emph{Box plots}. The win matrix effectively conveys how often one method beats another but does not capture the degree by which. To that end, for each method, we compute the relative percent improvement over some reference method on each dataset. We then build box-plots depicting the distribution of the relative improvement across datasets, plotting the observations as small points. We only consider datasets where the means of the method and the reference are different with $p$-value $0.20$. We use a larger $p$-value here than when computing the win ratio because otherwise some methods would not have enough points to make the box-plots meaningful.

{\bf Baselines.} We use the following baselines:
\begin{itemize}
    \item \emph{Label smoothing}~\citep{szegedy2016rethinking}, which has proved successful for accuracy~\citep{muller2019does} and label noise~\citep{lukasik2020does}. We use a weight of $0.1$ on the smoothing term. 
    \item \emph{Dropout}. We use standard dropout~\citep{srivastava2014dropout} using rate $0.04$ on all layers. Dropout has been shown to improve performance and robustness to label noise~\citep{rusiecki2020standard}.
    \item \emph{Mixup}~\citep{zhang2017mixup}, using $\alpha = 0.2$.
    \item \emph{Autoencoders}~\citep{rumelhart1985learning}. We use this as our key ablative pre-training baseline. We use the classical autoencoder (``no noise AE''), the denoising autoencoder~\citep{vincent2008extracting,vincent2010stacked} using Gaussian additive noise (``add. noise AE'') as well as \textsc{Scarf}'s corruption method (``\textsc{Scarf} AE''). We use MSE for the reconstruction loss. We try both pre-training and co-training with the supervised task, and when co-training, we add $0.1$ times the autoencoder loss to the supervised objective. We discuss co-training in the Appendix as it is less effective than pre-training.
    \item \emph{\textsc{Scarf} data-augmentation}. In order to isolate the effect of our proposed feature corruption technique, we skip pre-training and instead train on the corrupted inputs during supervised fine-tuning. We discuss results for this baseline in the Appendix as it is less effective than the others.
    \item \emph{Discriminative \textsc{Scarf}}. Here, our pre-training objective is to discriminate between original input features and their counterparts that have been corrupted using our proposed technique. To this end, we update our pre-training head network to include a final linear projection and swap the InfoNCE with a binary logistic loss. We use classification error, not logistic loss, as the validation metric for early stopping, as we found it to perform slightly better.
    \item \emph{Self-distillation}~\citep{hinton2015distilling,zhang2019your}. We first train the model on the labeled data and then train the final model on both the labeled and unlabeled data using the first models' predictions as soft labels for both sets.
    \item \emph{Deep $k$-NN}~\citep{bahri2020deep}, a recently proposed method for label noise. We set $k = 50$.
    \item \emph{Bi-tempered loss}~\citep{amid2019robust}, a recently proposed method for label noise. We use $5$ iterations, $t_1= 0.8$, and $t_2= 1.2$.
    \item \emph{Self-training}~\citep{yarowsky1995unsupervised,mcclosky2006effective}. A classical semi-supervised method -- each iteration, we train on pseudo-labeled data (initialized to be the original labeled dataset) and add highly confident predictions to the training set using the prediction as the label. We then train our final model on the final dataset. We use a softmax prediction threshold of $0.75$ and run for $10$ iterations.
    \item \emph{Tri-training}~\citep{zhou2005tri}. Like self-training, but using three models with different initial labeled data via bootstrap sampling. Each iteration, every model's training set is updated by adding only unlabeled points whose predictions made by the other two models agree. It was shown to be competitive in modern semi-supervised NLP tasks~\citep{ruder2018strong}. We use same hyperparameters as self-training.
\end{itemize}

\subsection{\textsc{Scarf} pre-training improves predictive performance}
Figure~\ref{fig:acc} shows our results. From the first win matrix plot, we see that all five pre-training techniques considered improve over no pre-training (control), and that \textsc{Scarf} outperforms the others and has more statistically significant wins. The second win matrix shows that \textsc{Scarf} pre-training boosts the performance of mixup, label smoothing, distillation, and dropout, and it does so better than alternatives. In other words, pre-training \emph{complements} existing solutions, suggesting that a distinct mechanism is at play here.
The box plots expand on the second win matrix, showing the relative improvement that each of the pre-training strategies confers over the baselines. Table~\ref{tab:relgain} summarizes the box-plots by the average relative gain. We observe that \textsc{Scarf} generally outperforms the alternatives and adds a 1-2\% relative gain across the board.

\begin{figure}[!t]
    \centering
    \includegraphics[width=0.49\textwidth]{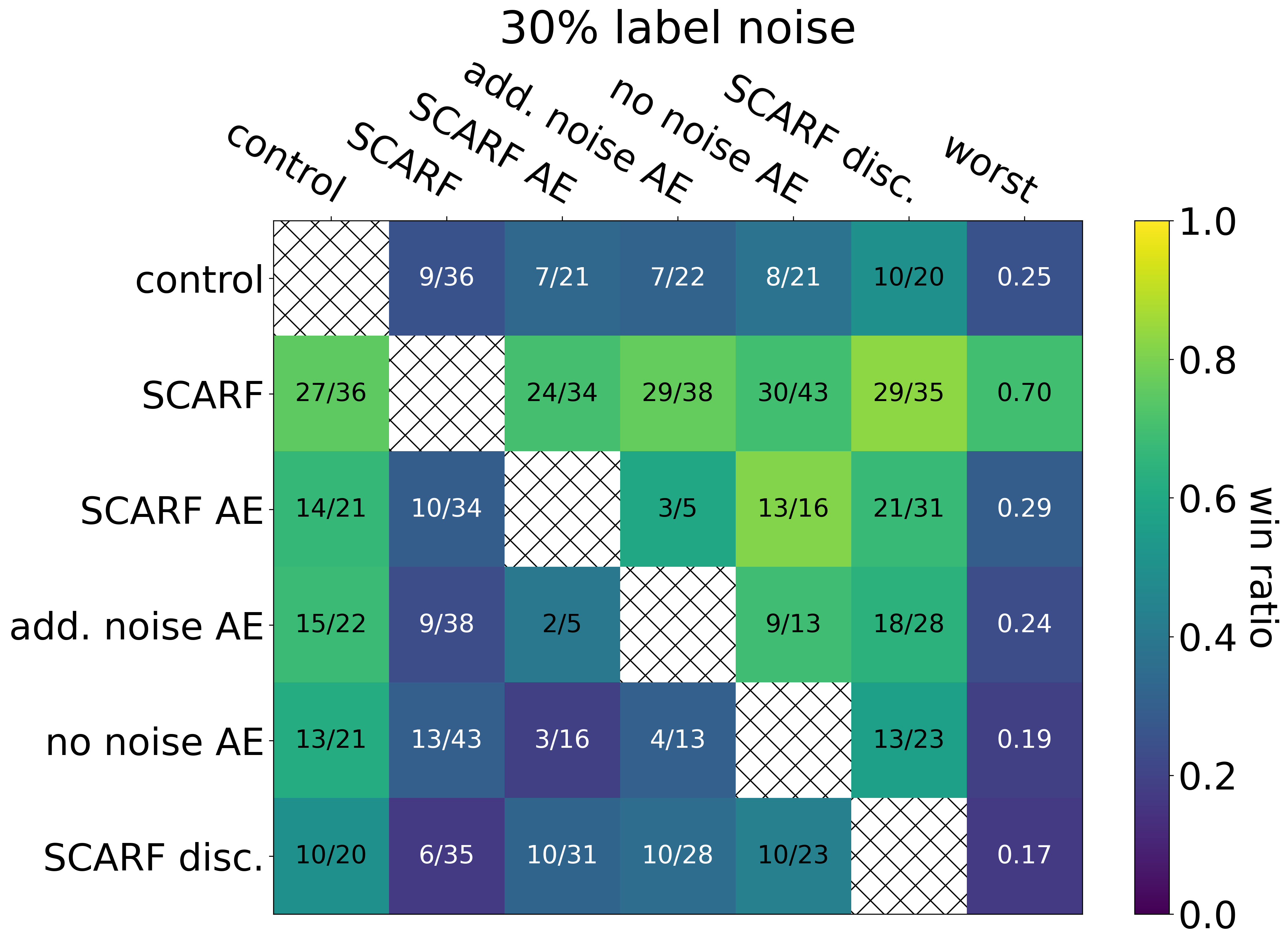}
    \includegraphics[width=0.49\textwidth]{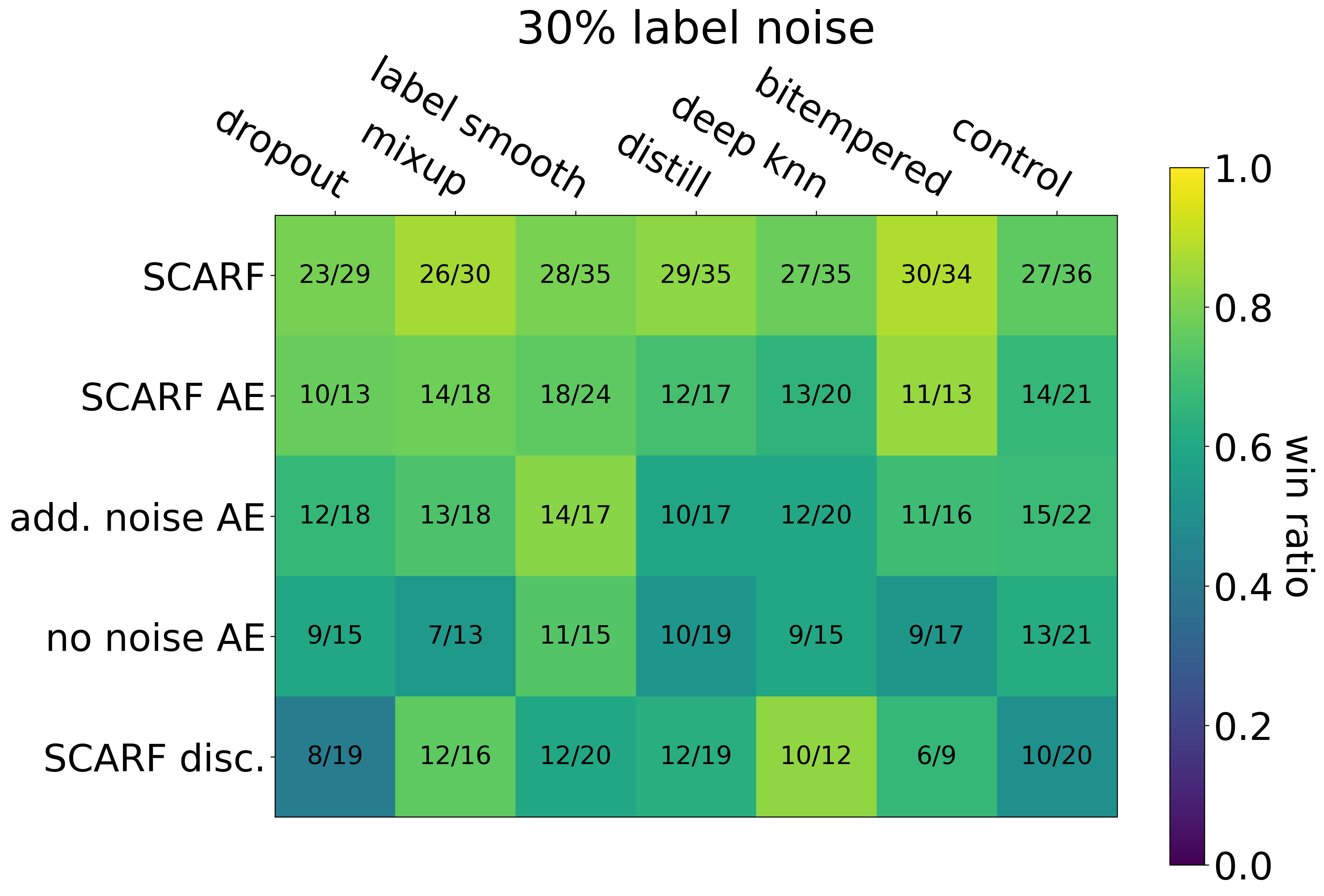}\\
    \includegraphics[width=0.32\textwidth]{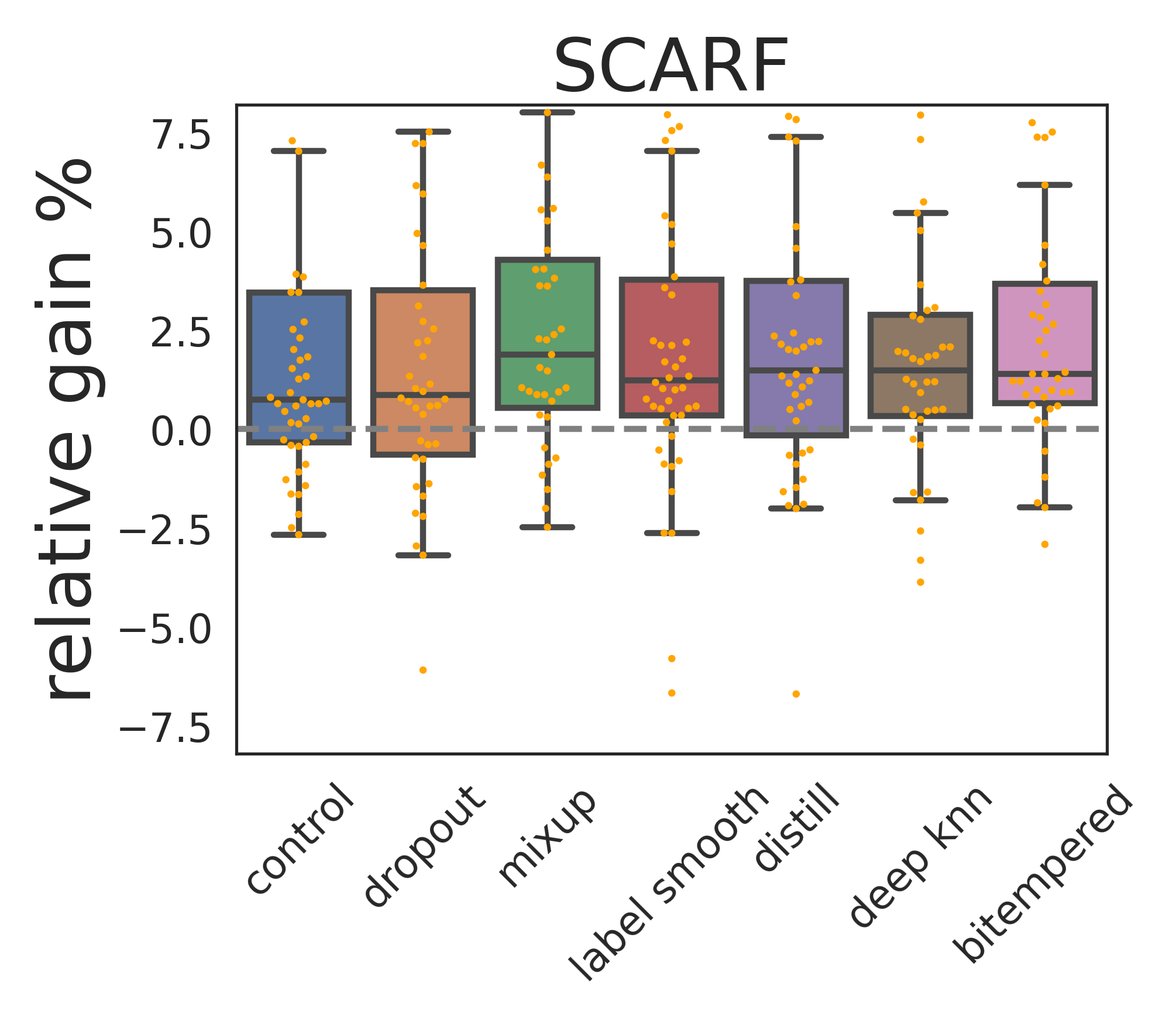} \includegraphics[width=0.32\textwidth]{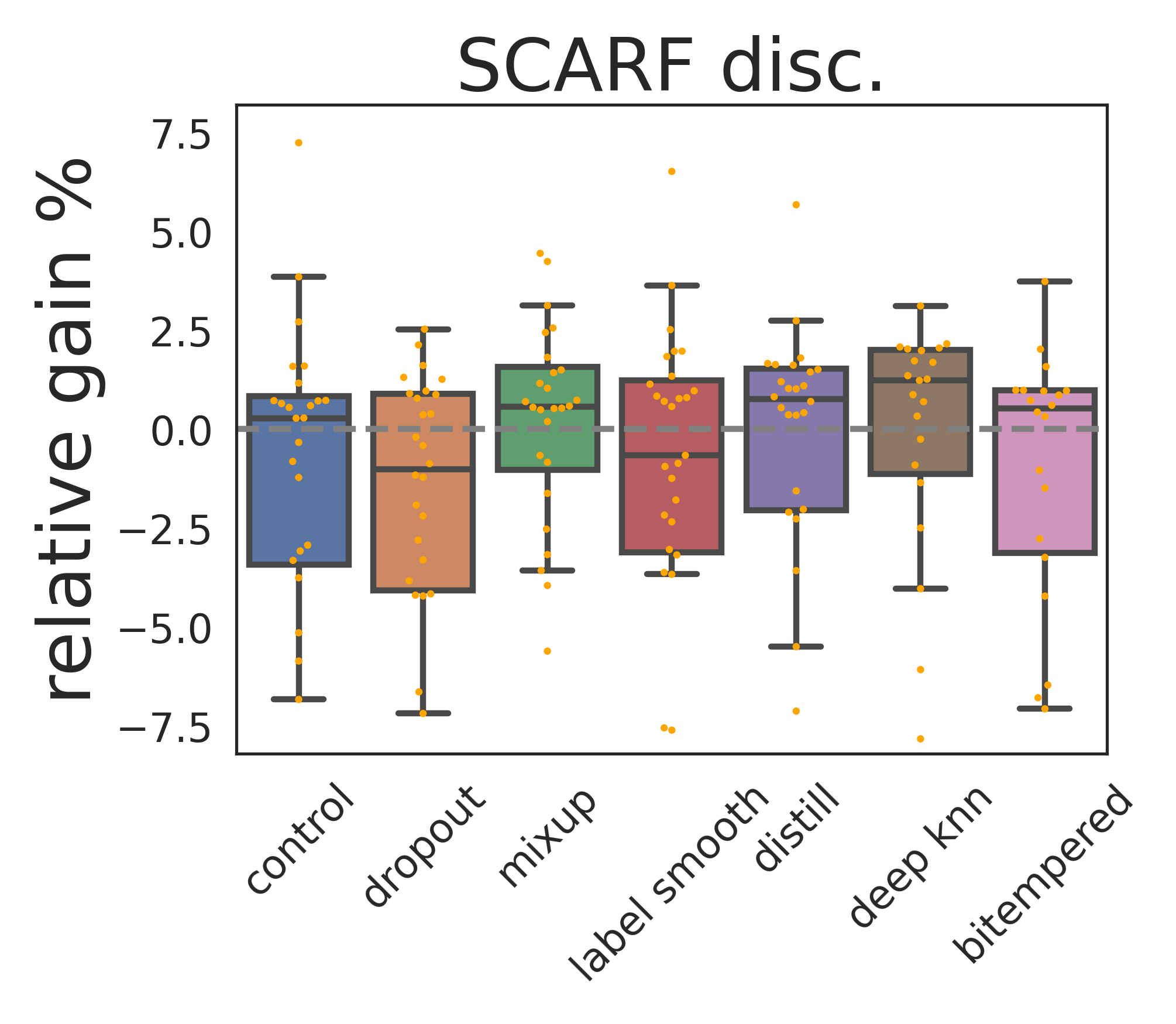}\\
    \includegraphics[width=0.32\textwidth]{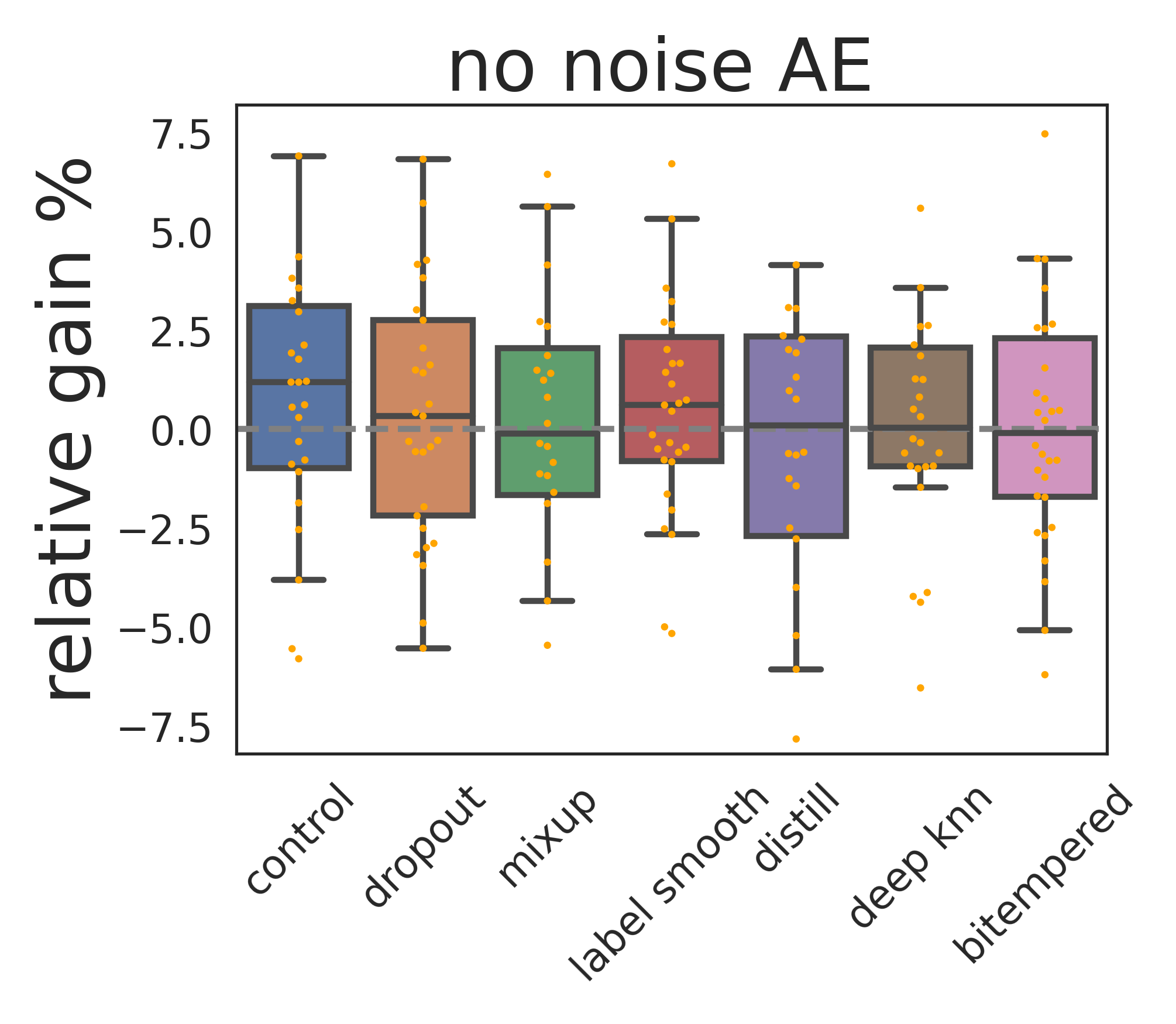}
    \includegraphics[width=0.32\textwidth]{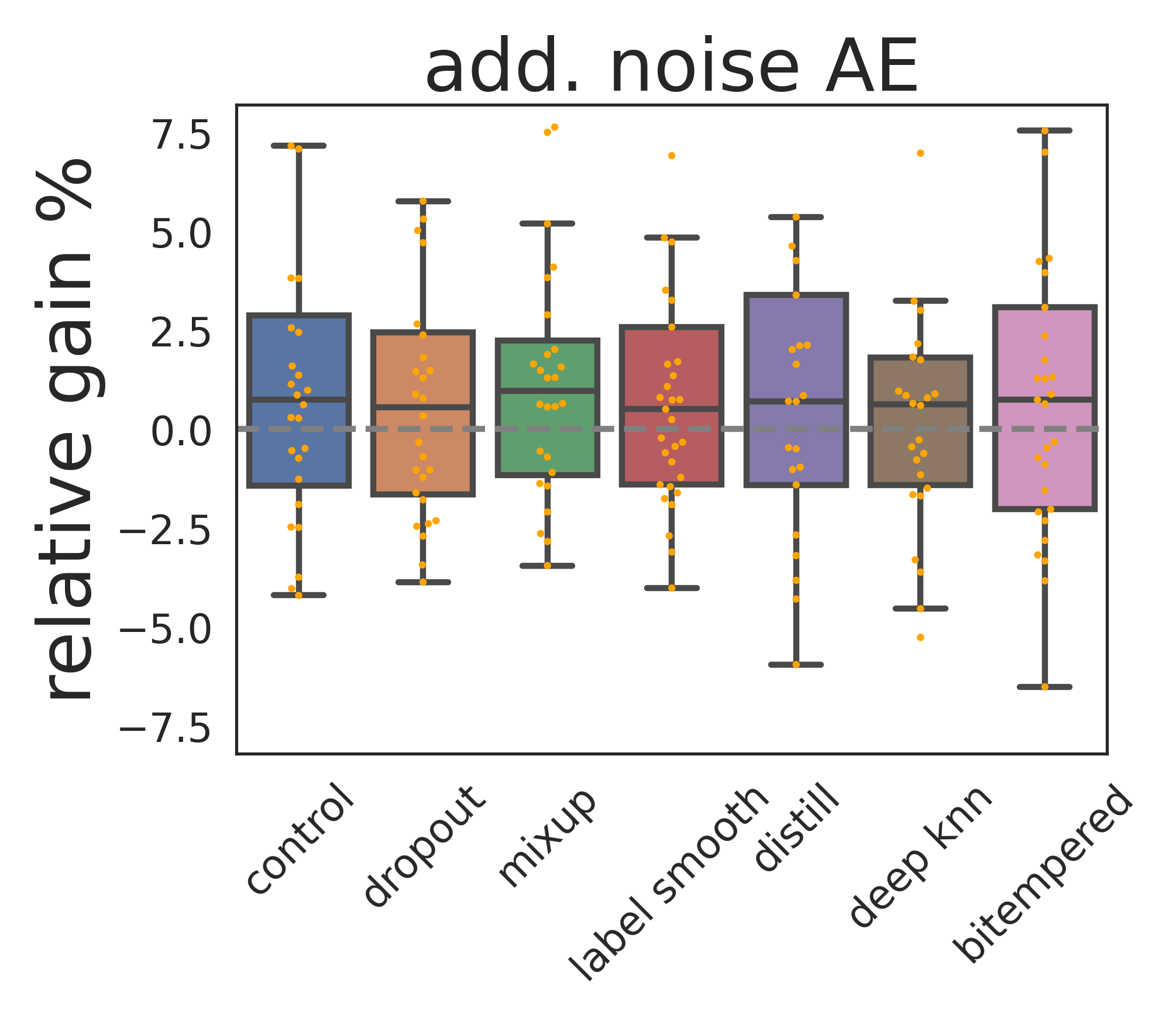}    
    \includegraphics[width=0.32\textwidth]{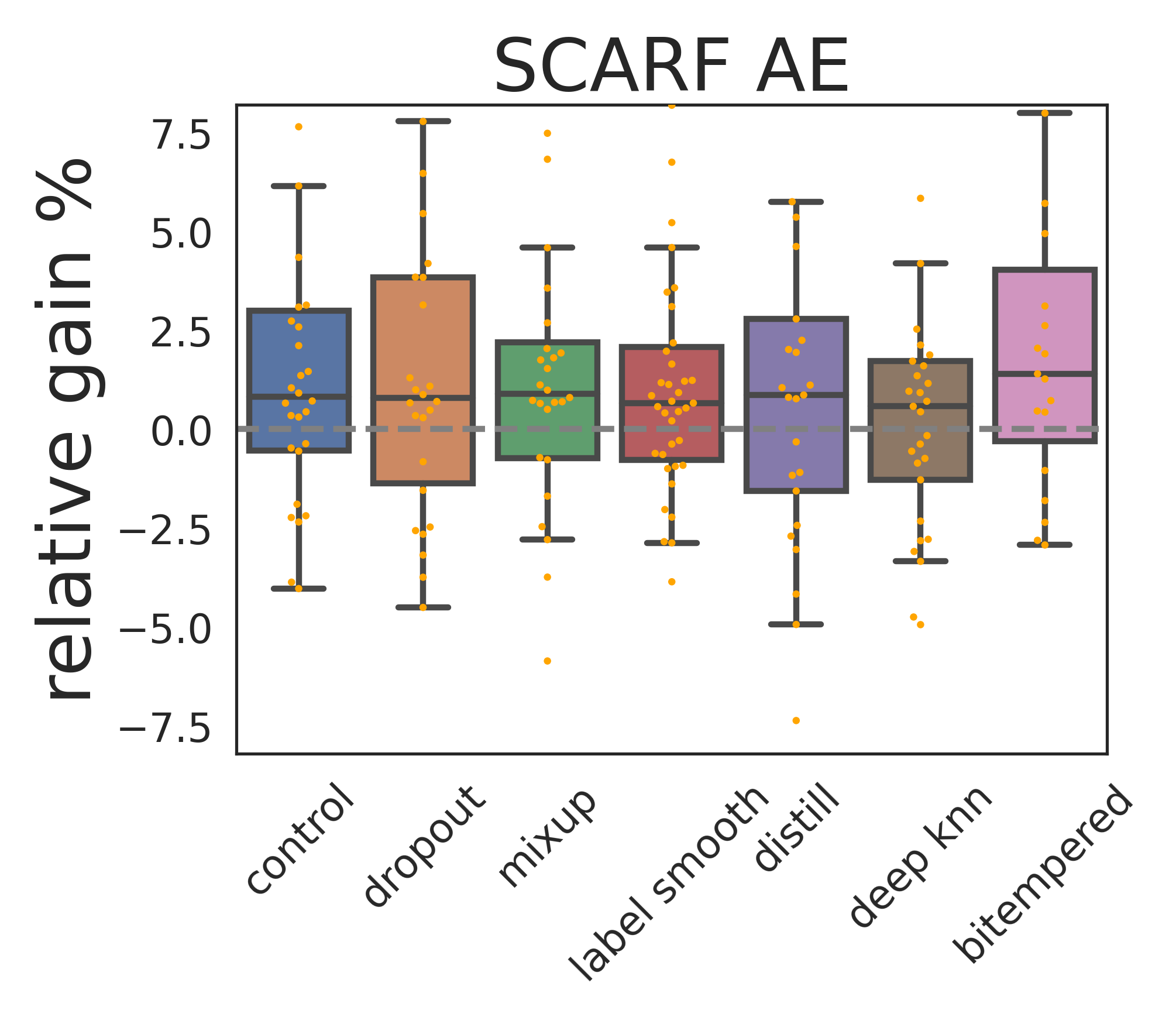}
    \caption{\textsc{Scarf} boosts baseline performance even when $30\%$ of the training labels are corrupted. Notably, it improves state-of-the-art label noise solutions like Deep $k$-NN.}
    \label{fig:noise30}
\end{figure}

\subsection{\textsc{Scarf} pre-training improves performance in the presence to label noise}
To show how pre-training improves model robustness when the data's labels are unreliable, we do as follows.
Firstly, label noise robustness is often studied in two distinct settings - (1) when some subset of the training data is guaranteed to be uncorrupted and that set is known in advance, (2) when the entire dataset is untrustworthy. For simplicity, we consider setting 2 in our experiments. We corrupt labels as follows:
leaving the validation and test splits uncorrupted, we select a random 30\% percent of the training data to corrupt and for each datapoint, we replace its label by one of the classes, uniformly over all classes (including the datapoint's true class). Results are shown in Figure~\ref{fig:noise30} and Table~\ref{tab:relgain}. Again, we see \textsc{Scarf} outperforms the rest and boosts all baselines by 2-3\%.

\begin{table}[!t]
\begin{center}

\begin{tabular}{lrrrrr}
\toprule
\emph{100\% labeled training} & \multicolumn{1}{l}{\textsc{Scarf}} & \multicolumn{1}{l}{\textsc{Scarf} AE} & \multicolumn{1}{l}{no noise AE} & \multicolumn{1}{l}{add. noise AE} & \multicolumn{1}{l}{\textsc{Scarf} disc.} \\
\midrule
control & \textbf{2.352} & 2.244 & 1.107 & 1.559 & 0.574 \\
dropout & \textbf{1.609} & 1.196 & 0.623 & 1.228 & -1.312 \\
mixup & \textbf{1.72} & 1.183 & -0.377 & 0.971 & -0.307 \\
label smooth & \textbf{1.522} & 0.711 & -0.002 & 1.04 & -0.894 \\
distill & \textbf{2.392} & 2.186 & 0.823 & 1.431 & -0.394 \\
\midrule
\emph{25\% labeled training} & \multicolumn{1}{l}{} & \multicolumn{1}{l}{} & \multicolumn{1}{l}{} & \multicolumn{1}{l}{} & \multicolumn{1}{l}{} \\
\midrule
control & \textbf{3.692} & 1.702 & 0.777 & 1.662 & 0.233 \\
dropout & \textbf{2.212} & 1.848 & 2.013 & 1.155 & -0.322 \\
mixup & \textbf{2.809} & 0.73 & 0.106 & 0.439 & 0.466 \\
label smooth & \textbf{2.303} & 0.705 & -0.564 & 0.196 & -0.206 \\
distill & \textbf{3.609} & 2.441 & 1.969 & 2.263 & 1.795 \\
self-train & \textbf{3.839} & 2.753 & 1.672 & 2.839 & 2.559 \\
tri-train & \textbf{3.549} & 2.706 & 1.455 & 2.526 & 1.92 \\
\midrule
\emph{30\% label noise} & \multicolumn{1}{l}{} & \multicolumn{1}{l}{} & \multicolumn{1}{l}{} & \multicolumn{1}{l}{} & \multicolumn{1}{l}{} \\
\midrule
control & \textbf{2.261} & 1.988 & 0.914 & 1.612 & -1.408 \\
dropout & 2.004 & \textbf{2.058} & 0.9 & 1.471 & -2.54 \\
mixup & \textbf{2.739} & 1.723 & 0.116 & 1.409 & 0.189 \\
label smooth & \textbf{2.558} & 1.474 & 0.703 & 1.395 & -1.337 \\
distill & \textbf{2.881} & 2.296 & -0.239 & 1.659 & -0.226 \\
deep knn & \textbf{2.001} & 1.281 & 0.814 & 1.348 & 0.088 \\
bitempered & 2.68 & \textbf{2.915} & 0.435 & 1.387 & -1.147 \\
\bottomrule
\end{tabular}

\caption{\label{tab:relgain} Results using the fully labeled training data, only 25\% of the labeled training data, and the full training data subject to 30\% label noise. Shown is the average relative gain in accuracy when adding the pre-training methods (columns) to the reference methods (rows). Like the box-plots, we filter out datasets using $p$-value 0.20. We see that \textsc{Scarf} consistently outperforms alternatives, not only in improving control but also in improving methods designed specifically for the setting.}
\end{center}
\end{table}

\subsection{\textsc{Scarf} pre-training improves performance when labeled data is limited}
To show how pre-training helps when there is more unlabeled data than labeled ones, we remove labels in the training split so that only $25\%$ of the original split remains labeled. Autoencoders, \textsc{Scarf}, self-training and tri-training all leverage the unlabeled remainder. Results are shown in Figure~\ref{fig:se25} (Appendix) and Table~\ref{tab:relgain}. \textsc{Scarf} outperforms the rest, adding a very impressive 2-4\% to all baselines.

\subsection{Ablations}
We now detail the importance of every factor in \textsc{Scarf}. We show only some of the results here; the rest are in the Appendix.

{\bf Other corruption strategies are less effective and are more sensitive to feature scaling.}
Here, we ablate the marginal sampling corruption technique we proposed, replacing it with the following other promising strategies, while keeping all else fixed.
\begin{enumerate}
\item \emph{No corruption.} We do not apply any corruption - i.e. $\tilde{x}^{(i)} = x^{(i)}$, in Algorithm~\ref{alg:main}. In this case, the cosine similarity between positive pairs is always one and the model is learning to make negative pairs as orthogonal as possible. Under the recent perspective~\citep{wang2020understanding} that the contrastive loss comprises two terms -- one that encourages alignment between views of the same example -- and one that encourages the hypersphere embeddings to be uniformly spread out -- we see that with no corruption, pre-training may just be learning to embed input examples uniformly on the hypersphere.
\item \emph{Mean corruption.} After determining which features to corrupt, we replace their entries with the empirical marginal distribution's mean.
\item \emph{Additive Gaussian noise}. We add i.i.d $\mathcal{N}(0, 0.5^2)$ noise to features.
\item \emph{Joint sampling.} Rather than replacing features by random draws from their marginals to form $\tilde{x}^{(i)}$, we instead randomly draw $\hat{x}^{(i)}$ from training data $\mathcal{X}$ -- i.e. we draw from the empirical (joint) data distribution -- and then set $\tilde{x}^{(i)}_j = \hat{x}_j^{(i)} \;\forall j \in \mathcal{I}_i$.
\item \emph{Missing feature corruption.} We mark the selected features as ``missing'' and add one learnable value per feature dimension to our model. When a feature is missing, it's filled in with the corresponding learnable missing value. 
\item \emph{Feature dropout.} We zero-out the selected features.
\end{enumerate}
We also examine the corruption strategies under the following ways of scaling the input features.
\begin{enumerate}
\item \emph{Z-score scaling.} Here, $x_j = \left[x_j-\mathrm{mean}\left({\mathcal{X}_j}\right)\right] / \mathrm{std}\left({\mathcal{X}_j}\right)$.
\item \emph{Min-max scaling.} $x_j = \left[x_j-\min\left({\mathcal{X}_j}\right)\right] / \left[\max\left({\mathcal{X}_j}\right)-\min\left({\mathcal{X}_j}\right)\right]$.
\item \emph{Mean scaling.} $x_j = \left[x-\mathrm{mean}\left({\mathcal{X}_j}\right)\right] / \left[\max\left({\mathcal{X}_j}\right)-\min\left({\mathcal{X}_j}\right)\right]$.
\end{enumerate}
Figure~\ref{fig:corruption_ablation} (Appendix) shows the results for z-score and min-max scaling. \textsc{Scarf} marginal sampling generally outperforms the other corruption strategies for different types of feature scaling. Marginal sampling is neat in that in addition to not having hyperparameters, it is invariant to scaling and preserves the ``units'' of each feature. In contrast, even a simple multiplicative scaling requires the additive noise to be scaled in the same way. 

{\bf \textsc{Scarf} is \emph{not} sensitive to batch size.} Contrastive methods like SimCLR~\citep{chen2020simple} have shown consistent improvements upon increasing the batch size, $N$. There is a tight coupling between the batch size and how hard the contrastive learning task is, since, in our case, the loss term for each example $i$ involves 1 positive and $N-1$ negatives.
The need for large (e.g. $5000$) batch sizes has motivated engineering solutions to support them~\citep{gao2021scaling} and have been seen as grounds for adopting other loss functions~\citep{zbontar2021barlow}. Figure~\ref{fig:rate_ablation} (Appendix) compares a range of batch sizes. We see that increasing the batch size past 128 did not result in significant improvements. A reasonable hypothesis here is that higher capacity models and harder tasks benefit more from negatives.

\noindent \paragraph{\textsc{Scarf} is fairly insensitive to the corruption rate and temperature.}
We study the effect of the corruption rate $c$ in Figure~\ref{fig:rate_ablation} (Appendix). We see that performance is stable when the rate is in the range $50\% - 80\%$. We thus recommend a default setting of $60\%$. We see a similar stability with respect to the temperature hyperparameter (see Appendix). We recommend using a default temperature of 1.

\noindent \paragraph{Tweaks to the corruption do not work any better.}
The Appendix shows the effect of four distinct tweaks to the corruption method. We do not see any reason to use any of them.

\noindent \paragraph{Alternatives to InfoNCE do not work any better.}
We investigate the importance of our choice of InfoNCE loss function and see the effects of swapping it with recently proposed alternatives Alignment and Uniformity~\citep{wang2020understanding} and Barlow Twins~\citep{zbontar2021barlow}. We found that these alternative losses almost match or perform worse than the original and popular InfoNCE in our setting. See the Appendix for details.

\section{Conclusion}

Self-supervised learning has seen profound success in important domains including computer vision and natural language processing, but little progress has been made for the general tabular setting. We propose a self-supervised learning method that's simple and versatile and learns representations that are effective in downstream classification tasks, even in the presence of limited labeled data or label noise. Potential negative side effects of this method may include learning representations that reinforce biases that appear in the input data. Finding ways to mitigate this during the training process is a potential direction for future research.

\bibliography{refs}

\begin{thebibliography}{85}
\providecommand{\natexlab}[1]{#1}
\providecommand{\url}[1]{\texttt{#1}}
\expandafter\ifx\csname urlstyle\endcsname\relax
  \providecommand{\doi}[1]{doi: #1}\else
  \providecommand{\doi}{doi: \begingroup \urlstyle{rm}\Url}\fi

\bibitem[Amid et~al.(2019)Amid, Warmuth, Anil, and Koren]{amid2019robust}
Ehsan Amid, Manfred~K Warmuth, Rohan Anil, and Tomer Koren.
\newblock Robust bi-tempered logistic loss based on bregman divergences.
\newblock \emph{arXiv preprint arXiv:1906.03361}, 2019.

\bibitem[Bahri et~al.(2020)Bahri, Jiang, and Gupta]{bahri2020deep}
Dara Bahri, Heinrich Jiang, and Maya Gupta.
\newblock Deep k-nn for noisy labels.
\newblock In \emph{International Conference on Machine Learning}, pp.\
  540--550. PMLR, 2020.

\bibitem[Bao et~al.(2020)Bao, Dong, Wei, Wang, Yang, Liu, Wang, Gao, Piao,
  Zhou, et~al.]{bao2020unilmv2}
Hangbo Bao, Li~Dong, Furu Wei, Wenhui Wang, Nan Yang, Xiaodong Liu, Yu~Wang,
  Jianfeng Gao, Songhao Piao, Ming Zhou, et~al.
\newblock Unilmv2: Pseudo-masked language models for unified language model
  pre-training.
\newblock In \emph{International Conference on Machine Learning}, pp.\
  642--652. PMLR, 2020.

\bibitem[Bischl et~al.(2017)Bischl, Casalicchio, Feurer, Hutter, Lang,
  Mantovani, van Rijn, and Vanschoren]{bischl2017openml}
Bernd Bischl, Giuseppe Casalicchio, Matthias Feurer, Frank Hutter, Michel Lang,
  Rafael~G Mantovani, Jan~N van Rijn, and Joaquin Vanschoren.
\newblock Openml benchmarking suites.
\newblock \emph{arXiv preprint arXiv:1708.03731}, 2017.

\bibitem[Bojanowski \& Joulin(2017)Bojanowski and
  Joulin]{bojanowski2017unsupervised}
Piotr Bojanowski and Armand Joulin.
\newblock Unsupervised learning by predicting noise.
\newblock In \emph{International Conference on Machine Learning}, pp.\
  517--526. PMLR, 2017.

\bibitem[Brown et~al.(2020)Brown, Mann, Ryder, Subbiah, Kaplan, Dhariwal,
  Neelakantan, Shyam, Sastry, Askell, et~al.]{brown2020language}
Tom~B Brown, Benjamin Mann, Nick Ryder, Melanie Subbiah, Jared Kaplan, Prafulla
  Dhariwal, Arvind Neelakantan, Pranav Shyam, Girish Sastry, Amanda Askell,
  et~al.
\newblock Language models are few-shot learners.
\newblock \emph{arXiv preprint arXiv:2005.14165}, 2020.

\bibitem[Chen et~al.(2018)Chen, Zhai, and Houlsby]{chen2018self}
Ting Chen, Xiaohua Zhai, and Neil Houlsby.
\newblock Self-supervised gan to counter forgetting.
\newblock \emph{arXiv preprint arXiv:1810.11598}, 2018.

\bibitem[Chen et~al.(2020)Chen, Kornblith, Norouzi, and Hinton]{chen2020simple}
Ting Chen, Simon Kornblith, Mohammad Norouzi, and Geoffrey Hinton.
\newblock A simple framework for contrastive learning of visual
  representations.
\newblock In \emph{International conference on machine learning}, pp.\
  1597--1607. PMLR, 2020.

\bibitem[Clark et~al.(2020)Clark, Luong, Le, and Manning]{clark2020electra}
Kevin Clark, Minh-Thang Luong, Quoc~V Le, and Christopher~D Manning.
\newblock Electra: Pre-training text encoders as discriminators rather than
  generators.
\newblock \emph{arXiv preprint arXiv:2003.10555}, 2020.

\bibitem[Collobert et~al.(2011)Collobert, Weston, Bottou, Karlen, Kavukcuoglu,
  and Kuksa]{collobert2011natural}
Ronan Collobert, Jason Weston, L{\'e}on Bottou, Michael Karlen, Koray
  Kavukcuoglu, and Pavel Kuksa.
\newblock Natural language processing (almost) from scratch.
\newblock \emph{Journal of machine learning research}, 12\penalty0
  (ARTICLE):\penalty0 2493--2537, 2011.

\bibitem[Cubuk et~al.(2020)Cubuk, Zoph, Shlens, and Le]{cubuk2020randaugment}
Ekin~D Cubuk, Barret Zoph, Jonathon Shlens, and Quoc~V Le.
\newblock Randaugment: Practical automated data augmentation with a reduced
  search space.
\newblock In \emph{Proceedings of the IEEE/CVF Conference on Computer Vision
  and Pattern Recognition Workshops}, pp.\  702--703, 2020.

\bibitem[de~Vries et~al.(2019)de~Vries, van Cranenburgh, Bisazza, Caselli, van
  Noord, and Nissim]{de2019bertje}
Wietse de~Vries, Andreas van Cranenburgh, Arianna Bisazza, Tommaso Caselli,
  Gertjan van Noord, and Malvina Nissim.
\newblock Bertje: A dutch bert model.
\newblock \emph{arXiv preprint arXiv:1912.09582}, 2019.

\bibitem[Devlin et~al.(2018)Devlin, Chang, Lee, and Toutanova]{devlin2018bert}
Jacob Devlin, Ming-Wei Chang, Kenton Lee, and Kristina Toutanova.
\newblock Bert: Pre-training of deep bidirectional transformers for language
  understanding.
\newblock \emph{arXiv preprint arXiv:1810.04805}, 2018.

\bibitem[Donahue et~al.(2016)Donahue, Kr{\"a}henb{\"u}hl, and
  Darrell]{donahue2016adversarial}
Jeff Donahue, Philipp Kr{\"a}henb{\"u}hl, and Trevor Darrell.
\newblock Adversarial feature learning.
\newblock \emph{arXiv preprint arXiv:1605.09782}, 2016.

\bibitem[Dong et~al.(2019)Dong, Yang, Wang, Wei, Liu, Wang, Gao, Zhou, and
  Hon]{dong2019unified}
Li~Dong, Nan Yang, Wenhui Wang, Furu Wei, Xiaodong Liu, Yu~Wang, Jianfeng Gao,
  Ming Zhou, and Hsiao-Wuen Hon.
\newblock Unified language model pre-training for natural language
  understanding and generation.
\newblock \emph{arXiv preprint arXiv:1905.03197}, 2019.

\bibitem[Feurer et~al.(2019)Feurer, van Rijn, Kadra, Gijsbers, Mallik, Ravi,
  Mueller, Vanschoren, and Hutter]{OpenMLPython2019}
Matthias Feurer, Jan~N. van Rijn, Arlind Kadra, Pieter Gijsbers, Neeratyoy
  Mallik, Sahithya Ravi, Andreas Mueller, Joaquin Vanschoren, and Frank Hutter.
\newblock Openml-python: an extensible python api for openml.
\newblock \emph{arXiv}, 1911.02490, 2019.
\newblock URL \url{https://arxiv.org/pdf/1911.02490.pdf}.

\bibitem[Freitag \& Roy(2018)Freitag and Roy]{freitag2018unsupervised}
Markus Freitag and Scott Roy.
\newblock Unsupervised natural language generation with denoising autoencoders.
\newblock \emph{arXiv preprint arXiv:1804.07899}, 2018.

\bibitem[Gao \& Zhang(2021)Gao and Zhang]{gao2021scaling}
Luyu Gao and Yunyi Zhang.
\newblock Scaling deep contrastive learning batch size with almost constant
  peak memory usage.
\newblock \emph{arXiv preprint arXiv:2101.06983}, 2021.

\bibitem[Gidaris et~al.(2018)Gidaris, Singh, and
  Komodakis]{gidaris2018unsupervised}
Spyros Gidaris, Praveer Singh, and Nikos Komodakis.
\newblock Unsupervised representation learning by predicting image rotations.
\newblock \emph{arXiv preprint arXiv:1803.07728}, 2018.

\bibitem[Gontijo-Lopes et~al.(2020)Gontijo-Lopes, Smullin, Cubuk, and
  Dyer]{gontijo2020affinity}
Raphael Gontijo-Lopes, Sylvia~J Smullin, Ekin~D Cubuk, and Ethan Dyer.
\newblock Affinity and diversity: Quantifying mechanisms of data augmentation.
\newblock \emph{arXiv preprint arXiv:2002.08973}, 2020.

\bibitem[Goodfellow et~al.(2014)Goodfellow, Pouget-Abadie, Mirza, Xu,
  Warde-Farley, Ozair, Courville, and Bengio]{goodfellow2014generative}
Ian~J Goodfellow, Jean Pouget-Abadie, Mehdi Mirza, Bing Xu, David Warde-Farley,
  Sherjil Ozair, Aaron Courville, and Yoshua Bengio.
\newblock Generative adversarial networks.
\newblock \emph{arXiv preprint arXiv:1406.2661}, 2014.

\bibitem[Grill et~al.(2020)Grill, Strub, Altch{\'e}, Tallec, Richemond,
  Buchatskaya, Doersch, Pires, Guo, Azar, et~al.]{grill2020bootstrap}
Jean-Bastien Grill, Florian Strub, Florent Altch{\'e}, Corentin Tallec,
  Pierre~H Richemond, Elena Buchatskaya, Carl Doersch, Bernardo~Avila Pires,
  Zhaohan~Daniel Guo, Mohammad~Gheshlaghi Azar, et~al.
\newblock Bootstrap your own latent: A new approach to self-supervised
  learning.
\newblock \emph{arXiv preprint arXiv:2006.07733}, 2020.

\bibitem[Gutmann \& Hyv{\"a}rinen(2010)Gutmann and
  Hyv{\"a}rinen]{gutmann2010noise}
Michael Gutmann and Aapo Hyv{\"a}rinen.
\newblock Noise-contrastive estimation: A new estimation principle for
  unnormalized statistical models.
\newblock In \emph{Proceedings of the Thirteenth International Conference on
  Artificial Intelligence and Statistics}, pp.\  297--304. JMLR Workshop and
  Conference Proceedings, 2010.

\bibitem[Hassani \& Khasahmadi(2020)Hassani and
  Khasahmadi]{hassani2020contrastive}
Kaveh Hassani and Amir~Hosein Khasahmadi.
\newblock Contrastive multi-view representation learning on graphs.
\newblock In \emph{International Conference on Machine Learning}, pp.\
  4116--4126. PMLR, 2020.

\bibitem[He et~al.(2020)He, Fan, Wu, Xie, and Girshick]{he2020momentum}
Kaiming He, Haoqi Fan, Yuxin Wu, Saining Xie, and Ross Girshick.
\newblock Momentum contrast for unsupervised visual representation learning.
\newblock In \emph{Proceedings of the IEEE/CVF Conference on Computer Vision
  and Pattern Recognition}, pp.\  9729--9738, 2020.

\bibitem[Henaff(2020)]{henaff2020data}
Olivier Henaff.
\newblock Data-efficient image recognition with contrastive predictive coding.
\newblock In \emph{International Conference on Machine Learning}, pp.\
  4182--4192. PMLR, 2020.

\bibitem[Hinton et~al.(2015)Hinton, Vinyals, and Dean]{hinton2015distilling}
Geoffrey Hinton, Oriol Vinyals, and Jeff Dean.
\newblock Distilling the knowledge in a neural network.
\newblock \emph{arXiv preprint arXiv:1503.02531}, 2015.

\bibitem[Hjelm et~al.(2018)Hjelm, Fedorov, Lavoie-Marchildon, Grewal, Bachman,
  Trischler, and Bengio]{hjelm2018learning}
R~Devon Hjelm, Alex Fedorov, Samuel Lavoie-Marchildon, Karan Grewal, Phil
  Bachman, Adam Trischler, and Yoshua Bengio.
\newblock Learning deep representations by mutual information estimation and
  maximization.
\newblock \emph{arXiv preprint arXiv:1808.06670}, 2018.

\bibitem[Ho et~al.(2019)Ho, Liang, Chen, Stoica, and Abbeel]{ho2019population}
Daniel Ho, Eric Liang, Xi~Chen, Ion Stoica, and Pieter Abbeel.
\newblock Population based augmentation: Efficient learning of augmentation
  policy schedules.
\newblock In \emph{International Conference on Machine Learning}, pp.\
  2731--2741. PMLR, 2019.

\bibitem[Iizuka et~al.(2017)Iizuka, Simo-Serra, and
  Ishikawa]{iizuka2017globally}
Satoshi Iizuka, Edgar Simo-Serra, and Hiroshi Ishikawa.
\newblock Globally and locally consistent image completion.
\newblock \emph{ACM Transactions on Graphics (ToG)}, 36\penalty0 (4):\penalty0
  1--14, 2017.

\bibitem[Jenni \& Favaro(2018)Jenni and Favaro]{jenni2018self}
Simon Jenni and Paolo Favaro.
\newblock Self-supervised feature learning by learning to spot artifacts.
\newblock In \emph{Proceedings of the IEEE Conference on Computer Vision and
  Pattern Recognition}, pp.\  2733--2742, 2018.

\bibitem[Jing \& Tian(2020)Jing and Tian]{jing2020self}
Longlong Jing and Yingli Tian.
\newblock Self-supervised visual feature learning with deep neural networks: A
  survey.
\newblock \emph{IEEE Transactions on Pattern Analysis and Machine
  Intelligence}, 2020.

\bibitem[Joshi et~al.(2020)Joshi, Chen, Liu, Weld, Zettlemoyer, and
  Levy]{joshi2020spanbert}
Mandar Joshi, Danqi Chen, Yinhan Liu, Daniel~S Weld, Luke Zettlemoyer, and Omer
  Levy.
\newblock Spanbert: Improving pre-training by representing and predicting
  spans.
\newblock \emph{Transactions of the Association for Computational Linguistics},
  8:\penalty0 64--77, 2020.

\bibitem[Lample \& Conneau(2019)Lample and Conneau]{lample2019cross}
Guillaume Lample and Alexis Conneau.
\newblock Cross-lingual language model pretraining.
\newblock \emph{arXiv preprint arXiv:1901.07291}, 2019.

\bibitem[Lan et~al.(2019)Lan, Chen, Goodman, Gimpel, Sharma, and
  Soricut]{lan2019albert}
Zhenzhong Lan, Mingda Chen, Sebastian Goodman, Kevin Gimpel, Piyush Sharma, and
  Radu Soricut.
\newblock Albert: A lite bert for self-supervised learning of language
  representations.
\newblock \emph{arXiv preprint arXiv:1909.11942}, 2019.

\bibitem[Larsson et~al.(2016)Larsson, Maire, and
  Shakhnarovich]{larsson2016learning}
Gustav Larsson, Michael Maire, and Gregory Shakhnarovich.
\newblock Learning representations for automatic colorization.
\newblock In \emph{European conference on computer vision}, pp.\  577--593.
  Springer, 2016.

\bibitem[Larsson et~al.(2017)Larsson, Maire, and
  Shakhnarovich]{larsson2017colorization}
Gustav Larsson, Michael Maire, and Gregory Shakhnarovich.
\newblock Colorization as a proxy task for visual understanding.
\newblock In \emph{Proceedings of the IEEE Conference on Computer Vision and
  Pattern Recognition}, pp.\  6874--6883, 2017.

\bibitem[Lewis et~al.(2019)Lewis, Liu, Goyal, Ghazvininejad, Mohamed, Levy,
  Stoyanov, and Zettlemoyer]{lewis2019bart}
Mike Lewis, Yinhan Liu, Naman Goyal, Marjan Ghazvininejad, Abdelrahman Mohamed,
  Omer Levy, Ves Stoyanov, and Luke Zettlemoyer.
\newblock Bart: Denoising sequence-to-sequence pre-training for natural
  language generation, translation, and comprehension.
\newblock \emph{arXiv preprint arXiv:1910.13461}, 2019.

\bibitem[Li et~al.(2016)Li, Paluri, Rehg, and Doll{\'a}r]{li2016unsupervised}
Yin Li, Manohar Paluri, James~M Rehg, and Piotr Doll{\'a}r.
\newblock Unsupervised learning of edges.
\newblock In \emph{Proceedings of the IEEE Conference on Computer Vision and
  Pattern Recognition}, pp.\  1619--1627, 2016.

\bibitem[Lim et~al.(2019)Lim, Kim, Kim, Kim, and Kim]{lim2019fast}
Sungbin Lim, Ildoo Kim, Taesup Kim, Chiheon Kim, and Sungwoong Kim.
\newblock Fast autoaugment.
\newblock \emph{arXiv preprint arXiv:1905.00397}, 2019.

\bibitem[Liu et~al.(2019)Liu, Ott, Goyal, Du, Joshi, Chen, Levy, Lewis,
  Zettlemoyer, and Stoyanov]{liu2019roberta}
Yinhan Liu, Myle Ott, Naman Goyal, Jingfei Du, Mandar Joshi, Danqi Chen, Omer
  Levy, Mike Lewis, Luke Zettlemoyer, and Veselin Stoyanov.
\newblock Roberta: A robustly optimized bert pretraining approach.
\newblock \emph{arXiv preprint arXiv:1907.11692}, 2019.

\bibitem[Lopes et~al.(2019)Lopes, Yin, Poole, Gilmer, and
  Cubuk]{lopes2019improving}
Raphael~Gontijo Lopes, Dong Yin, Ben Poole, Justin Gilmer, and Ekin~D Cubuk.
\newblock Improving robustness without sacrificing accuracy with patch gaussian
  augmentation.
\newblock \emph{arXiv preprint arXiv:1906.02611}, 2019.

\bibitem[Lukasik et~al.(2020)Lukasik, Bhojanapalli, Menon, and
  Kumar]{lukasik2020does}
Michal Lukasik, Srinadh Bhojanapalli, Aditya Menon, and Sanjiv Kumar.
\newblock Does label smoothing mitigate label noise?
\newblock In \emph{International Conference on Machine Learning}, pp.\
  6448--6458. PMLR, 2020.

\bibitem[McClosky et~al.(2006)McClosky, Charniak, and
  Johnson]{mcclosky2006effective}
David McClosky, Eugene Charniak, and Mark Johnson.
\newblock Effective self-training for parsing.
\newblock In \emph{Proceedings of the Human Language Technology Conference of
  the NAACL, Main Conference}, pp.\  152--159, 2006.

\bibitem[Mikolov et~al.(2013)Mikolov, Sutskever, Chen, Corrado, and
  Dean]{mikolov2013distributed}
Tomas Mikolov, Ilya Sutskever, Kai Chen, Greg Corrado, and Jeffrey Dean.
\newblock Distributed representations of words and phrases and their
  compositionality.
\newblock \emph{arXiv preprint arXiv:1310.4546}, 2013.

\bibitem[Misra \& Maaten(2020)Misra and Maaten]{misra2020self}
Ishan Misra and Laurens van~der Maaten.
\newblock Self-supervised learning of pretext-invariant representations.
\newblock In \emph{Proceedings of the IEEE/CVF Conference on Computer Vision
  and Pattern Recognition}, pp.\  6707--6717, 2020.

\bibitem[Mnih \& Kavukcuoglu(2013)Mnih and Kavukcuoglu]{mnih2013learning}
Andriy Mnih and Koray Kavukcuoglu.
\newblock Learning word embeddings efficiently with noise-contrastive
  estimation.
\newblock \emph{Advances in neural information processing systems},
  26:\penalty0 2265--2273, 2013.

\bibitem[M{\"u}ller et~al.(2019)M{\"u}ller, Kornblith, and
  Hinton]{muller2019does}
Rafael M{\"u}ller, Simon Kornblith, and Geoffrey Hinton.
\newblock When does label smoothing help?
\newblock \emph{arXiv preprint arXiv:1906.02629}, 2019.

\bibitem[Oord et~al.(2018)Oord, Li, and Vinyals]{oord2018representation}
Aaron van~den Oord, Yazhe Li, and Oriol Vinyals.
\newblock Representation learning with contrastive predictive coding.
\newblock \emph{arXiv preprint arXiv:1807.03748}, 2018.

\bibitem[Park et~al.(2019)Park, Chan, Zhang, Chiu, Zoph, Cubuk, and
  Le]{park2019specaugment}
Daniel~S Park, William Chan, Yu~Zhang, Chung-Cheng Chiu, Barret Zoph, Ekin~D
  Cubuk, and Quoc~V Le.
\newblock Specaugment: A simple data augmentation method for automatic speech
  recognition.
\newblock \emph{arXiv preprint arXiv:1904.08779}, 2019.

\bibitem[Perez \& Wang(2017)Perez and Wang]{perez2017effectiveness}
Luis Perez and Jason Wang.
\newblock The effectiveness of data augmentation in image classification using
  deep learning.
\newblock \emph{arXiv preprint arXiv:1712.04621}, 2017.

\bibitem[Purushwalkam \& Gupta(2020)Purushwalkam and
  Gupta]{purushwalkam2020demystifying}
Senthil Purushwalkam and Abhinav Gupta.
\newblock Demystifying contrastive self-supervised learning: Invariances,
  augmentations and dataset biases.
\newblock \emph{arXiv preprint arXiv:2007.13916}, 2020.

\bibitem[Qiu et~al.(2020)Qiu, Sun, Xu, Shao, Dai, and Huang]{qiu2020pre}
Xipeng Qiu, Tianxiang Sun, Yige Xu, Yunfan Shao, Ning Dai, and Xuanjing Huang.
\newblock Pre-trained models for natural language processing: A survey.
\newblock \emph{Science China Technological Sciences}, pp.\  1--26, 2020.

\bibitem[Radford et~al.(2015)Radford, Metz, and
  Chintala]{radford2015unsupervised}
Alec Radford, Luke Metz, and Soumith Chintala.
\newblock Unsupervised representation learning with deep convolutional
  generative adversarial networks.
\newblock \emph{arXiv preprint arXiv:1511.06434}, 2015.

\bibitem[Raffel et~al.(2019)Raffel, Shazeer, Roberts, Lee, Narang, Matena,
  Zhou, Li, and Liu]{raffel2019exploring}
Colin Raffel, Noam Shazeer, Adam Roberts, Katherine Lee, Sharan Narang, Michael
  Matena, Yanqi Zhou, Wei Li, and Peter~J Liu.
\newblock Exploring the limits of transfer learning with a unified text-to-text
  transformer.
\newblock \emph{arXiv preprint arXiv:1910.10683}, 2019.

\bibitem[Ratner et~al.(2017)Ratner, Ehrenberg, Hussain, Dunnmon, and
  R{\'e}]{ratner2017learning}
Alexander~J Ratner, Henry~R Ehrenberg, Zeshan Hussain, Jared Dunnmon, and
  Christopher R{\'e}.
\newblock Learning to compose domain-specific transformations for data
  augmentation.
\newblock \emph{Advances in neural information processing systems},
  30:\penalty0 3239, 2017.

\bibitem[Ruder \& Plank(2018)Ruder and Plank]{ruder2018strong}
Sebastian Ruder and Barbara Plank.
\newblock Strong baselines for neural semi-supervised learning under domain
  shift.
\newblock \emph{arXiv preprint arXiv:1804.09530}, 2018.

\bibitem[Rumelhart et~al.(1985)Rumelhart, Hinton, and
  Williams]{rumelhart1985learning}
David~E Rumelhart, Geoffrey~E Hinton, and Ronald~J Williams.
\newblock Learning internal representations by error propagation.
\newblock Technical report, California Univ San Diego La Jolla Inst for
  Cognitive Science, 1985.

\bibitem[Rusiecki(2020)]{rusiecki2020standard}
Andrzej Rusiecki.
\newblock Standard dropout as remedy for training deep neural networks with
  label noise.
\newblock In \emph{International Conference on Dependability and Complex
  Systems}, pp.\  534--542. Springer, 2020.

\bibitem[Song et~al.(2019)Song, Tan, Qin, Lu, and Liu]{song2019mass}
Kaitao Song, Xu~Tan, Tao Qin, Jianfeng Lu, and Tie-Yan Liu.
\newblock Mass: Masked sequence to sequence pre-training for language
  generation.
\newblock \emph{arXiv preprint arXiv:1905.02450}, 2019.

\bibitem[Song et~al.(2020)Song, Tan, Qin, Lu, and Liu]{song2020mpnet}
Kaitao Song, Xu~Tan, Tao Qin, Jianfeng Lu, and Tie-Yan Liu.
\newblock Mpnet: Masked and permuted pre-training for language understanding.
\newblock \emph{arXiv preprint arXiv:2004.09297}, 2020.

\bibitem[Srivastava et~al.(2014)Srivastava, Hinton, Krizhevsky, Sutskever, and
  Salakhutdinov]{srivastava2014dropout}
Nitish Srivastava, Geoffrey Hinton, Alex Krizhevsky, Ilya Sutskever, and Ruslan
  Salakhutdinov.
\newblock Dropout: a simple way to prevent neural networks from overfitting.
\newblock \emph{The journal of machine learning research}, 15\penalty0
  (1):\penalty0 1929--1958, 2014.

\bibitem[Szegedy et~al.(2016)Szegedy, Vanhoucke, Ioffe, Shlens, and
  Wojna]{szegedy2016rethinking}
Christian Szegedy, Vincent Vanhoucke, Sergey Ioffe, Jon Shlens, and Zbigniew
  Wojna.
\newblock Rethinking the inception architecture for computer vision.
\newblock In \emph{Proceedings of the IEEE conference on computer vision and
  pattern recognition}, pp.\  2818--2826, 2016.

\bibitem[Tamkin et~al.(2020)Tamkin, Wu, and Goodman]{tamkin2020viewmaker}
Alex Tamkin, Mike Wu, and Noah Goodman.
\newblock Viewmaker networks: Learning views for unsupervised representation
  learning.
\newblock \emph{arXiv preprint arXiv:2010.07432}, 2020.

\bibitem[Tian et~al.(2019)Tian, Krishnan, and Isola]{tian2019contrastive}
Yonglong Tian, Dilip Krishnan, and Phillip Isola.
\newblock Contrastive multiview coding.
\newblock \emph{arXiv preprint arXiv:1906.05849}, 2019.

\bibitem[Tran et~al.(2017)Tran, Pham, Carneiro, Palmer, and
  Reid]{tran2017bayesian}
Toan Tran, Trung Pham, Gustavo Carneiro, Lyle Palmer, and Ian Reid.
\newblock A bayesian data augmentation approach for learning deep models.
\newblock \emph{arXiv preprint arXiv:1710.10564}, 2017.

\bibitem[Tung et~al.(2017)Tung, Tung, Yumer, and Fragkiadaki]{tung2017self}
Hsiao-Yu~Fish Tung, Hsiao-Wei Tung, Ersin Yumer, and Katerina Fragkiadaki.
\newblock Self-supervised learning of motion capture.
\newblock \emph{arXiv preprint arXiv:1712.01337}, 2017.

\bibitem[Vanschoren et~al.(2013)Vanschoren, van Rijn, Bischl, and
  Torgo]{OpenML2013}
Joaquin Vanschoren, Jan~N. van Rijn, Bernd Bischl, and Luis Torgo.
\newblock Openml: Networked science in machine learning.
\newblock \emph{SIGKDD Explorations}, 15\penalty0 (2):\penalty0 49--60, 2013.
\newblock \doi{10.1145/2641190.2641198}.
\newblock URL \url{http://doi.acm.org/10.1145/2641190.2641198}.

\bibitem[Vincent et~al.(2008)Vincent, Larochelle, Bengio, and
  Manzagol]{vincent2008extracting}
Pascal Vincent, Hugo Larochelle, Yoshua Bengio, and Pierre-Antoine Manzagol.
\newblock Extracting and composing robust features with denoising autoencoders.
\newblock In \emph{Proceedings of the 25th international conference on Machine
  learning}, pp.\  1096--1103, 2008.

\bibitem[Vincent et~al.(2010)Vincent, Larochelle, Lajoie, Bengio, Manzagol, and
  Bottou]{vincent2010stacked}
Pascal Vincent, Hugo Larochelle, Isabelle Lajoie, Yoshua Bengio, Pierre-Antoine
  Manzagol, and L{\'e}on Bottou.
\newblock Stacked denoising autoencoders: Learning useful representations in a
  deep network with a local denoising criterion.
\newblock \emph{Journal of machine learning research}, 11\penalty0 (12), 2010.

\bibitem[Wang et~al.(2019)Wang, Zhao, Jia, Li, and Liu]{wang2019denoising}
Liang Wang, Wei Zhao, Ruoyu Jia, Sujian Li, and Jingming Liu.
\newblock Denoising based sequence-to-sequence pre-training for text
  generation.
\newblock \emph{arXiv preprint arXiv:1908.08206}, 2019.

\bibitem[Wang \& Isola(2020)Wang and Isola]{wang2020understanding}
Tongzhou Wang and Phillip Isola.
\newblock Understanding contrastive representation learning through alignment
  and uniformity on the hypersphere.
\newblock In \emph{International Conference on Machine Learning}, pp.\
  9929--9939. PMLR, 2020.

\bibitem[Wang \& Gupta(2015)Wang and Gupta]{wang2015unsupervised}
Xiaolong Wang and Abhinav Gupta.
\newblock Unsupervised learning of visual representations using videos.
\newblock In \emph{Proceedings of the IEEE international conference on computer
  vision}, pp.\  2794--2802, 2015.

\bibitem[Wu et~al.(2018)Wu, Xiong, Yu, and Lin]{wu2018unsupervised}
Zhirong Wu, Yuanjun Xiong, Stella Yu, and Dahua Lin.
\newblock Unsupervised feature learning via non-parametric instance-level
  discrimination.
\newblock \emph{arXiv preprint arXiv:1805.01978}, 2018.

\bibitem[Yang et~al.(2019)Yang, Dai, Yang, Carbonell, Salakhutdinov, and
  Le]{yang2019xlnet}
Zhilin Yang, Zihang Dai, Yiming Yang, Jaime Carbonell, Ruslan Salakhutdinov,
  and Quoc~V Le.
\newblock Xlnet: Generalized autoregressive pretraining for language
  understanding.
\newblock \emph{arXiv preprint arXiv:1906.08237}, 2019.

\bibitem[Yao et~al.(2020)Yao, Yi, Cheng, Yu, Chen, Menon, Hong, Chi, Tjoa,
  Kang, et~al.]{yao2020self}
Tiansheng Yao, Xinyang Yi, Derek~Zhiyuan Cheng, Felix Yu, Ting Chen, Aditya
  Menon, Lichan Hong, Ed~H Chi, Steve Tjoa, Jieqi Kang, et~al.
\newblock Self-supervised learning for large-scale item recommendations.
\newblock \emph{arXiv preprint arXiv:2007.12865}, 2020.

\bibitem[Yarowsky(1995)]{yarowsky1995unsupervised}
David Yarowsky.
\newblock Unsupervised word sense disambiguation rivaling supervised methods.
\newblock In \emph{33rd annual meeting of the association for computational
  linguistics}, pp.\  189--196, 1995.

\bibitem[Yoon et~al.(2020)Yoon, Zhang, Jordon, and van~der
  Schaar]{yoon2020vime}
Jinsung Yoon, Yao Zhang, James Jordon, and Mihaela van~der Schaar.
\newblock Vime: Extending the success of self-and semi-supervised learning to
  tabular domain.
\newblock \emph{Advances in Neural Information Processing Systems},
  33:\penalty0 11033--11043, 2020.

\bibitem[Zbontar et~al.(2021)Zbontar, Jing, Misra, LeCun, and
  Deny]{zbontar2021barlow}
Jure Zbontar, Li~Jing, Ishan Misra, Yann LeCun, and St{\'e}phane Deny.
\newblock Barlow twins: Self-supervised learning via redundancy reduction.
\newblock \emph{arXiv preprint arXiv:2103.03230}, 2021.

\bibitem[Zhai et~al.(2019)Zhai, Oliver, Kolesnikov, and Beyer]{zhai2019s4l}
Xiaohua Zhai, Avital Oliver, Alexander Kolesnikov, and Lucas Beyer.
\newblock S4l: Self-supervised semi-supervised learning.
\newblock In \emph{Proceedings of the IEEE/CVF International Conference on
  Computer Vision}, pp.\  1476--1485, 2019.

\bibitem[Zhang et~al.(2017)Zhang, Cisse, Dauphin, and
  Lopez-Paz]{zhang2017mixup}
Hongyi Zhang, Moustapha Cisse, Yann~N Dauphin, and David Lopez-Paz.
\newblock mixup: Beyond empirical risk minimization.
\newblock \emph{arXiv preprint arXiv:1710.09412}, 2017.

\bibitem[Zhang et~al.(2019{\natexlab{a}})Zhang, Song, Gao, Chen, Bao, and
  Ma]{zhang2019your}
Linfeng Zhang, Jiebo Song, Anni Gao, Jingwei Chen, Chenglong Bao, and Kaisheng
  Ma.
\newblock Be your own teacher: Improve the performance of convolutional neural
  networks via self distillation.
\newblock In \emph{Proceedings of the IEEE/CVF International Conference on
  Computer Vision}, pp.\  3713--3722, 2019{\natexlab{a}}.

\bibitem[Zhang et~al.(2016)Zhang, Isola, and Efros]{zhang2016colorful}
Richard Zhang, Phillip Isola, and Alexei~A Efros.
\newblock Colorful image colorization.
\newblock In \emph{European conference on computer vision}, pp.\  649--666.
  Springer, 2016.

\bibitem[Zhang et~al.(2019{\natexlab{b}})Zhang, Wang, Zhang, and
  Zhong]{zhang2019adversarial}
Xinyu Zhang, Qiang Wang, Jian Zhang, and Zhao Zhong.
\newblock Adversarial autoaugment.
\newblock \emph{arXiv preprint arXiv:1912.11188}, 2019{\natexlab{b}}.

\bibitem[Zhou \& Li(2005)Zhou and Li]{zhou2005tri}
Zhi-Hua Zhou and Ming Li.
\newblock Tri-training: Exploiting unlabeled data using three classifiers.
\newblock \emph{IEEE Transactions on knowledge and Data Engineering},
  17\penalty0 (11):\penalty0 1529--1541, 2005.

\end{thebibliography}
\bibliographystyle{iclr2022_conference}

\newpage
\appendix
\section{Appendix}

We now present findings held out from the main text. Unless noted otherwise, we use the same hyperparameters (i.e. $c$ = 0.6, temperature $\tau$ = 1, etc.).

\subsection{Results for data-limited experiments}
\begin{figure}[!t]
    \centering
    \includegraphics[width=0.49\textwidth]{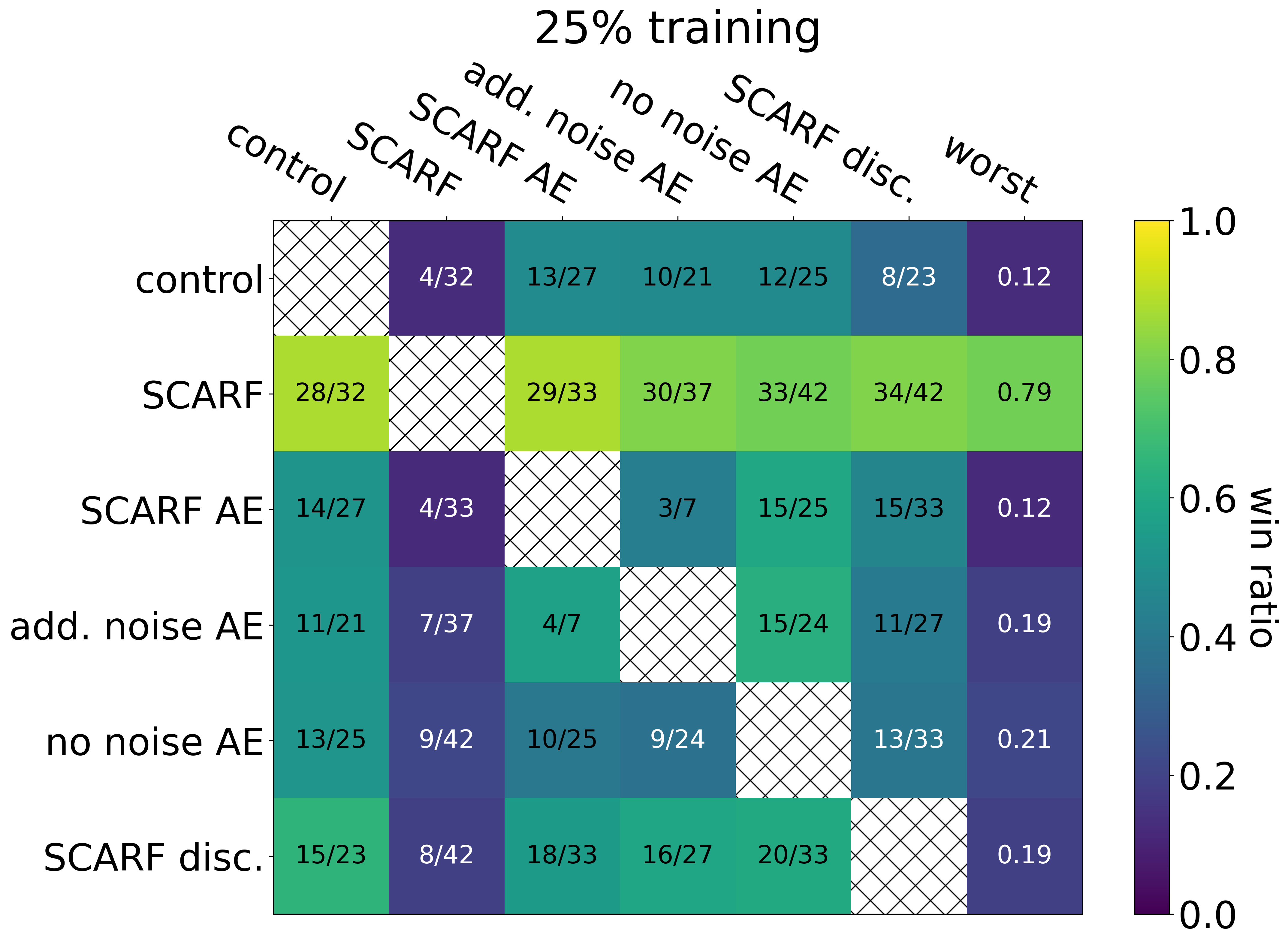}
    \includegraphics[width=0.49\textwidth]{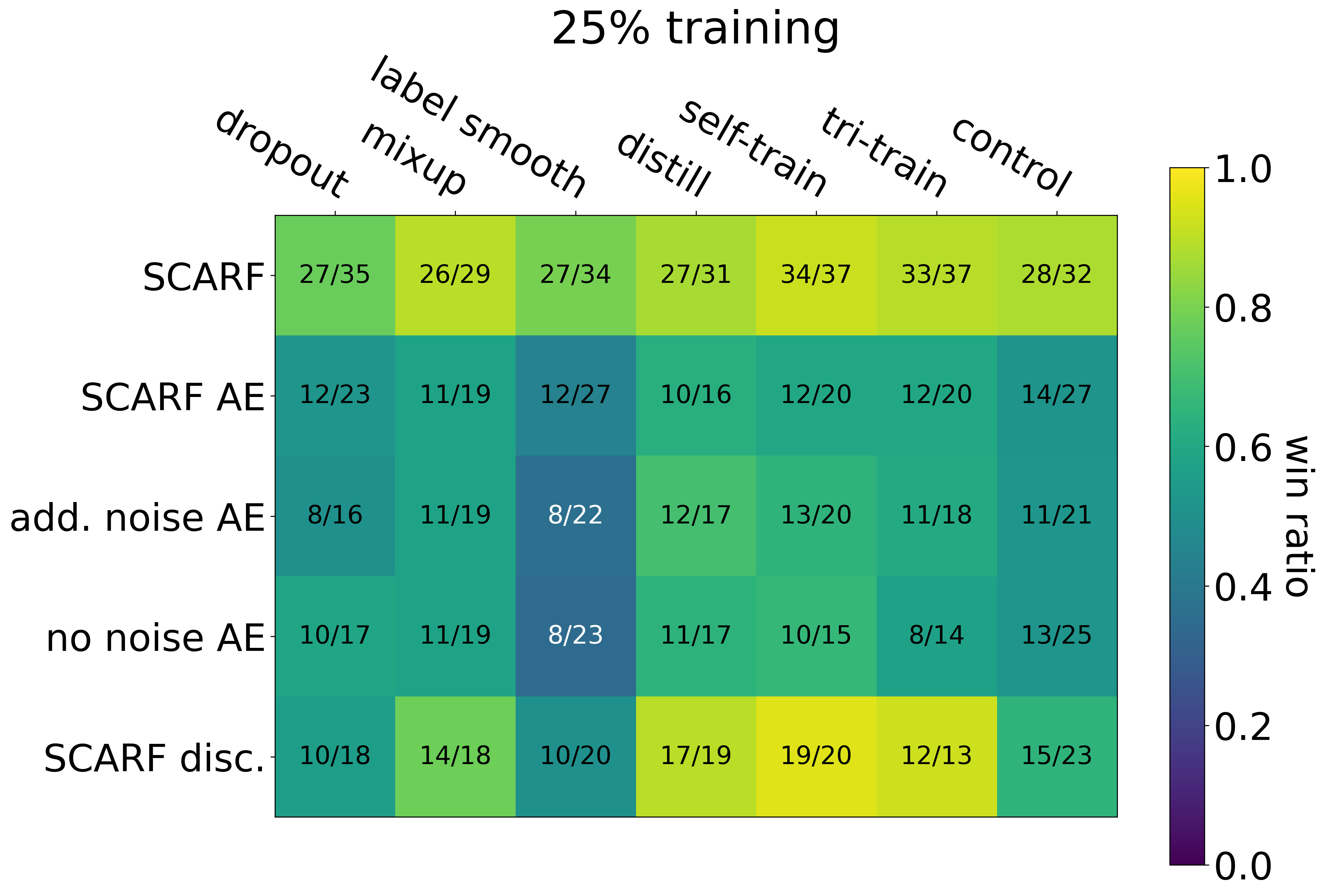}\\
    \includegraphics[width=0.32\textwidth]{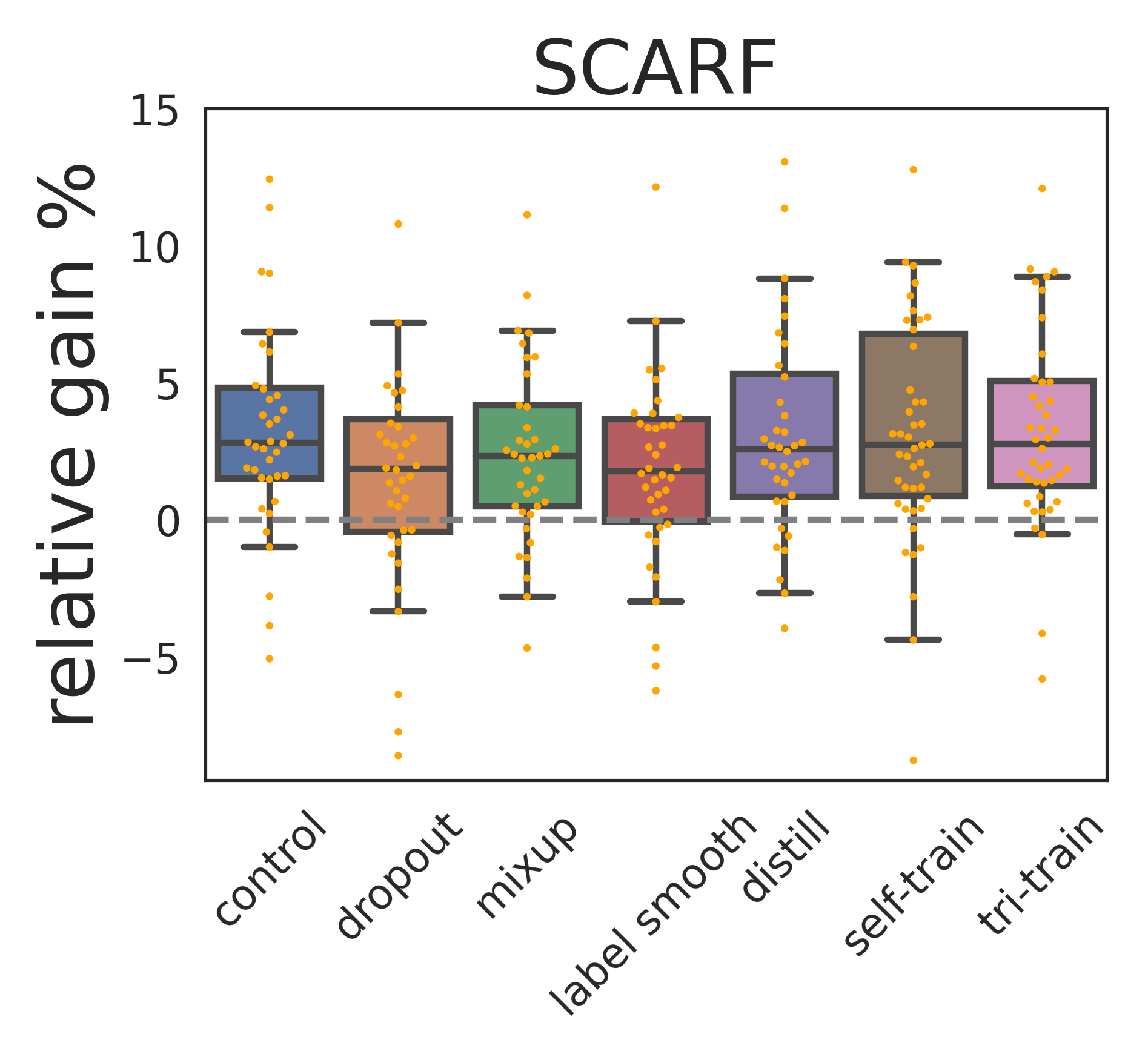} \includegraphics[width=0.32\textwidth]{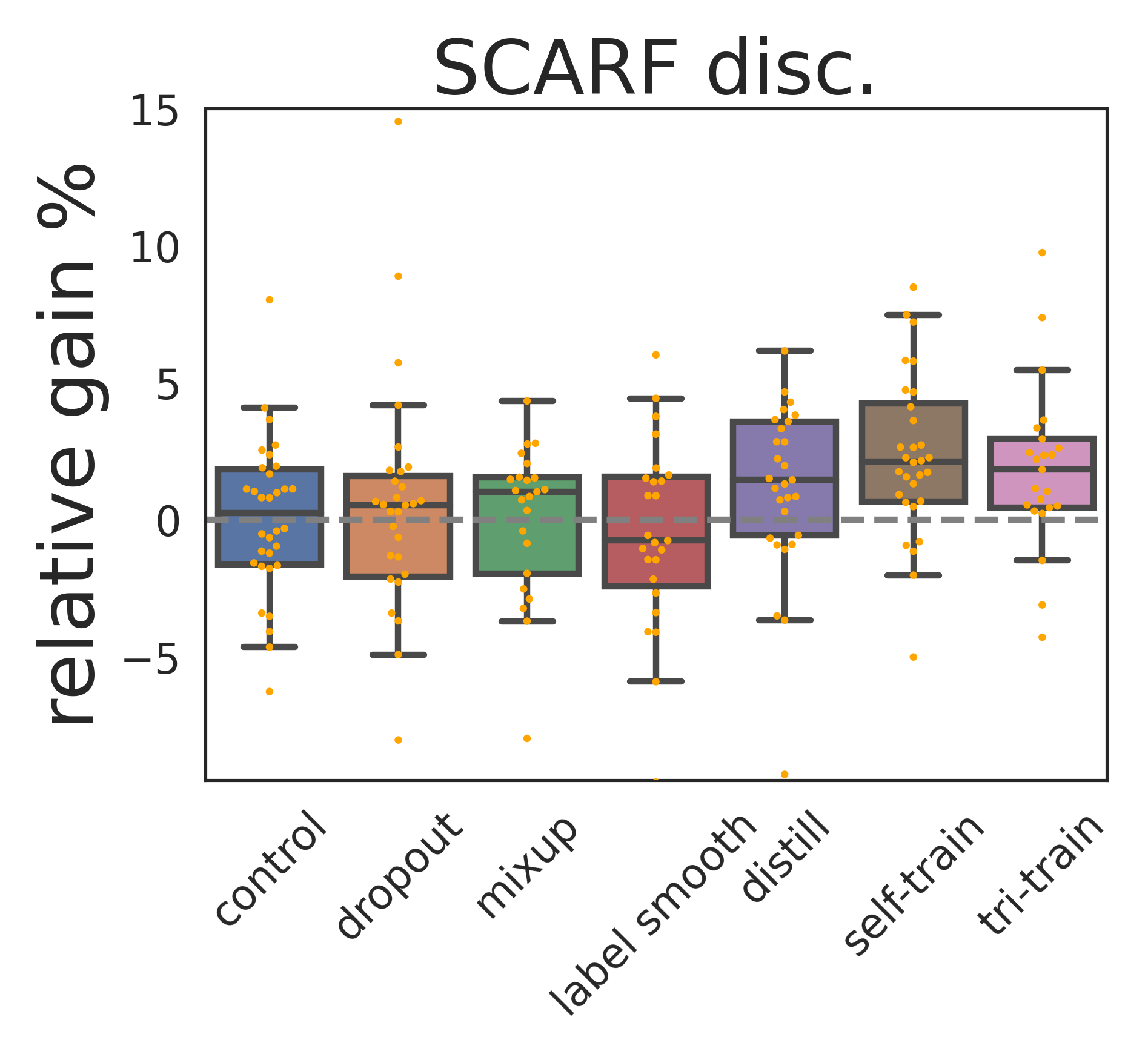}\\
    \includegraphics[width=0.32\textwidth]{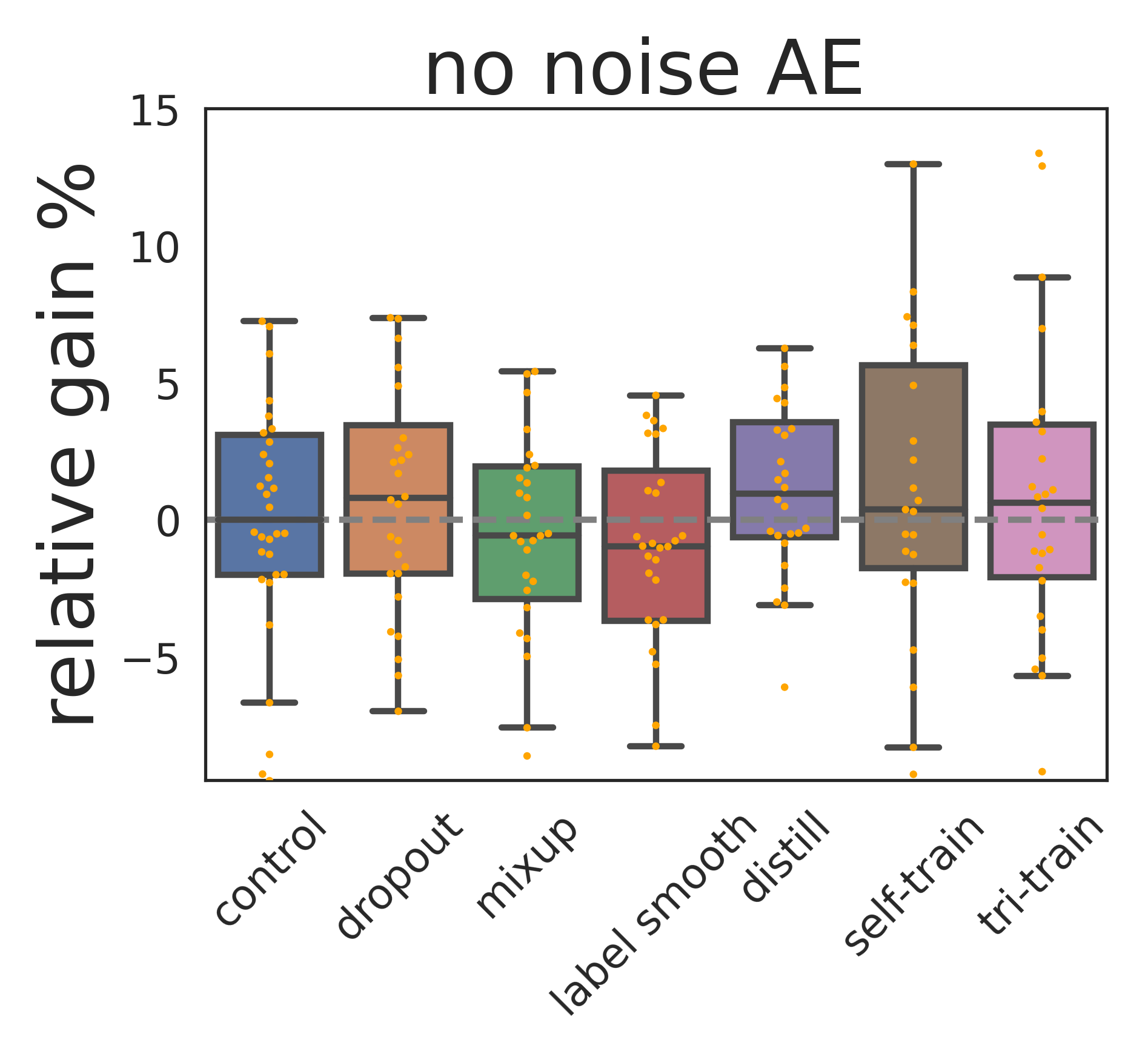}
    \includegraphics[width=0.32\textwidth]{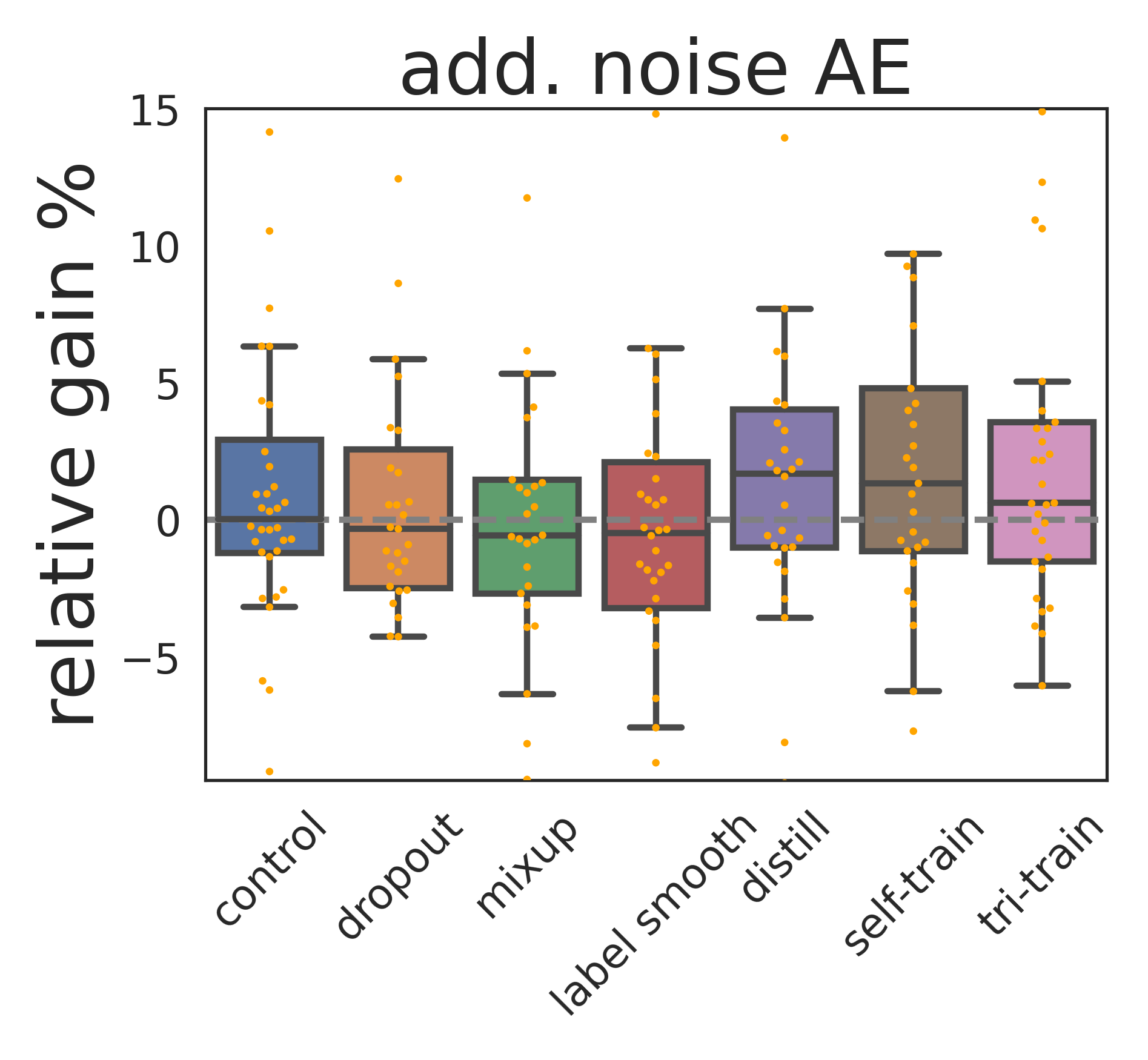}    
    \includegraphics[width=0.32\textwidth]{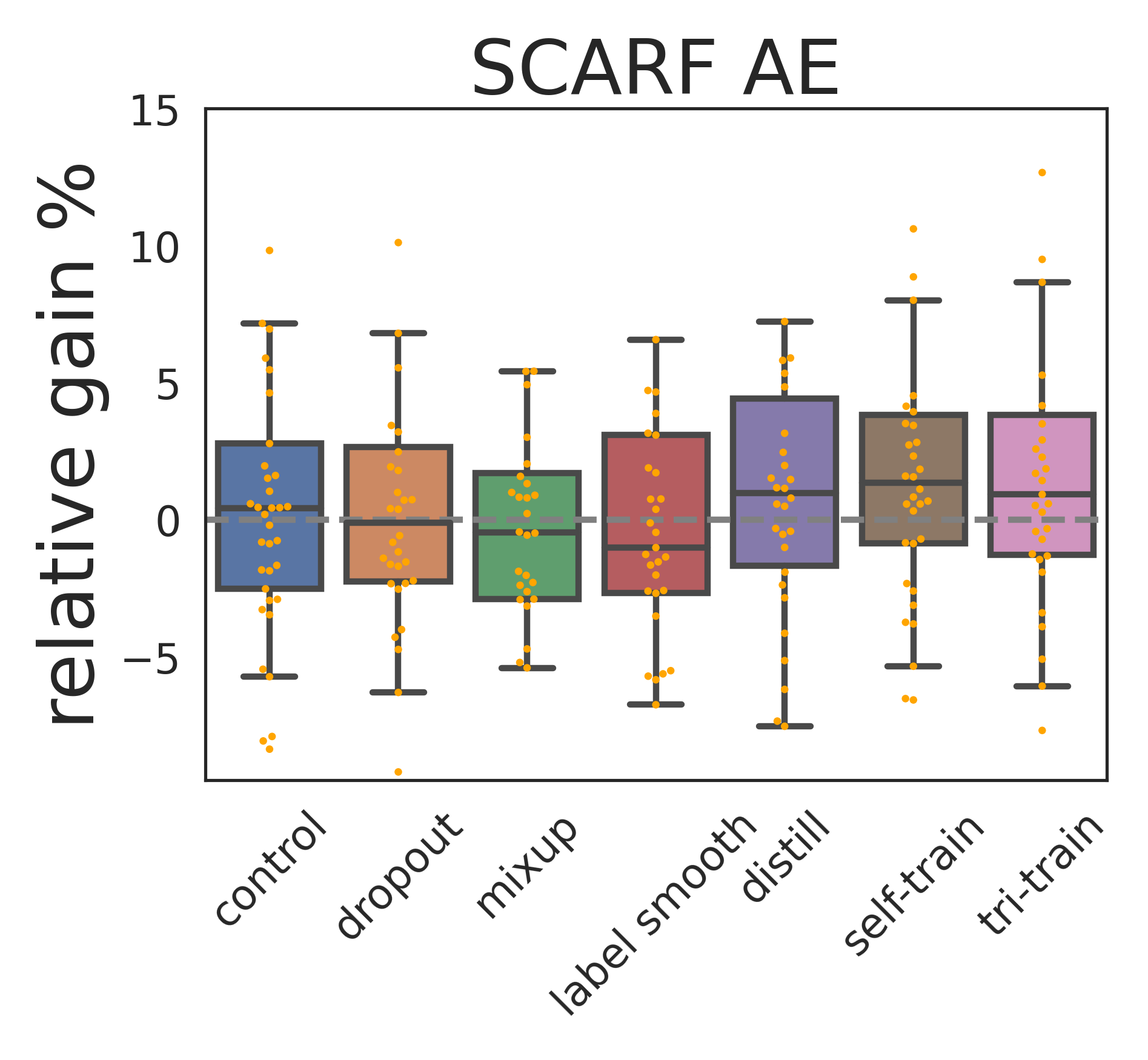}
    \caption{\textsc{Scarf} shows even more significant gain in the semi-supervised setting where $25\%$ of the data is labeled and the remaining $75\%$ is not. Strikingly, pre-training with \textsc{Scarf} boosts the performance of self-training and tri-training by several percent.} 
    \label{fig:se25}
\end{figure}

Figure~\ref{fig:se25} presents the results for the case when only $25\%$ of the training labels are available. We see that \textsc{Scarf} outperforms the rest here as well.

\subsection{Pre-training loss curves}
\begin{figure}[!t]
    \centering
    \includegraphics[width=0.48\textwidth]{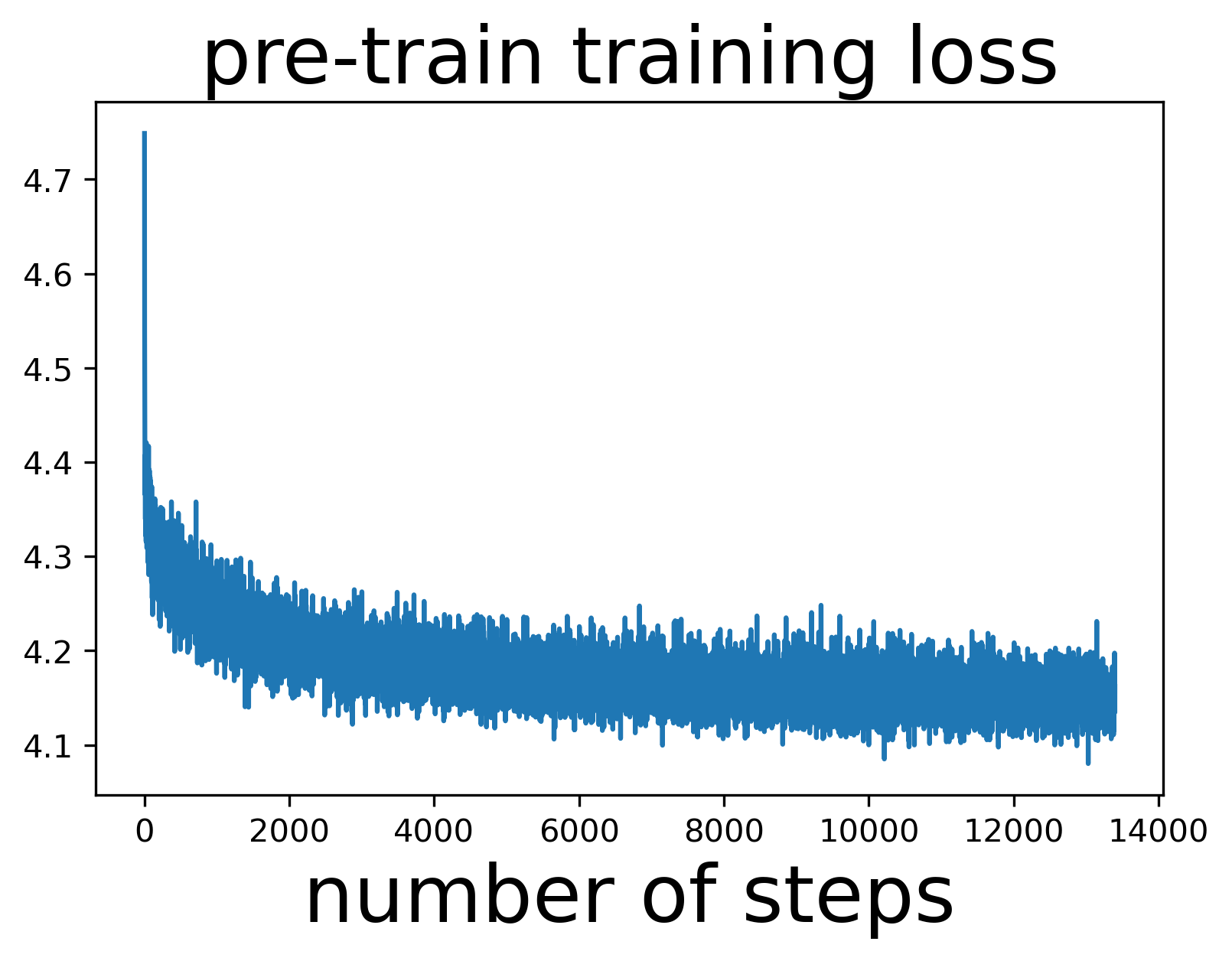}
    \includegraphics[width=0.48\textwidth]{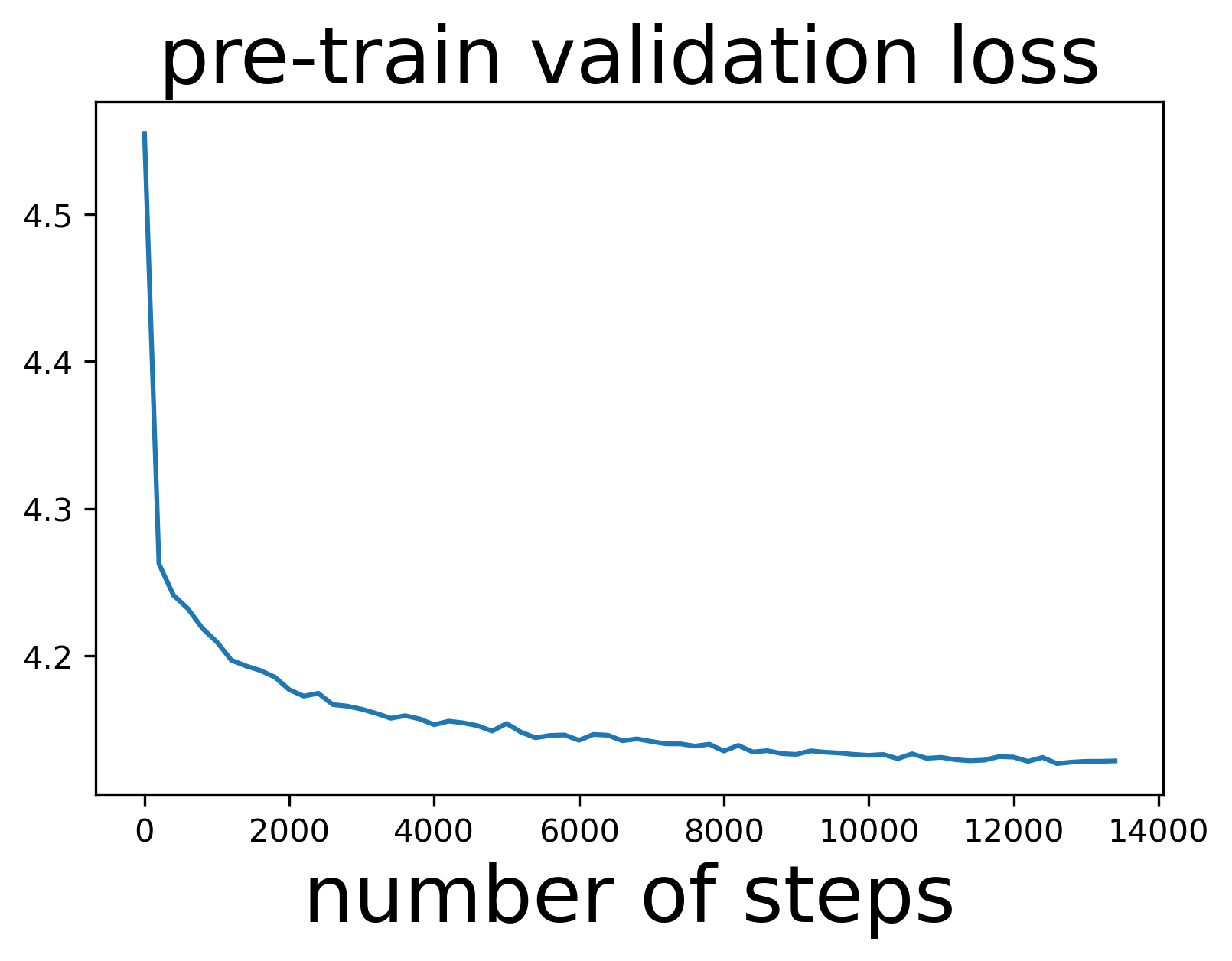}
    \caption{Training and validation loss curves for \textsc{Scarf} pre-training on Phonemes (dataset id 1489). We observe that both losses drop rapidly initially and then taper off. The training curve is jittery because of the random corruptions, but the validation curve isn't because the validation dataset is built once before training and is static throughout training.}
    \label{fig:pretrain_curves}
\end{figure}

Figure~\ref{fig:pretrain_curves} shows prototypical pre-training training and validation loss curves for \textsc{Scarf}. We see that both losses tend to decrease rapidly at the start of training and then diminish slowly until the early stopping point. We also notice the noisy (high variance) nature of the training loss - this is due to the stochasticity of our corruption method.

\subsection{Ablations Continued}
We present more ablation results, where the metric is accuracy using $100\%$ of the labeled data.

\subsection*{Impact of batch size and corruption rates}

\begin{figure}[!t]
    \centering
    \includegraphics[width=0.49\textwidth]{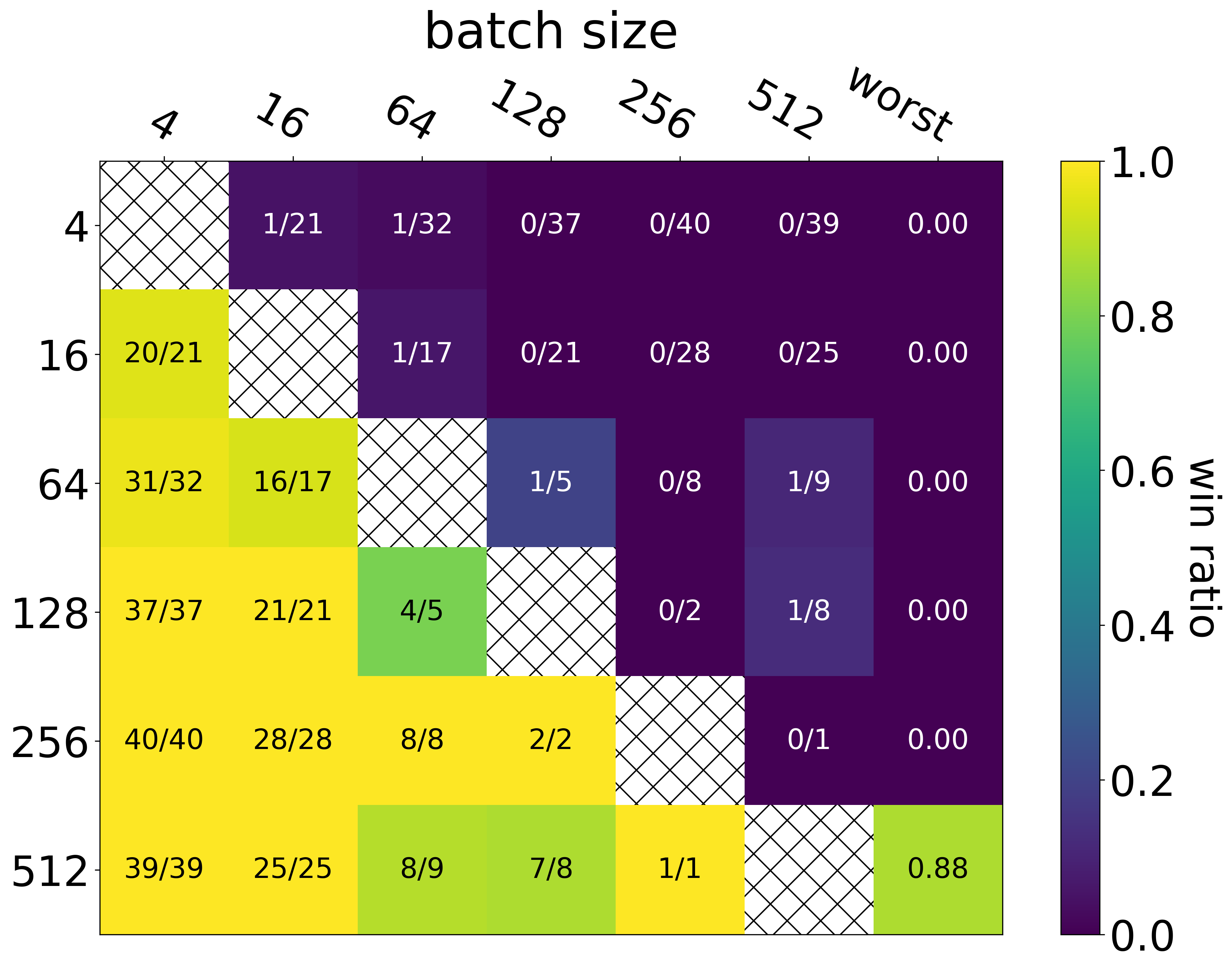}
    \includegraphics[width=0.49\textwidth]{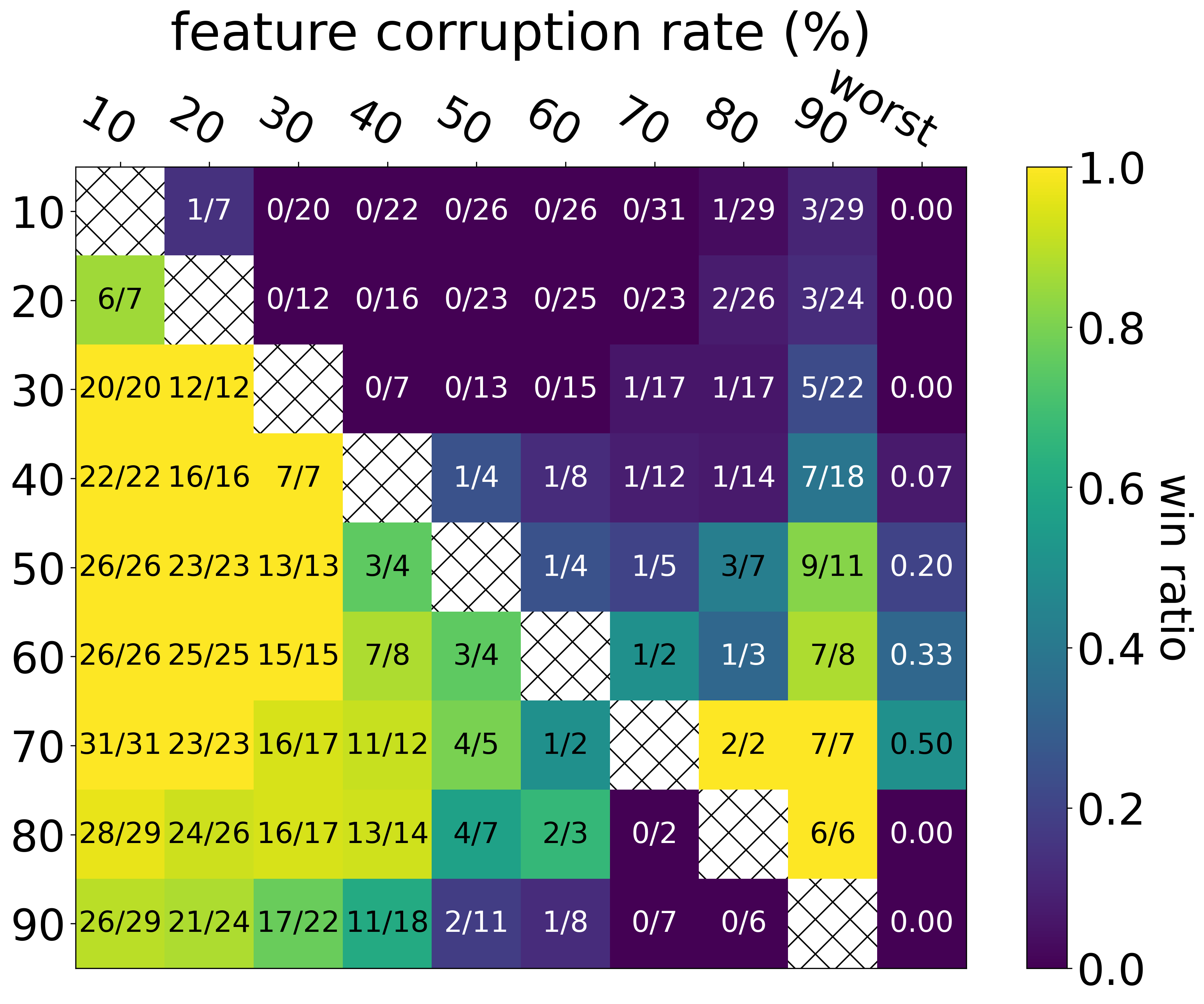}
    \caption{Win matrix for various batch sizes (\textbf{Left}) and corruption rates (\textbf{Right}) for the fully labeled, noiseless setting.}
    \label{fig:rate_ablation}
\end{figure}

Figure~\ref{fig:rate_ablation} examines the impact of the batch size and corruption rate for the fully labeled, noiseless setting.

\subsection*{Impact of corruption strategies}

\begin{figure}[!t]
    \centering
    \includegraphics[width=0.49\textwidth]{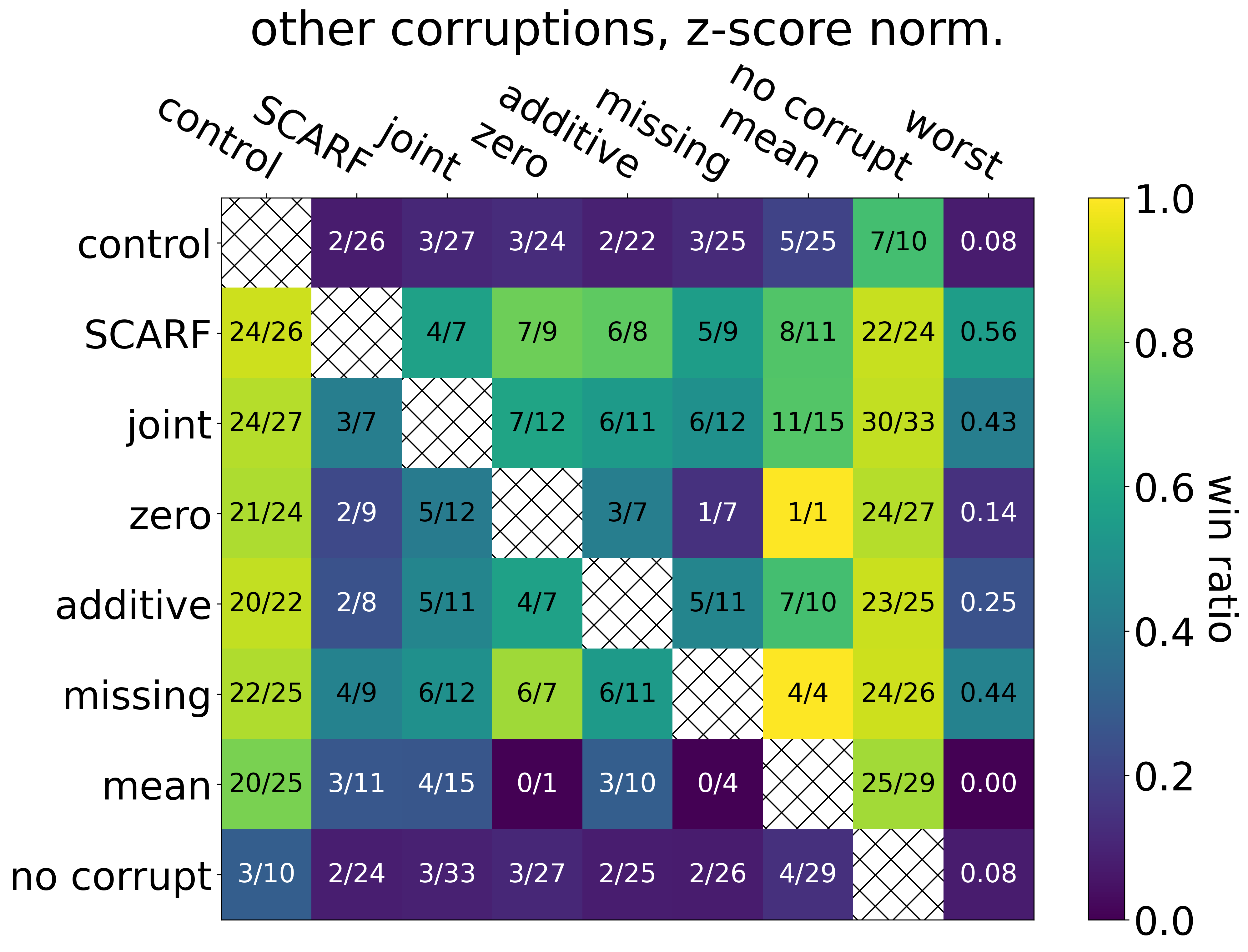}
    \includegraphics[width=0.49\textwidth]{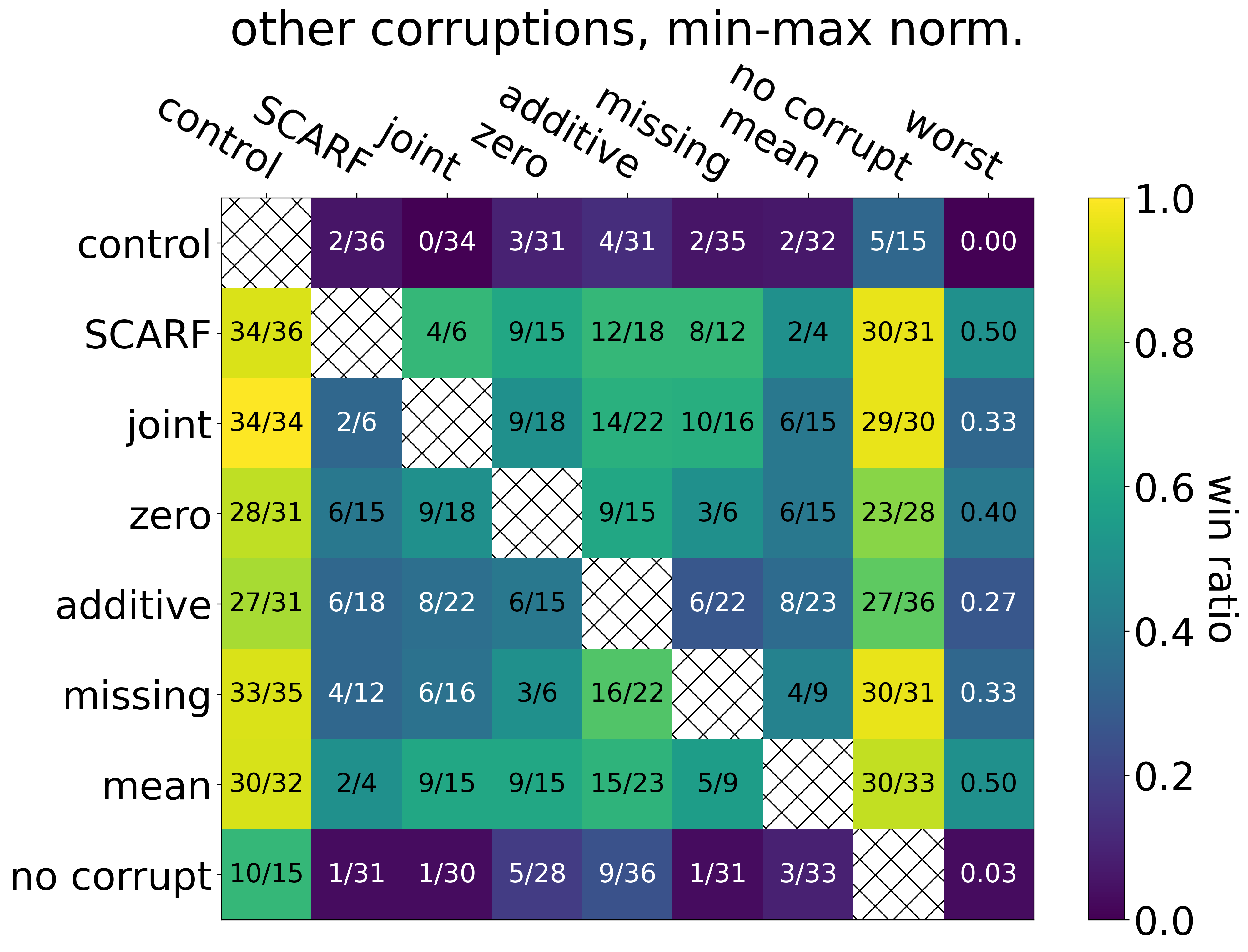}
    \caption{\textbf{Left:} Win matrix comparing different corruption strategies when z-score feature normalization is used in the fully labeled, noiseless setting. \textbf{Right:} The same matrix but when min-max feature scaling is used. We see that \textsc{Scarf} is better than alternative corruption strategies for different types of feature scaling.} 
    \label{fig:corruption_ablation}
\end{figure}

Figure~\ref{fig:corruption_ablation} shows the performance of a variety of alternative corruption strategies for \textsc{SCARF} under z-score and min-max feature normalization. Results for mean scaling are shown in Figure~\ref{fig:mean_scaling}.

\subsection*{Impact of temperature}
Figure~\ref{fig:loss_temp_ablation} shows the impact of the temperature term. While prior work considers temperature an important hyperparameter that needs to be tuned, we see that a default of 1 (i.e. just softmax) works the best in our setting.

\subsubsection*{More corruption ablations}
\begin{figure}[!t]
    \centering
    \includegraphics[width=0.48\textwidth]{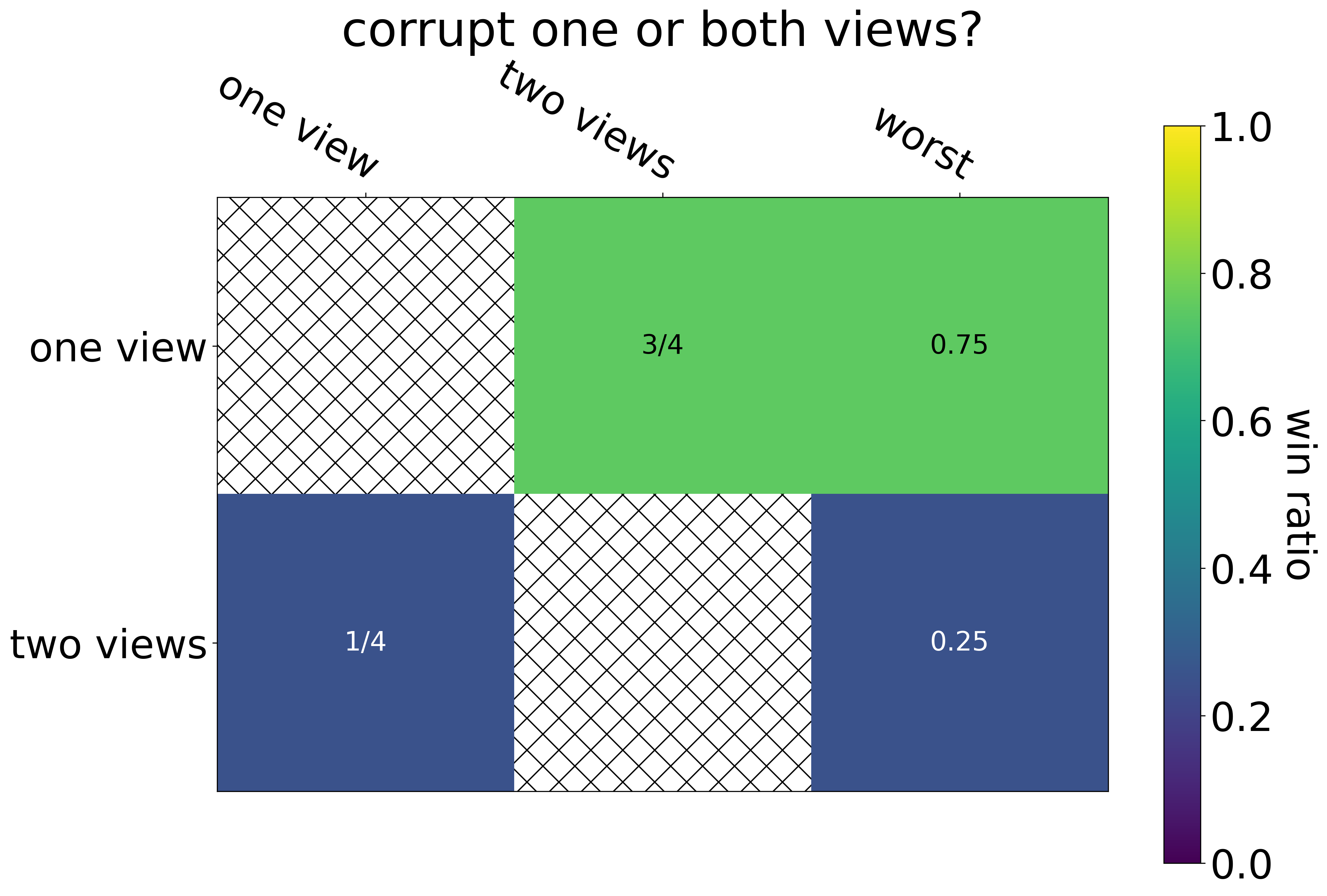}
    \includegraphics[width=0.48\textwidth]{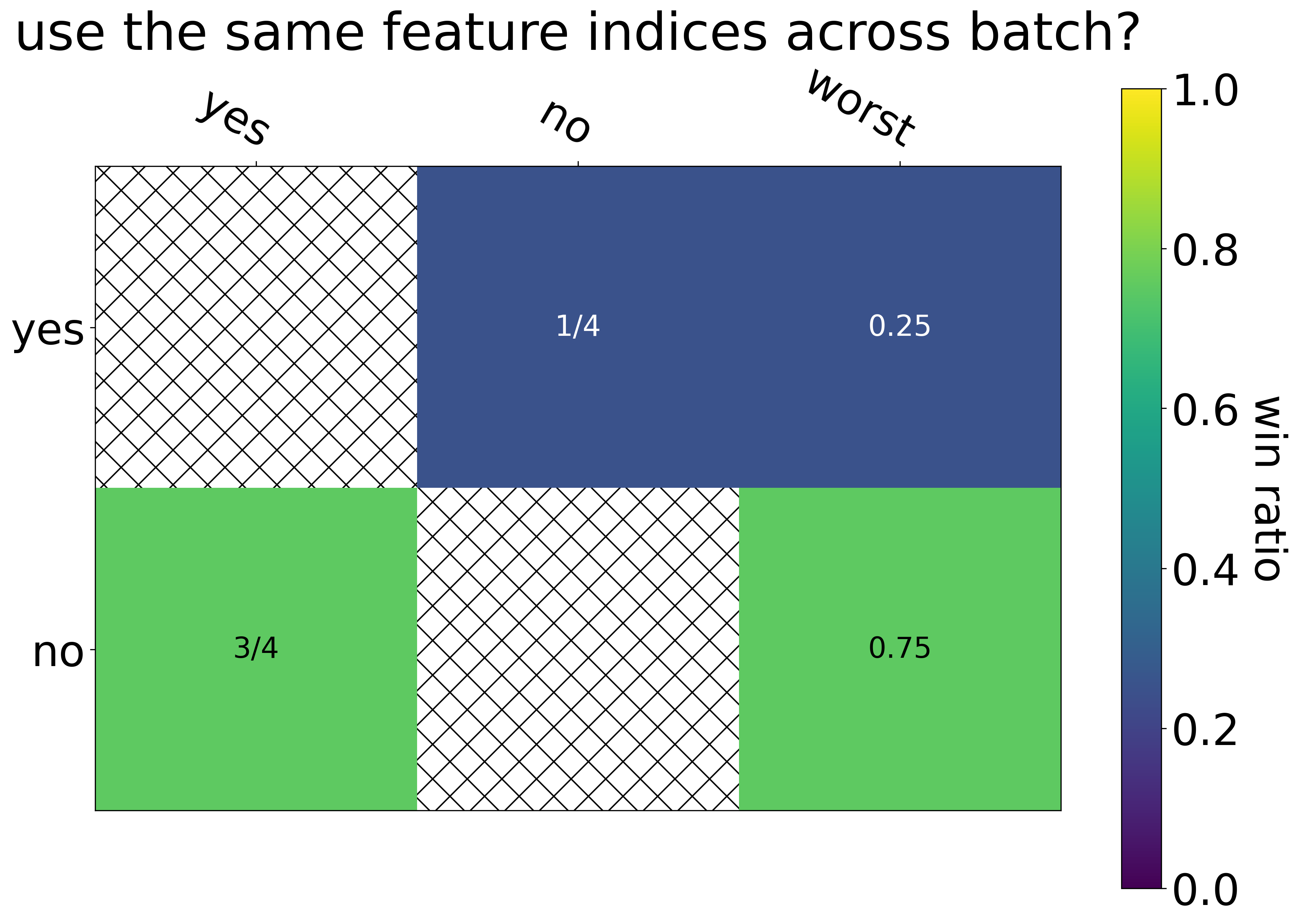}\\
    \includegraphics[width=0.48\textwidth]{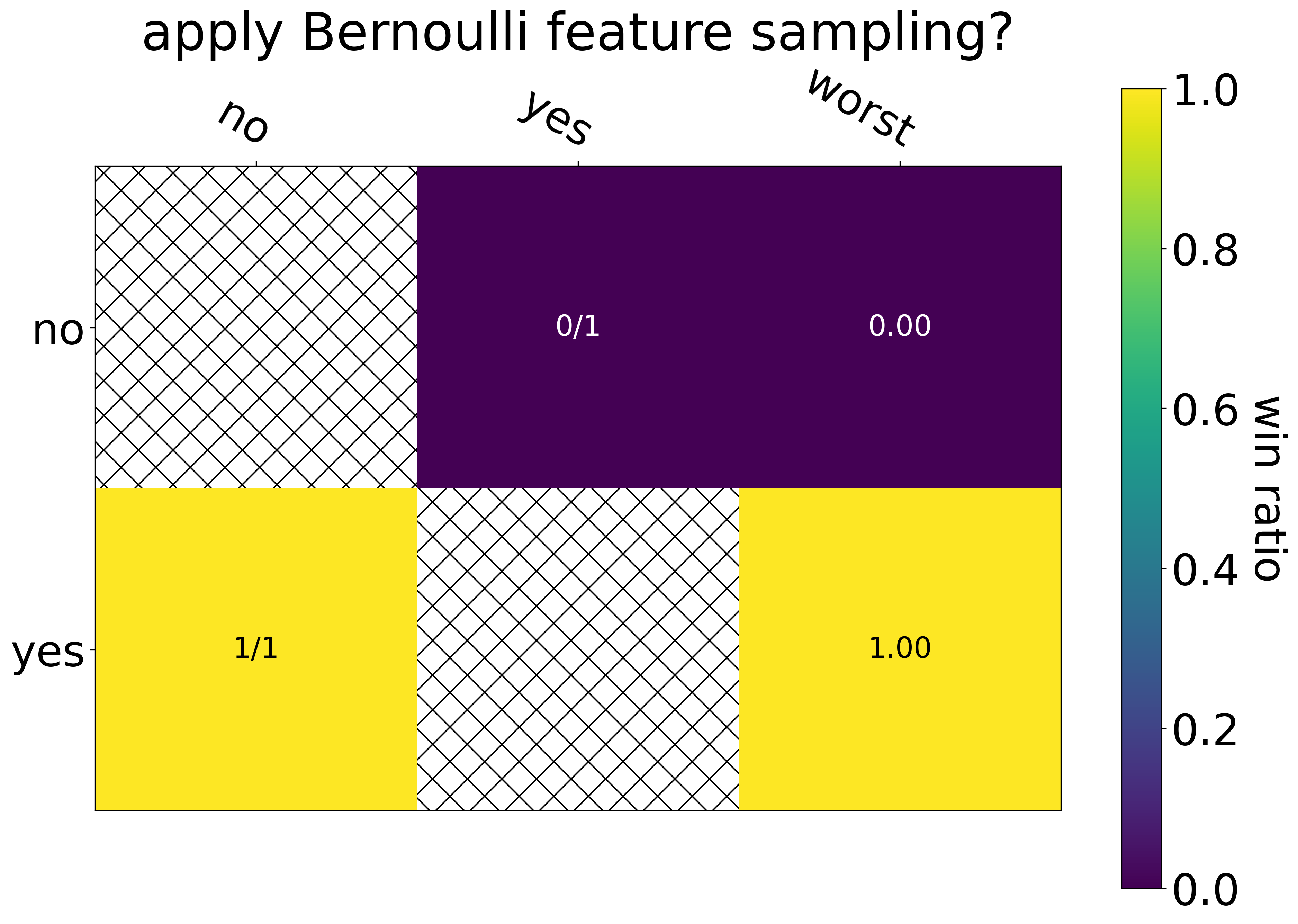}
    \includegraphics[width=0.48\textwidth]{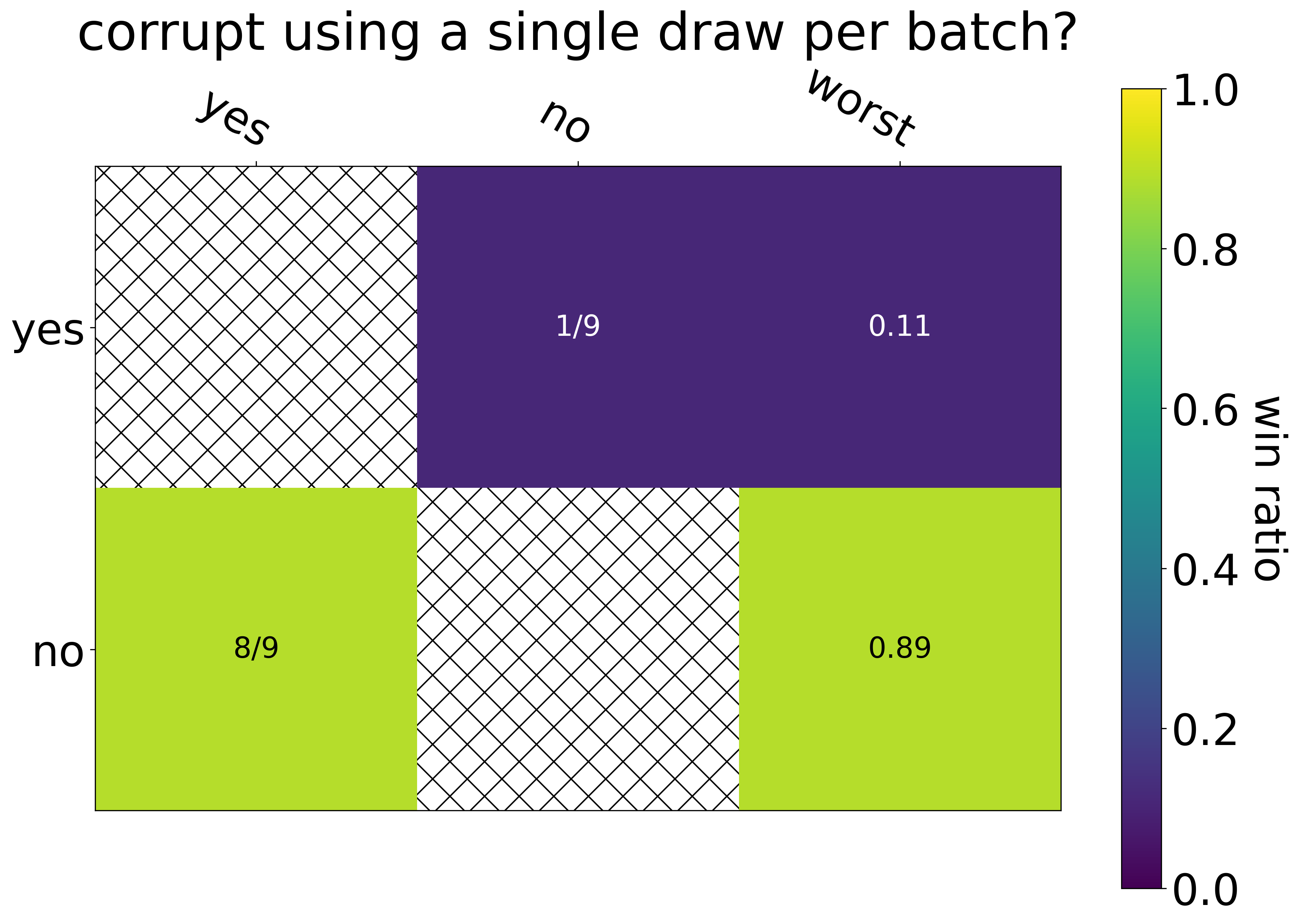}
    \caption{\textbf{Top left}: Corrupting one view is better than corrupting both. \textbf{Top right}: Using different random feature indices for each example in the mini-batch is better than using a same set across the batch. \textbf{Bottom left}: Selecting a variable number of feature indices via coin flips performs similar to the method described in Algorithm~\ref{alg:main}. \textbf{Bottom right}: Corrupting by replacing the features by the features of a \emph{single} drawn example for the whole mini-batch performs worse.} 
    \label{fig:single_batch_ablation}
\end{figure}

Figure~\ref{fig:single_batch_ablation} shows the following points.
\begin{itemize}
\item Corrupting one view is better than corrupting both the views. The likely explanation for this is that at a corruption rate of $60\%$, corrupting both views i.i.d means that the fraction of feature indices that contain the same value for both views is small - in other words, there is less information between the two views and the contrastive task is harder.

\item Corrupting the same set of feature indices within the mini-batch performs slightly worse than random sampling for every example in the mini-batch.

\item An alternative way to select feature indices to corrupt is to use Bernoulli's: for each feature index, corrupt it with probability (corruption rate) $c$, ensuring that at least one index is corrupted. In the selection method we describe in Algorithm~\ref{alg:main}, a constant number of indices are corrupted, whereas here it is variable. This alternative way of selecting indices performs roughly the same.

\item Once the feature indices to corrupt are determined, rather than corrupting them by drawing from the empirical marginal distribution, we can instead draw a random example from the training set and use its feature entries for corrupting \emph{all} views for the mini-batch. This performs worse.

\end{itemize}

\subsubsection*{Alternative losses}
\begin{figure}[!t]
    \centering
    \includegraphics[width=0.48\textwidth]{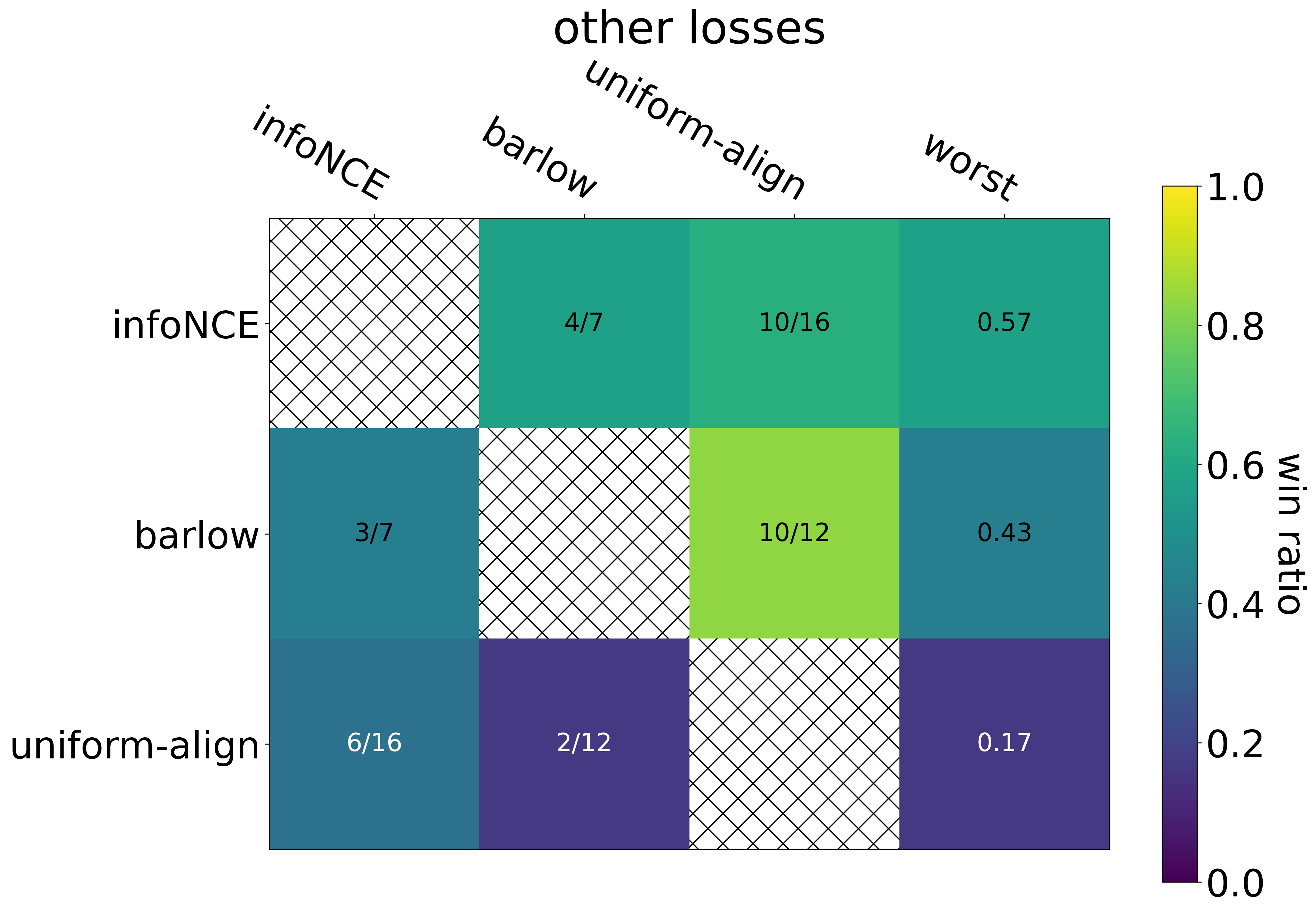}
    \includegraphics[width=0.48\textwidth]{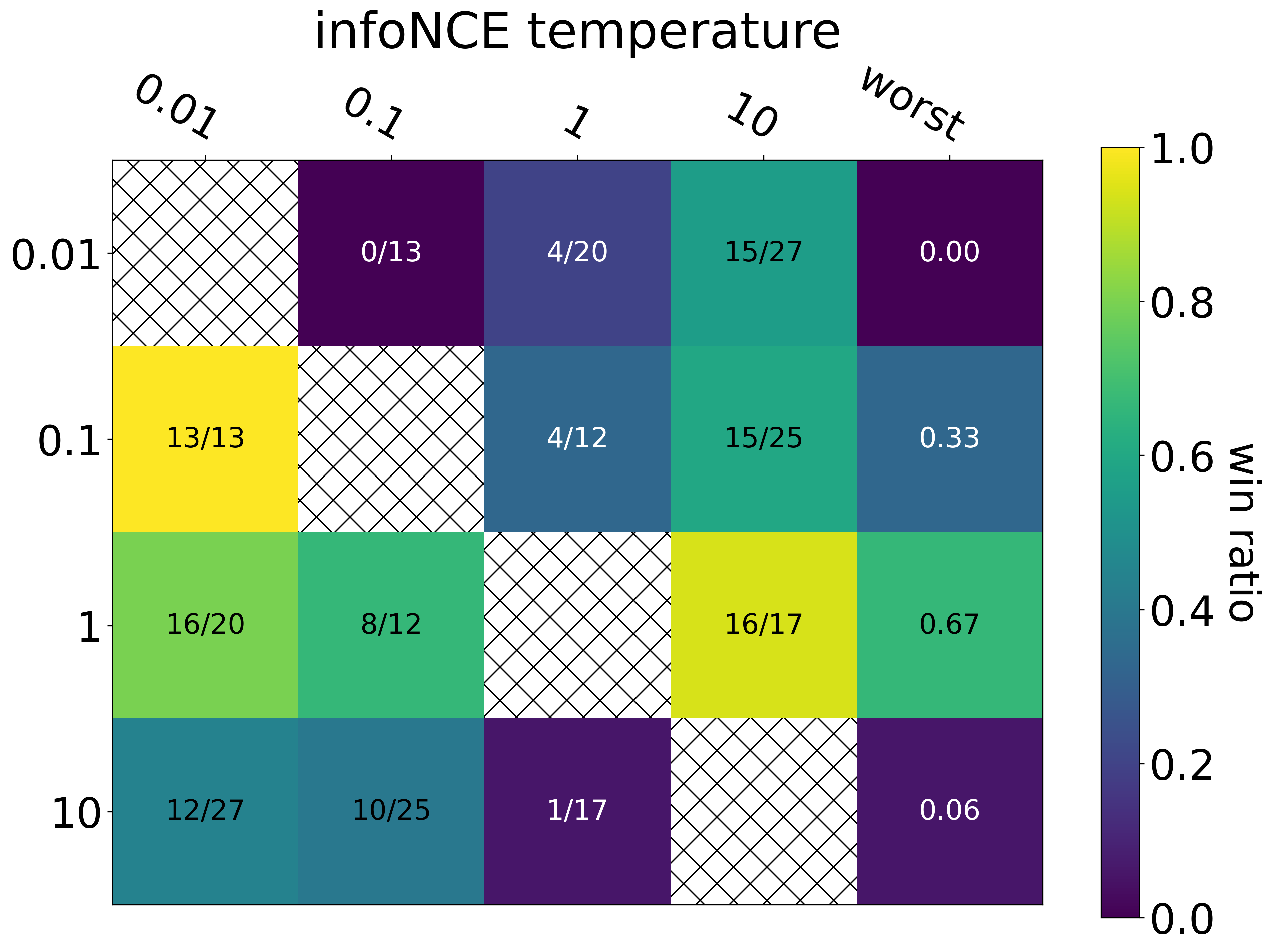}
    \caption{\textbf{Left}: Barlow Twins loss performs similar to InfoNCE while Uniform-Align performs worse. \textbf{Right}: InfoNCE softmax temperature 1 performs well.} 
    \label{fig:loss_temp_ablation}
\end{figure}

Figure~\ref{fig:loss_temp_ablation} shows the impact of using two recently-proposed alternatives to the InfoNCE loss: Uniform and Align~\citep{wang2020understanding} and Barlow Twins~\citep{zbontar2021barlow}. We use 5e-3 for the hyperparameter in Barlow, and equal weighting between the align and uniform loss terms. We find no benefit in using these other losses.

\begin{figure}[!t]
    \centering
    \includegraphics[width=0.48\textwidth]{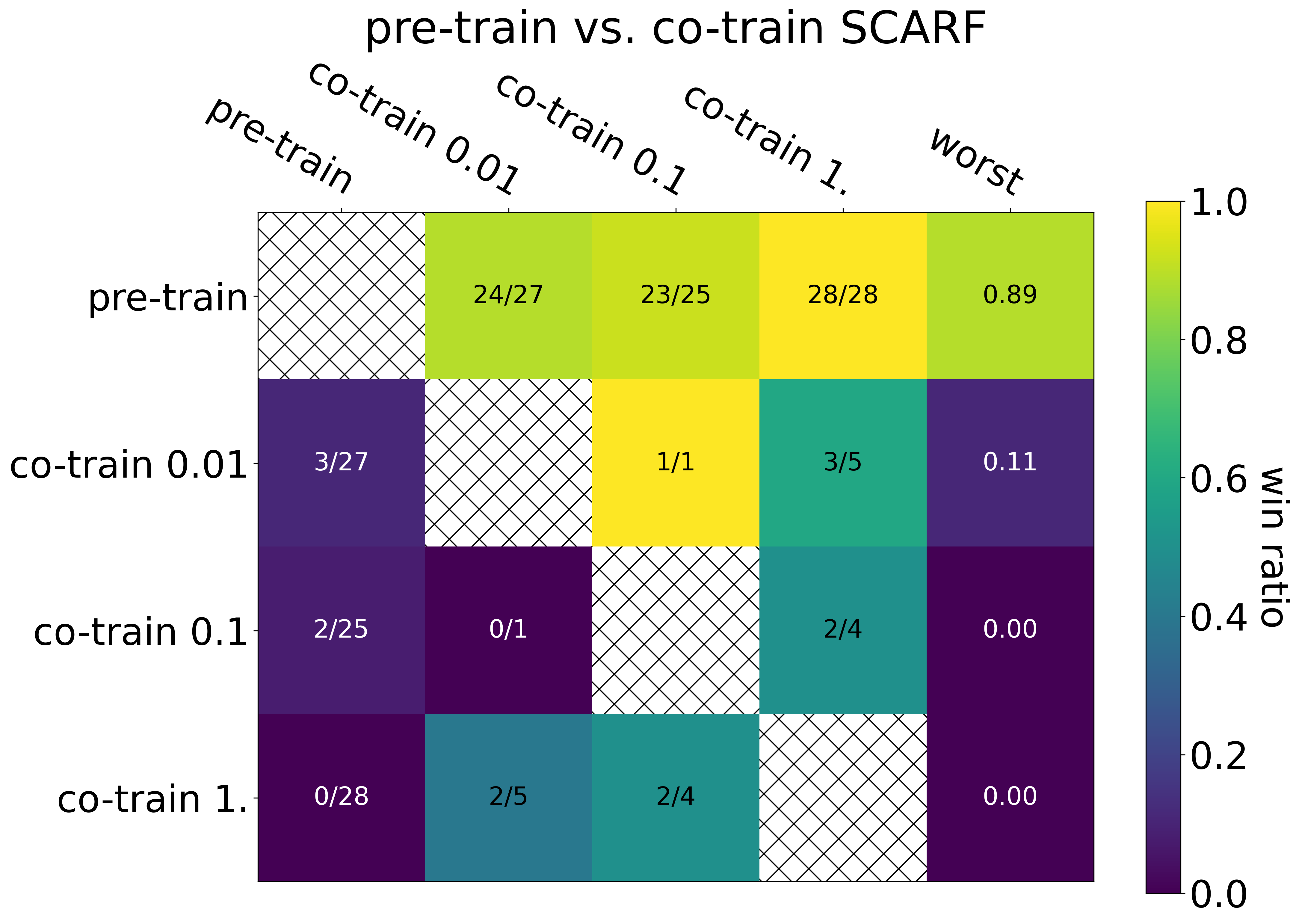}
    \includegraphics[width=0.48\textwidth]{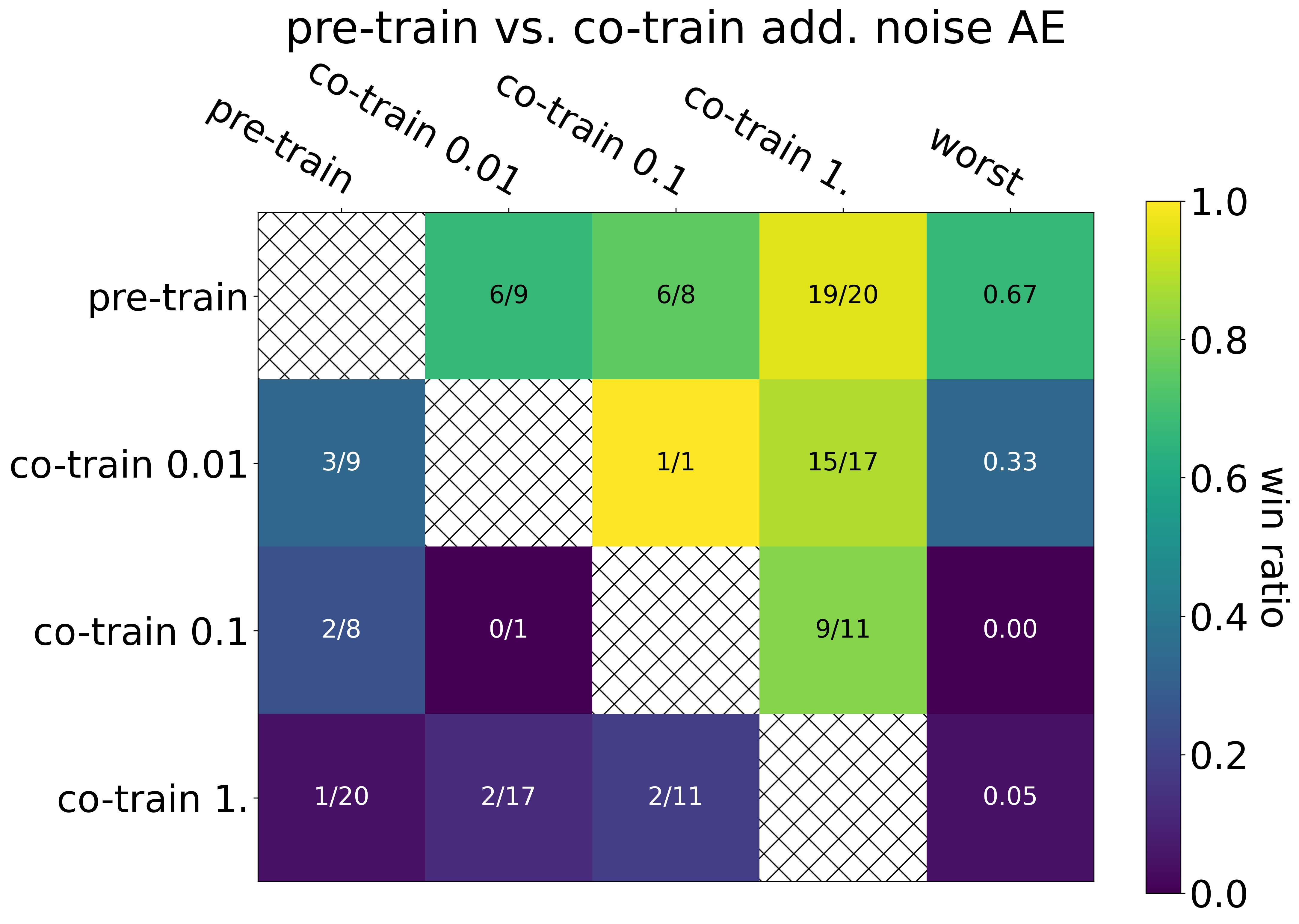}
    \caption{\textbf{Left:} Pre-training with \textsc{Scarf} beats co-training for a range of weights on the co-training term. \textbf{Right:} The same is true for additive noise autoencoders.} 
    \label{fig:cotrain_ablation}
\end{figure}

\begin{figure}[!t]
    \centering
    \includegraphics[width=0.48\textwidth]{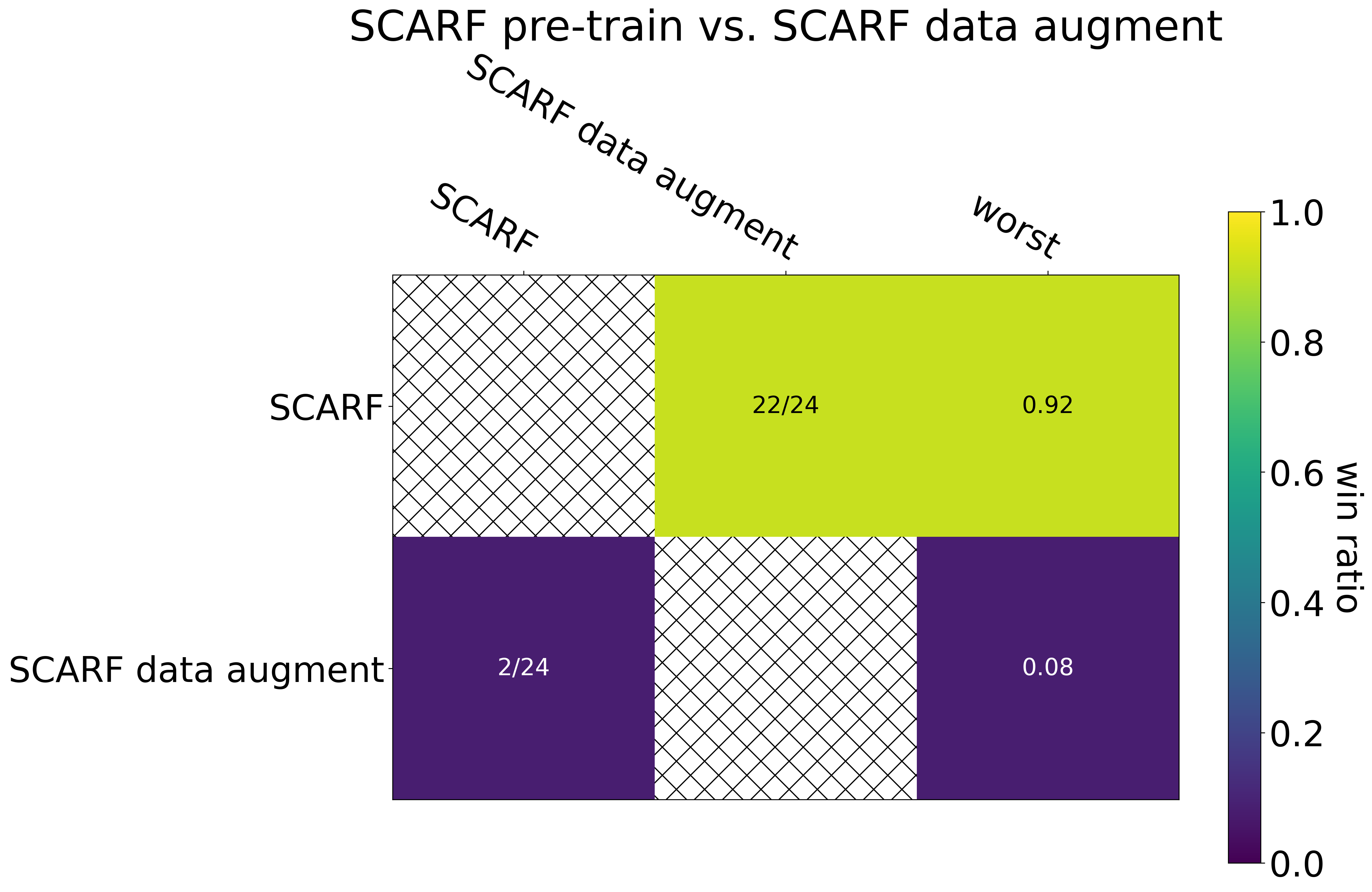}    
    \includegraphics[width=0.48\textwidth]{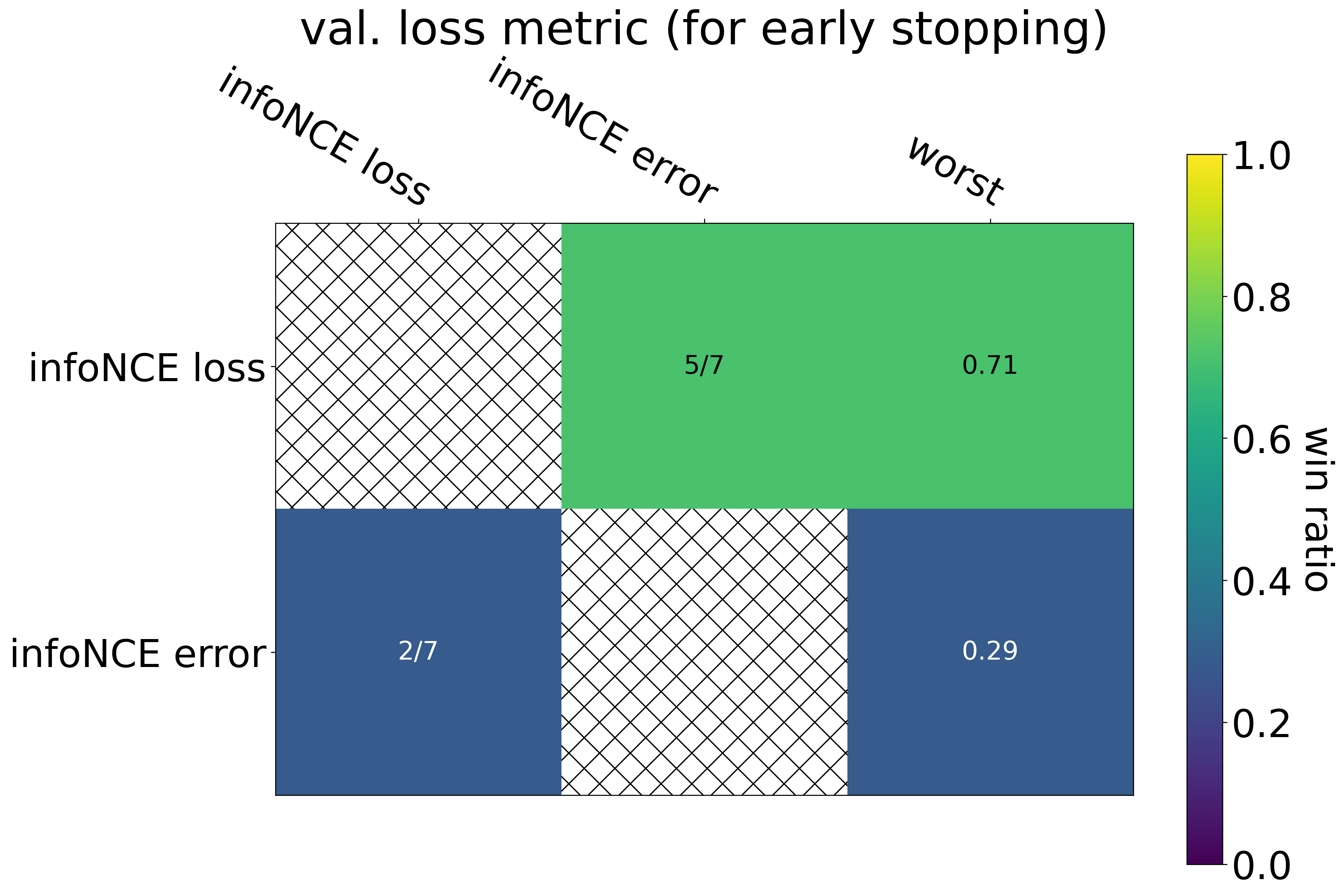}
    \caption{\textbf{Left}: Using \textsc{Scarf} only for data augmentation during supervised training performs worse than using it for pre-training. \textbf{Right}: Using InfoNCE error instead of InfoNCE loss as the validation metric for early stopping degrades performance.} 
    \label{fig:pretrain_val_metric}
\end{figure}

\subsubsection*{\textsc{Scarf} pre-training outperforms \textsc{Scarf} co-training}
Here we address the question: is it better to pre-train using \textsc{Scarf} or to apply it as a term in the supervised training loss? In particular,
$\mathcal{L}_\text{co-train} = \mathcal{L}_\text{supervised} + \lambda_\text{cont} \mathcal{L}_\text{cont}$, where $\mathcal{L}_\text{cont}$ is as described in Algorithm~\ref{alg:main}. Figure~\ref{fig:cotrain_ablation} shows that pre-training outperforms co-training for a range of different $\lambda_\text{cont}$. We see this is also the case for additive noise autoencoders.

\subsubsection*{\textsc{Scarf} pre-training outperforms \textsc{Scarf} data augmentation}
Figure~\ref{fig:pretrain_val_metric} shows that using \textsc{Scarf} only for data augmentation during supervised training performs far worse than using it for pre-training.

\subsubsection*{Using InfoNCE error as the pre-training validation metric is worse}
\textsc{Scarf} uses the same loss function (InfoNCE) for training and validation. If we instead use the InfoNCE error for validation - where an error occurs when the model predicts an off-diagonal entry of the batch-size by batch-size similarity matrix - downstream performance degrades, as shown in Figure~\ref{fig:pretrain_val_metric}.

\begin{figure}[!t]
    \centering
    \includegraphics[width=0.48\textwidth]{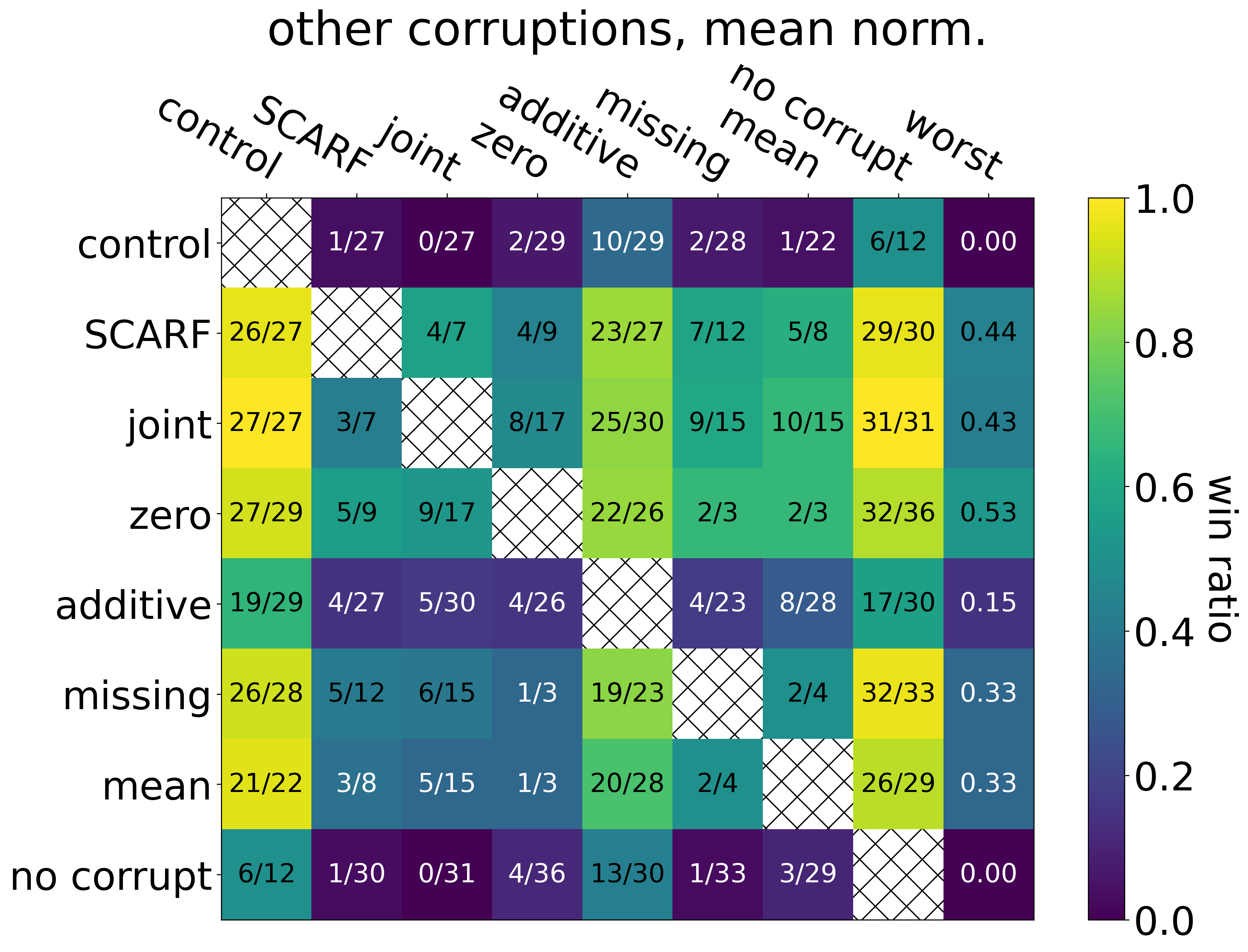}
    \includegraphics[width=0.48\textwidth]{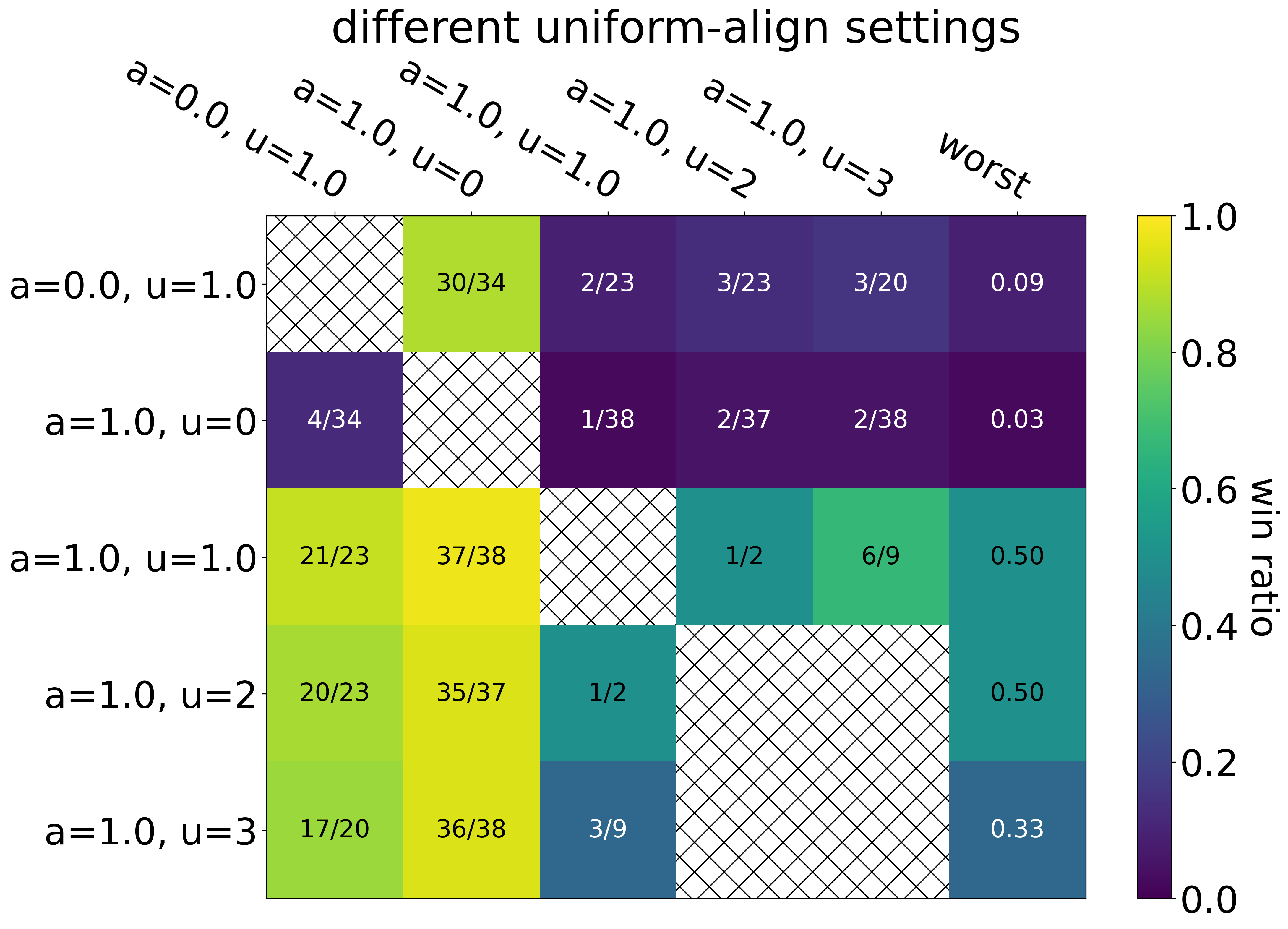}
    \caption{\textbf{Left:} \textsc{Scarf}'s corruption strategy remains competitive for ``mean'' feature scaling, as was the case for both z-score and min-max scaling. \textbf{Right:} Comparison of different hyperparameters for the uniform-align loss. $a$ and $u$ are the weights on the align and uniform terms respectively.}
    \label{fig:mean_scaling}
\end{figure}

\subsubsection*{More on uniform-align loss}
Figure~\ref{fig:mean_scaling} compares \textsc{Scarf} using the uniform-align loss, for different weights between the align and uniform loss terms. The best performance is achieved with equal weight between the two, and Figure~\ref{fig:loss_temp_ablation} shows that this underperforms vanilla InfoNCE. We thus recommend using the vanilla InfoNCE loss.

\subsubsection*{Absolute performance}
We have thus far presented results in terms of relative improvement over control since relative gain is easier to summarize across the many datasets we consider. In this section, we present the absolute test accuracy for all methods and datasets, for the keen reader. We also add a gradient-boosted decision tree baseline using the XGBoost package, since decision trees are common among practitioners working with tabular data. 
We use the Python API~\footnote{\url{https://xgboost.readthedocs.io/en/latest/python/python_api.html}}, version 0.6, and choose the default settings for XGBClassifier (max depth of 3, 100 estimators, learning rate of 0.1).

Tables \ref{tab:absolute_acc}, \ref{tab:absolute_labelnoise}, \ref{tab:absolute_se} show the absolute test accuracies (averaged over the 30 trials) for each dataset and baseline for the 100\% training data, the 30\% label noise, and the 25\% (semi-supervised) training data settings respectively.
\begin{table}[!t]
\scriptsize
\centering
\begin{tabular}{|c|c|c|c|c|c|c|c|}
\hline
\textbf{dataset id} & \textbf{control} & \textbf{SCARF} & \textbf{SCARF AE} & \textbf{add. noise AE} & \textbf{no noise AE} & \textbf{SCARF disc.} & \textbf{XGBoost} \\ \hline
6                   & 94.69            & 95.75          & 94.6              & 94.56                  & 94.36                & 95.1                 & 88.01            \\
6332                & 74.24            & 75.4           & 71.76             & 69.35                  & 66.6                 & 69.34                & 77.59            \\
54                  & 75.45            & 76.75          & 75.14             & 75.83                  & 75.88                & 78.44                & 76.16            \\
50                  & 90.54            & 97.53          & 98.68             & 97.33                  & 95.91                & 96.66                & 90.61            \\
46                  & 93.47            & 94.59          & 94.63             & 94.96                  & 94.33                & 93.36                & 95.72            \\
469                 & 18.54            & 19.08          & 19.89             & 19.14                  & 19.79                & 19.17                & 21.21            \\
458                 & 98.96            & 99.51          & 99.34             & 99.2                   & 99.11                & 99.19                & 98.46            \\
4538                & 56.28            & 60.32          & 56.78             & 56.89                  & 56.84                & 57.06                & 56.31            \\
4534                & 96.08            & 95.85          & 95.91             & 95.8                   & 95.9                 & 96.46                & 94.66            \\
44                  & 93.4             & 94.46          & 93.53             & 93.45                  & 93.49                & 93.55                & 94.54            \\
4134                & 75.53            & 75.51          & 76.25             & 75.52                  & 75.77                & 75.47                & 78.31            \\
41027               & 90.96            & 91.13          & 91.73             & 92.15                  & 92.32                & 91.68                & 81.75            \\
40994               & 90.97            & 90.91          & 91.02             & 91.5                   & 90.99                & 90.86                & 94.29            \\
40984               & 90.56            & 91.97          & 90.32             & 90.46                  & 90.53                & 90.69                & 92.66            \\
40983               & 98.59            & 98.7           & 98.64             & 98.66                  & 98.66                & 98.77                & 98.6             \\
40982               & 73.48            & 73.27          & 73.47             & 73.7                   & 73.56                & 72.99                & 78.07            \\
40979               & 96.68            & 97.45          & 96.36             & 96.39                  & 96.6                 & 96.59                & 95.91            \\
40978               & 94.84            & 95.13          & 95.36             & 95.81                  & 94.75                & 94.74                & 97.33            \\
40975               & 95.27            & 97.26          & 96.86             & 97.02                  & 96.76                & 96.7                 & 95.87            \\
40966               & 98.44            & 98.77          & 98.21             & 98.46                  & 98.68                & 97.87                & 97.05            \\
40923               & 2.17             & 2.19           & 2.19              & 2.19                   & 2.19                 & 2.19                 & 77.49            \\
40701               & 91.73            & 92.36          & 92.36             & 92.27                  & 92.14                & 92.29                & 95.31            \\
40670               & 92.37            & 92.88          & 92.88             & 93.27                  & 92.25                & 92.53                & 96.27            \\
40668               & 83.38            & 82.97          & 83.29             & 83.17                  & 83.26                & 83.17                & 74.11            \\
40499               & 99.05            & 98.9           & 98.65             & 98.64                  & 98.95                & 98.63                & 97.58            \\
3                   & 97.55            & 98.53          & 97.72             & 97.65                  & 97.94                & 98.06                & 97.6             \\
38                  & 97.14            & 97.75          & 97.09             & 97.19                  & 97.18                & 97.45                & 98.42            \\
37                  & 75.21            & 75.36          & 75.15             & 75.58                  & 75.57                & 74.58                & 76.47            \\
32                  & 98.95            & 99.25          & 98.81             & 98.82                  & 98.89                & 98.9                 & 98.39            \\
31                  & 73.55            & 73.86          & 73.3              & 72.68                  & 72.92                & 71.92                & 76.38            \\
307                 & 94.89            & 96.45          & 94.29             & 95.17                  & 94.17                & 94.44                & 86.48            \\
300                 & 94.46            & 94.37          & 94.3              & 94.44                  & 94.18                & 94.47                & 94.73            \\
29                  & 85.82            & 86.35          & 85.62             & 85.57                  & 85.14                & 83.7                 & 86.16            \\
28                  & 97.56            & 97.88          & 97.63             & 97.76                  & 97.44                & 97.55                & 97.29            \\
23                  & 54.64            & 54.25          & 53.22             & 53.27                  & 53.03                & 53.41                & 55.77            \\
23517               & 51.59            & 51.55          & 51.53             & 51.51                  & 51.69                & 51.57                & 52               \\
23381               & 60.59            & 59.07          & 56.78             & 56.68                  & 53.22                & 54.62                & 59.7             \\
22                  & 82.05            & 80.86          & 81.18             & 81.29                  & 81.48                & 82.05                & 77.43            \\
18                  & 72.59            & 74.06          & 73.51             & 72.94                  & 73.69                & 73.06                & 72.12            \\
188                 & 64.24            & 65.02          & 67.34             & 66.09                  & 64.95                & 64.7                 & 64.89            \\
182                 & 89.21            & 90.43          & 89.15             & 89.17                  & 88.98                & 89.42                & 89.63            \\
16                  & 95.68            & 96.55          & 95.51             & 95.53                  & 95.47                & 95.14                & 94.85            \\
15                  & 96.45            & 97.15          & 97.06             & 96.39                  & 96.04                & 96.07                & 95.83            \\
1590                & 85.55            & 85.6           & 85.51             & 85.51                  & 85.54                & 85.5                 & 86.45            \\
151                 & 82.77            & 83.55          & 82.69             & 82.67                  & 82.34                & 83.18                & 83.27            \\
1510                & 96.94            & 96.49          & 96.27             & 96.54                  & 96.21                & 96.58                & 96.26            \\
1501                & 91.41            & 93.76          & 91.55             & 91.6                   & 92.49                & 92.35                & 91.33            \\
14                  & 79.27            & 80.99          & 81.05             & 80.67                  & 80.89                & 78.82                & 82.34            \\
1497                & 88.73            & 93.65          & 89.19             & 88.51                  & 88.94                & 89.21                & 99.69            \\
1494                & 86.76            & 86.71          & 86.96             & 86.59                  & 87.29                & 86.44                & 86.32            \\
1489                & 85.25            & 87.55          & 85.84             & 85.86                  & 85.03                & 87.15                & 85.63            \\
1487                & 93.47            & 94.25          & 94.27             & 94.31                  & 94.27                & 93.96                & 94.34            \\
1486                & 95.74            & 95.66          & 95.74             & 95.79                  & 95.76                & 95.74                & 95.63            \\
1485                & 54.2             & 70.16          & 68.76             & 68.24                  & 64.12                & 56.73                & 71.6             \\
1480                & 70.16            & 69.74          & 69.49             & 69.19                  & 69.23                & 69.64                & 68.01            \\
1478                & 96.85            & 97.16          & 97.08             & 97.12                  & 97.07                & 94.98                & 98.38            \\
1475                & 54.23            & 54.54          & 54.02             & 53.85                  & 53.7                 & 53.97                & 55.72            \\
1468                & 93.45            & 94.04          & 92.95             & 91.68                  & 92.72                & 85.79                & 91.27            \\
1464                & 79.29            & 78.53          & 79.31             & 78.6                   & 79.13                & 78.04                & 77.58            \\
1462                & 99.5             & 99.94          & 99.75             & 99.72                  & 99.77                & 99.92                & 99.49            \\
1461                & 90.42            & 90.46          & 90.43             & 90.43                  & 90.41                & 90.45                & 90.36            \\
12                  & 96.82            & 96.46          & 95.96             & 95.99                  & 95.97                & 96.42                & 96.17            \\
11                  & 95.98            & 97.27          & 96.87             & 97.15                  & 97.05                & 97.16                & 87.95            \\
1068                & 93.04            & 93.18          & 93.23             & 93.06                  & 93.12                & 93.18                & 93.78            \\
1067                & 85.6             & 85.69          & 85.66             & 85.57                  & 85.45                & 85.46                & 85.36            \\
1063                & 84.17            & 84.06          & 84.06             & 83.71                  & 82.16                & 83.7                 & 83.27            \\
1053                & 81.47            & 81.19          & 81.28             & 81.21                  & 81.27                & 81.19                & 81.44            \\
1050                & 89.84            & 89.49          & 89.24             & 89.27                  & 89.4                 & 89.7                 & 88.9             \\
1049                & 90.65            & 89.97          & 90.2              & 90.38                  & 90.74                & 90.17                & 91.2 \\
\hline
\end{tabular}

\caption{\label{tab:absolute_acc} Absolute test accuracy (percent; averaged over 30 trials) for the 100\% training data setting, for each dataset and every baseline, including a gradient-boosted decision tree (XGBoost) baseline (max depth of 3, 100 estimators).}
\end{table}
\begin{table}[!t]
\centering
\scriptsize
\begin{tabular}{|l|l|l|l|l|l|l|l|}
\hline
\textbf{dataset id} & \textbf{control} & \textbf{SCARF} & \textbf{SCARF AE} & \textbf{add. noise AE} & \textbf{no noise AE} & \textbf{SCARF disc.} & \textbf{XGBoost} \\ \hline
6                   & 90.45            & 90.57          & 90.02             & 90.01                  & 89.65                & 90.7                 & 85.53            \\
6332                & 69.39            & 70.99          & 63.41             & 62.78                  & 60.25                & 62.96                & 72.75            \\
54                  & 72.21            & 71.94          & 70.84             & 71.11                  & 71.93                & 72.76                & 72.92            \\
50                  & 79.64            & 87.66          & 90.52             & 86.7                   & 85.14                & 88.12                & 87.4             \\
46                  & 84.43            & 92.1           & 92.76             & 93.05                  & 91.89                & 85.77                & 94               \\
469                 & 19.47            & 19.02          & 20.08             & 19.89                  & 19.41                & 18.14                & 19.52            \\
458                 & 96.65            & 99.27          & 98.06             & 97.97                  & 97.1                 & 95.47                & 96.07            \\
4538                & 52.24            & 54.25          & 52.07             & 52.15                  & 51.94                & 52.35                & 54.88            \\
4534                & 94.13            & 94.38          & 94.2              & 94.09                  & 94                   & 94.31                & 93.66            \\
44                  & 91.37            & 93.21          & 91.67             & 91.34                  & 91.31                & 91.76                & 93.13            \\
4134                & 72.09            & 71.17          & 71.99             & 72.13                  & 72.2                 & 72.31                & 76.95            \\
41027               & 85.54            & 85.31          & 85.43             & 85.42                  & 85.46                & 85.53                & 81.54            \\
40994               & 91.33            & 91.37          & 90.74             & 91.16                  & 91.53                & 91.91                & 90.12            \\
40984               & 88.63            & 89.99          & 88.44             & 88.51                  & 88.08                & 88.68                & 90.71            \\
40983               & 97.87            & 98.5           & 98.3              & 98.14                  & 98.07                & 98.57                & 97.86            \\
40982               & 70.54            & 70.92          & 71.26             & 70.97                  & 71.11                & 70.19                & 73.69            \\
40979               & 92.68            & 96.31          & 92.62             & 92.7                   & 92.57                & 93.76                & 92.43            \\
40978               & 91.87            & 92.77          & 93.81             & 94.12                  & 93.64                & 92.52                & 96.48            \\
40975               & 88.64            & 90.26          & 91.42             & 90.91                  & 91.27                & 88.54                & 91.36            \\
40966               & 90.78            & 88.52          & 88.65             & 90.12                  & 91.86                & 87.37                & 89.58            \\
40923               & 2.16             & 2.17           & 2.17              & 2.17                   & 2.17                 & 2.17                 & 75.47            \\
40701               & 88.77            & 89.34          & 88.87             & 88.54                  & 88.92                & 89.41                & 93.7             \\
40670               & 84.41            & 90.34          & 89.6              & 90.39                  & 87.63                & 87.66                & 93.94            \\
40668               & 80.62            & 80.27          & 80.66             & 80.7                   & 80.71                & 80.55                & 73.67            \\
40499               & 98.18            & 97.85          & 97.82             & 98.11                  & 98.16                & 98.29                & 95.69            \\
3                   & 94.07            & 94.77          & 94.93             & 95.14                  & 95.21                & 93.59                & 96.07            \\
38                  & 96.2             & 96.88          & 96.1              & 96.19                  & 95.9                 & 96.3                 & 97.81            \\
37                  & 75.68            & 74.6           & 73.99             & 73.8                   & 74.27                & 73.17                & 72.71            \\
32                  & 98.47            & 98.6           & 98.43             & 98.51                  & 98.43                & 98.56                & 97.12            \\
31                  & 71.15            & 72.12          & 70.66             & 71.04                  & 70.63                & 66.98                & 73               \\
307                 & 78.65            & 81.37          & 80.8              & 79.83                  & 78.7                 & 78.07                & 74.95            \\
300                 & 89.13            & 87.21          & 89.1              & 89.21                  & 88.9                 & 89.2                 & 90.89            \\
29                  & 84.12            & 83.21          & 84.28             & 83.86                  & 84.18                & 81.53                & 82.44            \\
28                  & 95.77            & 96.54          & 96.1              & 95.94                  & 95.77                & 95.45                & 95.22            \\
23                  & 51               & 51.89          & 50.29             & 50.03                  & 50.21                & 51.37                & 53.5             \\
23517               & 51.33            & 51.12          & 51.29             & 51.25                  & 51.38                & 51.26                & 51.7             \\
23381               & 58.07            & 57.88          & 55.73             & 55.73                  & 54.7                 & 53.25                & 55.47            \\
22                  & 78.75            & 78.05          & 78.56             & 78.86                  & 79.15                & 78.8                 & 74.94            \\
18                  & 70.8             & 73.25          & 72.63             & 71.93                  & 73.1                 & 72.72                & 71.23            \\
188                 & 58.3             & 53.19          & 58.02             & 57.64                  & 57.09                & 55.3                 & 60.29            \\
182                 & 88.31            & 88.88          & 87.82             & 87.83                  & 87.36                & 87.59                & 88.25            \\
16                  & 88.19            & 94.62          & 92.02             & 91.56                  & 89.75                & 87.96                & 89.57            \\
15                  & 96.31            & 97.2           & 96.77             & 96.17                  & 95.56                & 95.87                & 93.6             \\
1590                & 84.81            & 85.19          & 85.07             & 85.06                  & 85.06                & 85.05                & 86.23            \\
151                 & 80.08            & 81.15          & 80.06             & 80.14                  & 80.02                & 80.52                & 82.97            \\
1510                & 94.58            & 94.37          & 92.51             & 92.24                  & 92.18                & 93.83                & 92.13            \\
1501                & 78.73            & 89.29          & 81.16             & 81.73                  & 82.16                & 84.43                & 85.6             \\
14                  & 72.34            & 78.57          & 77.87             & 77.52                  & 74.92                & 71.53                & 77.64            \\
1497                & 82.85            & 90.2           & 82.64             & 82.24                  & 82.82                & 82.72                & 98.75            \\
1494                & 84.18            & 84.97          & 85.03             & 85.06                  & 85.18                & 84.72                & 83.4             \\
1489                & 83.29            & 85.39          & 84.16             & 84.11                  & 83.48                & 84.62                & 84.26            \\
1487                & 92.61            & 92.97          & 93.27             & 93.41                  & 92.9                 & 93.16                & 93.55            \\
1486                & 94.89            & 94.71          & 94.91             & 94.95                  & 94.91                & 94.96                & 95.12            \\
1485                & 53.97            & 67.03          & 66.74             & 66.62                  & 62.55                & 54.21                & 65.67            \\
1480                & 71.63            & 69.72          & 68.86             & 68.95                  & 68.9                 & 69.53                & 68.06            \\
1478                & 94.13            & 94.42          & 94.75             & 94.71                  & 94.71                & 94.74                & 96.37            \\
1475                & 51.47            & 51.9           & 51.19             & 51.29                  & 51.12                & 51.28                & 53.8             \\
1468                & 83.07            & 90.76          & 83.39             & 79.58                  & 78.46                & 64.89                & 87.3             \\
1464                & 76.65            & 76.91          & 77.69             & 76.9                   & 78.28                & 76.63                & 75.71            \\
1462                & 98.97            & 99.55          & 99.14             & 99.21                  & 99.52                & 99.69                & 98.07            \\
1461                & 90.06            & 90.21          & 90.11             & 90.11                  & 90.11                & 90.05                & 90.37            \\
12                  & 93.99            & 95.18          & 93.93             & 93.77                  & 93.8                 & 94.2                 & 92.98            \\
11                  & 90.25            & 88.77          & 89.77             & 89.11                  & 89.88                & 89.68                & 82.43            \\
1068                & 92.45            & 92.44          & 92.32             & 92.24                  & 92.1                 & 92.73                & 92.82            \\
1067                & 84.74            & 84.87          & 84.96             & 84.53                  & 84.62                & 84.33                & 85.33            \\
1063                & 83.81            & 83.76          & 83.59             & 83.63                  & 82.95                & 83.95                & 80.95            \\
1053                & 81.05            & 81.05          & 81.06             & 81.04                  & 81.09                & 81.04                & 81.07            \\
1050                & 89.08            & 88.94          & 88.96             & 89.11                  & 89.46                & 89.32                & 88.81            \\
1049                & 88.95            & 87.48          & 88.53             & 88.69                  & 88.93                & 88.58                & 89.2             \\ \hline
\end{tabular}

\caption{\label{tab:absolute_labelnoise} Absolute test accuracy (percent; averaged over 30 trials) for the 30\% label noise setting, for each dataset and every baseline, including a gradient-boosted decision tree (XGBoost) baseline (max depth of 3, 100 estimators).}
\end{table}
\begin{table}[!t]
\scriptsize
\centering
\begin{tabular}{|c|c|c|c|c|c|c|c|}
\hline
\textbf{dataset id} & \textbf{control} & \textbf{SCARF} & \textbf{SCARF AE} & \textbf{add. noise AE} & \textbf{no noise AE} & \textbf{SCARF disc.} & \textbf{XGBoost} \\ \hline
6                   & 87.22            & 89.71          & 87.12             & 87.38                  & 87.01                & 88.87                & 85.02            \\
6332                & 63.67            & 66.71          & 58.53             & 57.82                  & 57.25                & 61.5                 & 69.41            \\
54                  & 69.27            & 66.59          & 68.31             & 68.48                  & 69.5                 & 68.85                & 69.04            \\
50                  & 72.63            & 79.2           & 86.85             & 82.9                   & 75.78                & 88.62                & 81.34            \\
46                  & 87.14            & 93.1           & 93.2              & 93.86                  & 93.28                & 89.35                & 94.6             \\
469                 & 18.02            & 18.53          & 19.31             & 19.16                  & 19.11                & 17.85                & 18.81            \\
458                 & 98.03            & 99.48          & 98.6              & 98.21                  & 97.54                & 98.16                & 95.58            \\
4538                & 50.65            & 53.75          & 50.79             & 50.32                  & 50.45                & 50.03                & 54.05            \\
4534                & 94.1             & 94.31          & 93.89             & 94.06                  & 93.93                & 94.3                 & 94.22            \\
44                  & 91.6             & 93.6           & 92                & 91.95                  & 91.63                & 92.5                 & 93.41            \\
4134                & 68.31            & 68.44          & 69.65             & 69                     & 68.78                & 68.45                & 74.58            \\
41027               & 84.3             & 84.63          & 84.7              & 84.66                  & 84.68                & 85.24                & 80.84            \\
40994               & 90.63            & 90.96          & 90.96             & 90.93                  & 90.97                & 90.96                & 92.69            \\
40984               & 86.48            & 88.05          & 86.12             & 85.83                  & 86.34                & 86.49                & 89.96            \\
40983               & 94.72            & 97.65          & 94.85             & 94.65                  & 97.72                & 98.58                & 97.92            \\
40982               & 68.25            & 67.99          & 67.62             & 68.08                  & 68.62                & 67.11                & 70.14            \\
40979               & 92.81            & 96.21          & 92.58             & 92.15                  & 92.56                & 93.55                & 91.33            \\
40978               & 86.05            & 89.95          & 91.12             & 91.49                  & 88.9                 & 85.88                & 96.32            \\
40975               & 72.47            & 90.55          & 91.14             & 90.53                  & 90.02                & 71.68                & 91.98            \\
40966               & 86.53            & 84.11          & 83.98             & 86.01                  & 88.3                 & 83.48                & 80.65            \\
40923               & 2.16             & 2.16           & 2.16              & 2.16                   & 2.16                 & 2.16                 & 74.35            \\
40701               & 87.44            & 89.59          & 88.76             & 88.49                  & 88.44                & 88.34                & 93.46            \\
40670               & 81.67            & 90.97          & 89.69             & 90.27                  & 87.58                & 88.22                & 94.37            \\
40668               & 79.64            & 79.28          & 79.41             & 79.72                  & 79.74                & 79.37                & 74.05            \\
40499               & 97.88            & 97.65          & 97.13             & 97.59                  & 98.13                & 97.78                & 94.59            \\
3                   & 93.04            & 95.44          & 94.02             & 94.84                  & 94.34                & 94.59                & 96.16            \\
38                  & 95.13            & 96.92          & 95.39             & 95.17                  & 94.53                & 95.39                & 97.7             \\
37                  & 70.37            & 72.24          & 71.15             & 71.2                   & 70.21                & 71.58                & 72.94            \\
32                  & 98.39            & 98.61          & 98.19             & 98.15                  & 97.94                & 98.4                 & 96.9             \\
31                  & 70.53            & 71.66          & 68.48             & 68.54                  & 65.82                & 69.33                & 73.12            \\
307                 & 69.68            & 72.47          & 73.49             & 72.6                   & 72.31                & 70.88                & 63.97            \\
300                 & 91.66            & 87.01          & 89.98             & 90.42                  & 90.58                & 91.06                & 90.3             \\
29                  & 83.6             & 83.57          & 80.7              & 81.46                  & 81.93                & 82.28                & 84.59            \\
28                  & 96.43            & 96.56          & 96.38             & 96.08                  & 96.23                & 96.12                & 95.07            \\
23                  & 49.56            & 51.45          & 45.64             & 46.48                  & 44.84                & 49.95                & 51.48            \\
23517               & 51.17            & 51.04          & 51.03             & 50.98                  & 51.03                & 51.23                & 51.52            \\
23381               & 58.83            & 58.7           & 55.62             & 55.37                  & 53.37                & 56.43                & 55.5             \\
22                  & 77.72            & 77.03          & 76.27             & 76.8                   & 77.8                 & 76.34                & 73.72            \\
18                  & 70.71            & 71.37          & 71.05             & 70.53                  & 72.39                & 71.11                & 70.11            \\
188                 & 55.18            & 46.41          & 50.56             & 53.42                  & 53.06                & 51.72                & 57.82            \\
182                 & 86.98            & 88.36          & 87.37             & 87.34                  & 86.53                & 86.62                & 87.89            \\
16                  & 88.25            & 93.91          & 90.7              & 90.44                  & 89.6                 & 89.97                & 87.54            \\
15                  & 96.65            & 97.01          & 96.7              & 96.45                  & 94.54                & 95.54                & 95.19            \\
1590                & 84.9             & 85.01          & 84.89             & 84.89                  & 84.92                & 84.86                & 86.34            \\
151                 & 79.49            & 80.7           & 79.48             & 79.61                  & 79.4                 & 80.13                & 82.68            \\
1510                & 94.63            & 94.36          & 93.8              & 93.87                  & 92.73                & 94.23                & 94.12            \\
1501                & 81.01            & 88.29          & 79.66             & 80.16                  & 81.53                & 83.97                & 81.09            \\
14                  & 73.13            & 76.71          & 76.51             & 76.3                   & 75.2                 & 72.6                 & 76.36            \\
1497                & 80.19            & 90.15          & 80.88             & 80.14                  & 80.18                & 82.09                & 98.96            \\
1494                & 84.13            & 84.11          & 84.46             & 84.65                  & 84.39                & 83.67                & 83.11            \\
1489                & 81.95            & 84.81          & 83.27             & 82.72                  & 82.04                & 84.18                & 83.97            \\
1487                & 93.71            & 93.72          & 93.59             & 93.41                  & 93.68                & 93.7                 & 94.02            \\
1486                & 94.75            & 94.64          & 94.75             & 94.84                  & 94.83                & 94.81                & 95.38            \\
1485                & 53.69            & 66.82          & 67.1              & 66.82                  & 62.54                & 53.86                & 61.14            \\
1480                & 70.5             & 69.57          & 68.72             & 69.42                  & 69.62                & 70.63                & 68.52            \\
1478                & 93.82            & 94.06          & 93.92             & 94.41                  & 93.68                & 93.45                & 96.9             \\
1475                & 49.24            & 50.63          & 49.67             & 49.35                  & 49.65                & 49.06                & 52.08            \\
1468                & 85.01            & 88.74          & 80.14             & 74.45                  & 77.73                & 71.39                & 81.67            \\
1464                & 75.26            & 75.37          & 75.7              & 75.36                  & 75.4                 & 75.17                & 75.82            \\
1462                & 98.48            & 99.13          & 98.9              & 98.78                  & 99.39                & 99.58                & 97.36            \\
1461                & 89.87            & 89.94          & 90.04             & 89.94                  & 89.92                & 89.81                & 90.22            \\
12                  & 91.83            & 94.38          & 92.78             & 92.68                  & 92.95                & 92.86                & 91.49            \\
11                  & 88.38            & 87.99          & 85.48             & 85.84                  & 89.13                & 84.27                & 82.43            \\
1068                & 92.97            & 92.75          & 92.97             & 92.91                  & 92.93                & 93                   & 92.39            \\
1067                & 84.88            & 84.87          & 84.77             & 84.7                   & 84.27                & 84.06                & 84.24            \\
1063                & 82.22            & 82.78          & 81.81             & 81.3                   & 80.33                & 82.38                & 81.21            \\
1053                & 80.82            & 80.91          & 80.87             & 80.89                  & 80.95                & 80.94                & 80.98            \\
1050                & 90.2             & 89.3           & 89.46             & 89.17                  & 89.06                & 89.73                & 88.43            \\
1049                & 87.59            & 87.14          & 87.33             & 87.47                  & 88.11                & 87.95                & 89.18            \\ \hline
\end{tabular}

\caption{\label{tab:absolute_se} Absolute test accuracy (percent; averaged over 30 trials) for the 25\% training data (semi-supervised) setting, for each dataset and every baseline, including a gradient-boosted decision tree (XGBoost) baseline (max depth of 3, 100 estimators).}
\end{table}

\subsubsection*{Training Speed}
We use an early stopping criteria when fine-tuning \textsc{Scarf} and baselines.
Table~\ref{tab:training_epochs} shows the number of actual training epochs used (averaged over the 30 trials) for each dataset and baseline for the 100\% training data setting.
\begin{table}[!t]
\centering
\scriptsize
\begin{tabular}{|c|c|c|c|c|c|c|}
\hline
\textbf{dataset id} & \textbf{control} & \textbf{SCARF} & \textbf{SCARF AE} & \textbf{add. noise AE} & \textbf{no noise AE} & \textbf{SCARF disc.} \\ \hline
6                   & 20.1             & 17.23          & 19.3              & 18.42                  & 18.77                & 18.1                 \\
6332                & 10.39            & 9.72           & 12.93             & 12.53                  & 11.1                 & 9.85                 \\
54                  & 12.03            & 12.68          & 11.83             & 12                     & 11.5                 & 13.95                \\
50                  & 12.92            & 13.58          & 10.63             & 11.65                  & 14.63                & 9.02                 \\
46                  & 9.32             & 7.8            & 7.55              & 7.15                   & 7.53                 & 7.93                 \\
469                 & 7.36             & 7.38           & 7.43              & 7.02                   & 6.97                 & 7.9                  \\
458                 & 7.62             & 5.17           & 6.12              & 6.03                   & 7.03                 & 6.57                 \\
4538                & 13.17            & 15.18          & 16.3              & 15.38                  & 14.97                & 14.13                \\
4534                & 14.03            & 12.1           & 14.88             & 13.93                  & 13.2                 & 12.68                \\
44                  & 8.86             & 8.55           & 9.85              & 9.3                    & 9.78                 & 7.52                 \\
4134                & 9.13             & 12.45          & 11.07             & 11.05                  & 10.85                & 10.48                \\
41027               & 31.3             & 33.37          & 35.62             & 37.9                   & 37.62                & 36.2                 \\
40994               & 5                & 5.3            & 5.37              & 6.03                   & 5.28                 & 5                    \\
40984               & 13.46            & 13.7           & 13.47             & 13.02                  & 13.23                & 13.53                \\
40983               & 9.91             & 7.78           & 8.8               & 9.3                    & 8.45                 & 6.75                 \\
40982               & 10.94            & 10.45          & 12.5              & 12.38                  & 11.45                & 12.57                \\
40979               & 10.65            & 7.77           & 10                & 10                     & 10.65                & 10.32                \\
40978               & 13               & 13.38          & 15.3              & 13.15                  & 13.92                & 13.3                 \\
40975               & 15.09            & 12.63          & 14.12             & 13.78                  & 14.13                & 15.33                \\
40966               & 12.62            & 11.37          & 12.28             & 11.7                   & 11.1                 & 13.87                \\
40923               & 5                & 5              & 5                 & 5                      & 5                    & 5                    \\
40701               & 10.53            & 12.72          & 11.25             & 11.48                  & 11.72                & 10.38                \\
40670               & 9.91             & 8.93           & 10.4              & 9.92                   & 9.72                 & 10.27                \\
40668               & 10.93            & 12.3           & 11.45             & 12.07                  & 12.15                & 11.92                \\
40499               & 13.19            & 10.95          & 12.55             & 12                     & 11.35                & 13.17                \\
3                   & 10.72            & 12.07          & 11.9              & 11.5                   & 12.5                 & 11.63                \\
38                  & 9.53             & 9.03           & 9.83              & 10.53                  & 9.9                  & 9.88                 \\
37                  & 8.11             & 7.33           & 7.15              & 7.53                   & 7.77                 & 7.05                 \\
32                  & 10.73            & 9.73           & 10.68             & 10.35                  & 11.3                 & 10.62                \\
31                  & 8.45             & 8.35           & 9.58              & 9.13                   & 9.87                 & 7.5                  \\
307                 & 17.08            & 13.27          & 16.78             & 16.7                   & 17.43                & 16.88                \\
300                 & 12.2             & 16.5           & 13.38             & 14.4                   & 13.28                & 14.4                 \\
29                  & 6.92             & 8.05           & 8.57              & 9.48                   & 9.15                 & 7.38                 \\
28                  & 10.68            & 9.85           & 11.83             & 12.15                  & 11.6                 & 10.98                \\
23                  & 9.06             & 8.05           & 10.38             & 10.48                  & 10.65                & 7.8                  \\
23517               & 8.37             & 7.83           & 8.45              & 8.4                    & 7.8                  & 7.62                 \\
23381               & 7.88             & 6.67           & 7.9               & 6.95                   & 7.17                 & 6.83                 \\
22                  & 10.81            & 9.37           & 11.55             & 10.48                  & 10.05                & 12.2                 \\
18                  & 13.2             & 10.52          & 12.73             & 11.93                  & 11.07                & 12.52                \\
188                 & 10.65            & 14.62          & 12.62             & 11.6                   & 11.7                 & 12.15                \\
182                 & 12.23            & 12.35          & 13.4              & 12.6                   & 12.03                & 14.77                \\
16                  & 12.68            & 8.5            & 10.25             & 10.63                  & 11.52                & 10.82                \\
15                  & 5.59             & 5.3            & 6                 & 6.05                   & 7.02                 & 6.72                 \\
1590                & 7.5              & 7.12           & 7.4               & 7.67                   & 7.53                 & 7.33                 \\
151                 & 19.5             & 14.55          & 18.27             & 19.07                  & 18.37                & 16.77                \\
1510                & 7.24             & 6.67           & 7.42              & 7.37                   & 7.85                 & 7.25                 \\
1501                & 11.84            & 9.75           & 12.48             & 12.32                  & 11.97                & 12.53                \\
14                  & 10.65            & 10.02          & 11.22             & 11.38                  & 11.38                & 12.12                \\
1497                & 13.42            & 11.82          & 15.87             & 14.83                  & 14.45                & 13.85                \\
1494                & 8.68             & 8.28           & 8.65              & 8.55                   & 8.25                 & 9.02                 \\
1489                & 13.84            & 12.52          & 14.12             & 13.97                  & 14.62                & 12.6                 \\
1487                & 7.81             & 7.98           & 7.93              & 8.17                   & 7.33                 & 7.75                 \\
1486                & 11.3             & 11.35          & 11.05             & 11.6                   & 10.33                & 11.5                 \\
1485                & 8.03             & 8.42           & 7.77              & 7.35                   & 7.02                 & 8.53                 \\
1480                & 5.44             & 6.63           & 6.75              & 6.57                   & 6.6                  & 6.25                 \\
1478                & 11.4             & 13.8           & 12.38             & 12.17                  & 11.78                & 12.57                \\
1475                & 13.16            & 11.57          & 14.57             & 13.65                  & 13.42                & 14                   \\
1468                & 12.25            & 11.12          & 14.1              & 16.35                  & 14.93                & 13.62                \\
1464                & 7.83             & 8.18           & 8.07              & 8.6                    & 6.95                 & 8.12                 \\
1462                & 8.03             & 6.4            & 7.58              & 7.43                   & 7.28                 & 5.68                 \\
1461                & 7.53             & 6.92           & 7.75              & 7.5                    & 8.02                 & 6.45                 \\
12                  & 11.16            & 8.13           & 9.73              & 10.07                  & 9.7                  & 11.68                \\
11                  & 10.31            & 10.97          & 11.48             & 10.78                  & 9.58                 & 11.38                \\
1068                & 5.69             & 6.78           & 6.18              & 6.3                    & 6                    & 6.47                 \\
1067                & 7.97             & 9.87           & 8.23              & 8.55                   & 8.1                  & 9.03                 \\
1063                & 6.45             & 6.6            & 6.28              & 6.3                    & 7.03                 & 6.8                  \\
1053                & 8.44             & 8.78           & 8.12              & 9.07                   & 8.43                 & 9.1                  \\
1050                & 5.13             & 7.27           & 7.18              & 6.95                   & 6.53                 & 5.88                 \\
1049                & 8.68             & 10.92          & 9.38              & 9.42                   & 8.8                  & 9.5                  \\ \hline
\end{tabular}

\caption{\label{tab:training_epochs} Number of actual training epochs used (averaged over 30 trials) for the 100\% training data setting, for each dataset and every baseline.}
\end{table}

\end{document}